\documentclass[10pt]{article} %
\usepackage[preprint]{tmlr}

\usepackage{hyperref}
\usepackage{url}

\title{Tractable Representation Learning with Probabilistic Circuits}

\author{%
   \name Steven Braun \email steven.braun@cs.tu-darmstadt.de \\
   \addr Technische Universität Darmstadt, Germany
   \AND
   \name Sahil Sidheekh \email sahil.sidheekh@utdallas.edu \\
   \addr University of Texas, Dallas, United States
   \AND
   \name Antonio Vergari \email avergari@ed.ac.uk \\
   \addr University of Edinburgh, United Kingdom
   \AND
   \name Martin Mundt \email mundtm@uni-bremen.de \\
   \addr Universität Bremen, Germany
   \AND
   \name Sriraam Natarajan \email sriraam.natarajan@utdallas.edu \\
   \addr University of Texas at Dallas, United States
   \AND
   \name Kristian Kersting \email kristian.kersting@cs.tu-darmstadt.de \\
   \addr Technische Universität Darmstadt, Germany
}

\def\month{7}  %
\def\year{2025} %
\def\openreview{\url{https://openreview.net/forum?id=XXXX}} %

\usepackage[utf8]{inputenc} %
\usepackage[T1]{fontenc}    %
\usepackage{hyperref}       %
\usepackage{url}            %
\usepackage{booktabs}       %
\usepackage{amsfonts}       %
\usepackage{nicefrac}       %
\usepackage{microtype}      %
\usepackage{xcolor}         %

\usepackage{tikz} %

\usepackage{amsmath,amsfonts,bm}

\def\eqref#1{equation~\ref{#1}}

\def\1{\bm{1}}

\def\rvw{{\mathbf{w}}}
\def\rvx{{\mathbf{x}}}

\def\rvz{{\mathbf{z}}}

\def\rmX{{\mathbf{X}}}

\DeclareMathAlphabet{\mathsfit}{\encodingdefault}{\sfdefault}{m}{sl}
\SetMathAlphabet{\mathsfit}{bold}{\encodingdefault}{\sfdefault}{bx}{n}

\def\gC{{\mathcal{C}}}

\DeclareMathOperator*{\argmax}{arg\,max}

\DeclareFontFamily{U}{mathx}{}
\DeclareFontShape{U}{mathx}{m}{n}{<-> mathx10}{}
\DeclareSymbolFont{mathx}{U}{mathx}{m}{n}
\DeclareMathAccent{\widecheck}{0}{mathx}{"71}

\NewDocumentCommand{\intz}{o}{%
  \IfValueTF{#1}{%
    \tilde{z}_{#1}%
  }{%
    \tilde{\rvz}%
  }%
}
\NewDocumentCommand{\intw}{o}{%
  \IfValueTF{#1}{%
    \tilde{w}_{#1}%
  }{%
    \tilde{\rvw}%
  }%
}

\newcommand{\inscope}{\ensuremath{\text{in}}}

\newsavebox{\inputunitbox}
\savebox{\inputunitbox}{%
	\tikz[x=.5em,y=1ex,baseline=0]{%
		\draw[line width=.5pt] plot[domain=-.66:.66] (\x,{1.2*exp(-14*\x*\x)});
		\draw[line width=.5pt] (0,.625ex) circle[radius=1.15ex];
	}%
}
\newcommand{\inputunit}{\usebox\inputunitbox\xspace}

\usepackage{float} %
\usepackage{lipsum}
\usepackage{booktabs} %
\usepackage{graphicx}
\usepackage{xspace}

\usepackage[tableposition=below]{caption}

\usepackage{pifont}%
\newcommand{\Cmark}{{\color{tab10green}\ding{51}}\hspace{0.5em}}%
\newcommand{\Xmark}{{\color{tab10red}\ding{55}}\hspace{0.5em}}%
\newcommand{\XmarkBlack}{\ding{55}\hspace{0.5em}}%

\usepackage{xargs}
\usepackage[colorinlistoftodos,textsize=small]{todonotes} %
\newcommandx{\todoc}[2][1=]{{\todo[linecolor=tab10orange,backgroundcolor=tab10orange!25,bordercolor=tab10orange!35,#1]{
      \scriptsize TODO: #2}}}
\newcommandx{\unsure}[2][1=]{{\todo[linecolor=tab10olive,backgroundcolor=tab10olive!25,bordercolor=tab10olive!35,#1]{
      \scriptsize UNSURE: #2}}}
\newcommandx{\change}[2][1=]{{\todo[linecolor=tab10blue,backgroundcolor=tab10blue!25,bordercolor=tab10blue!35,#1]{
      \scriptsize CHANGE: #2}}}
\newcommandx{\info}[2][1=]{{\todo[linecolor=tab10green,backgroundcolor=tab10green!25,bordercolor=tab10green!35,#1]{
      \scriptsize INFO: #2}}}
\newcommandx{\improvement}[2][1=]{{\todo[linecolor=tab10purple,backgroundcolor=tab10purple!25,bordercolor=tab10purple!35,#1]{
      \scriptsize IMPROVEMENT: #2}}}
\newcommandx{\thiswillnotshow}[2][1=]{{\todo[disable,#1]{\tiny THIS WILL NOT SHOW:
      #2}}}

\usepackage{bm}

\usepackage[utf8]{inputenc}
\usepackage[T1]{fontenc}    %
\usepackage{hyperref}       %
\usepackage{url}            %
\usepackage{booktabs}       %
\usepackage{amsfonts}       %
\usepackage{nicefrac}       %
\usepackage{microtype}      %
\usepackage{xcolor}         %
\usepackage{multicol}
\usepackage{multirow}
\usepackage{array}
\usepackage{amsmath, amsthm, amssymb, amsfonts}
\usepackage{mathtools}
\usepackage{bm}
\usepackage{enumitem}
\usepackage{float}
\usepackage{tcolorbox}
\usepackage{tabularx} %
\usepackage{soul}
\usepackage{capt-of}
\usepackage{doi}
\usepackage{algorithm}
\usepackage{xspace}
\usepackage{wrapfig,graphbox}
\usepackage{relsize}
\usepackage{caption}
\usepackage{subcaption}
\usepackage{xstring}
\usepackage{ifthen}
\usepackage{svg}

\usepackage{algorithm}
\usepackage[noend]{algpseudocode}

\usepackage{amsmath,amssymb}

\usepackage[capitalise,nameinlink]{cleveref}
\crefname{appfig}{Appendix Figure}{Appendix Figures}
\crefname{apptab}{Appendix Table}{Appendix Tables}
\crefname{appalg}{Appendix Algorithm}{Appendix Algorithms}

\usetikzlibrary{colorbrewer} \usetikzlibrary{positioning}
\usetikzlibrary{shapes.geometric} \usetikzlibrary{arrows}
\usetikzlibrary{arrows.meta} \usetikzlibrary{patterns,decorations.pathreplacing,decorations.markings}
\usetikzlibrary{fit} \usetikzlibrary{backgrounds}
\usetikzlibrary{shapes,backgrounds,calc} \usetikzlibrary{shadows}
\usetikzlibrary{patterns} \definecolor{mygray}{RGB}{190,190,190}

\newcommand{\node}{\mathsf{N}}

\usepackage{xparse}
\usepackage{xstring}

\NewDocumentCommand{\fn}{m m O{0}}{%
  \IfEqCase{#3}{%
    {0}{#1\! \left( #2 \right)}%
    {1}{#1 \bigl( #2 \bigr)}%
    {2}{#1 \Bigl( #2 \Bigr)}%
    {3}{#1 \biggl( #2 \biggr)}%
    {4}{#1 \Biggl( #2 \Biggr)}%
  }%
}

\newcommand{\cbar}{\,|\,}
\newcommand{\ccbar}{\,||\,}
\newcommand{\X}{\mathbf{X}}
\newcommand{\V}{\mathbf{V}}

\newcommand{\Z}{\mathbf{Z}}
\newcommand{\x}{\mathbf{x}}

\newcommand{\z}{\mathbf{z}}

\newlength{\nodethickness}
\newlength{\nodedist}
\newlength{\prodnodedist}
\newlength{\leafnodedist}
\definecolor{tab10green}{HTML}{2CA02C}
\definecolor{tab10blue}{HTML}{1f77b4}
\definecolor{tab10red}{HTML}{d62728}

\definecolor{tab10blue}{RGB}{31,119,180}
\definecolor{tab10orange}{RGB}{255,127,14}
\definecolor{tab10green}{RGB}{44,160,44}
\definecolor{tab10red}{RGB}{214,39,40}
\definecolor{tab10purple}{RGB}{148,103,189}
\definecolor{tab10brown}{RGB}{140,86,75}
\definecolor{tab10pink}{RGB}{227,119,194}
\definecolor{tab10gray}{RGB}{127,127,127}
\definecolor{tab10olive}{RGB}{188,189,34}
\definecolor{tab10cyan}{RGB}{23,190,207}

\newcommand{\citepacronym}[2]{(#1; \citealp{#2})}

\newcommand{\res}[2]{$#1$ {\tiny $\pm #2$}}
\newcommand{\resB}[2]{$\bm{#1}$ {\tiny $\bm{\pm #2}$}}

\newcommand{\pc}{$_{\text{pc}}$}
\newcommand{\header}[1]{\multicolumn{1}{c}{#1}}

\theoremstyle{definition}

\newif\ifshowbooktabrules \showbooktabrulesfalse %

\newcommand{\maybetoprule}{\ifshowbooktabrules\toprule\fi}
\newcommand{\maybebottomrule}{\ifshowbooktabrules\bottomrule\fi}
\newcommand{\tablesize}{\normalsize}

\newcommand{\method}{APC\xspace}
\newcommand{\methods}{APCs\xspace}

\definecolor{avorange}{RGB}{215,102,4}  %
\newcommand{\av}[2][]{%
  \ifthenelse{\equal{#1}{q}}%
    {{\color{avorange}\textit{\textbf{[AV] #2}}}}%
    {{\color{avorange}\textbf{[AV] #2}}}%
}

\newcommand{\sbr}[2][]{%
  \ifthenelse{\equal{#1}{q}}%
    {{\color{tab10blue}\textit{\textbf{[SB] #2}}}}%
      {{\color{tab10blue}\textbf{[SB] #2}}}%
}

\hypersetup{ colorlinks, linkcolor={red!50!black}, citecolor=tab10blue, urlcolor={blue!80!black} }

\begin{document}

\maketitle

\begin{abstract}
Probabilistic circuits (PCs) are powerful probabilistic models that enable exact and tractable inference, making them highly suitable for probabilistic reasoning and inference tasks.
While dominant in neural networks, representation learning with PCs remains underexplored, with prior approaches relying on external neural embeddings or activation-based encodings.
To address this gap, we introduce autoencoding probabilistic circuits (\methods), a novel framework leveraging the tractability of PCs to model probabilistic embeddings explicitly.
\methods extend PCs by jointly modeling data and embeddings, obtaining embedding representations through tractable probabilistic inference.
The PC encoder allows the framework to natively handle arbitrary missing data and is seamlessly integrated with a neural decoder in a hybrid, end-to-end trainable architecture enabled by differentiable sampling.
Our empirical evaluation demonstrates that \methods outperform existing PC-based autoencoding methods in reconstruction quality, generate embeddings competitive with, and exhibit superior robustness in handling missing data compared to neural autoencoders.
These results highlight \methods as a powerful and flexible representation learning method that exploits the probabilistic inference capabilities of PCs, showing promising directions for robust inference, out-of-distribution detection, and knowledge distillation.

\end{abstract}

\section{Introduction}
\label{sec:intro}

\begin{wrapfigure}{R}{0.5\textwidth}
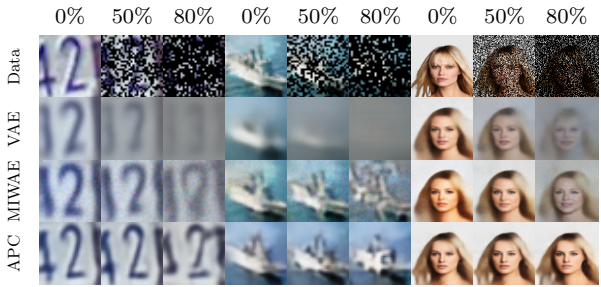

  \vspace{-1.0em}
  \tablesize \newcommand{\rbox}[1]{\rotatebox{90}{\small #1}}
\centering

\setlength{\tabcolsep}{0.0pt}  %
\renewcommand{\arraystretch}{0.0}  %

\newcommand{\imwidth}{1.0\linewidth}
\newcommand{\colwidth}{0.1\linewidth}
\newcommand{\imda}[5]{\includegraphics[width=\imwidth]{./res/rec-missing-images/#1/#2/#3/data/#4_#5.png }}
\newcommand{\im}[5]{\includegraphics[width=\imwidth]{./res/rec-missing-images/#1/#2/#3/recs/#4_#5.png }}
\newcommand{\mc}[1]{\multicolumn{1}{c}{#1}}
\newcommand{\svhnIdx}{2}
\newcommand{\mnistIdx}{8}
\newcommand{\cifarIdx}{28}
\newcommand{\celebaIdx}{10}
\newcommand{\imagenetIdx}{0}
\newcommand{\p}[1]{\mc{\footnotesize #1}}
\newcommand{\rowlabel}[1]{\rbox{\scalebox{0.7}{#1}}}

\begin{tabular}{m{0.05\linewidth} m{\colwidth} m{\colwidth} m{\colwidth} m{\colwidth} m{\colwidth} m{\colwidth} m{\colwidth} m{\colwidth} m{\colwidth} m{\colwidth} m{\colwidth} m{\colwidth}}
& \p{0\%}                            & \p{50\%}                           & \p{80\%}
& \p{0\%}                            & \p{50\%}                           & \p{80\%}
& \p{0\%}                            & \p{50\%}                           & \p{80\%} \\[0.5em]
\rowlabel{Data}
& \imda{svhn-extra}{VAE}{mcar}{\svhnIdx}{0}                 & \imda{svhn-extra}{VAE}{mcar}{\svhnIdx}{50}                  & \imda{svhn-extra}{VAE}{mcar}{\svhnIdx}{80}
& \imda{cifar}{VAE}{mcar}{\cifarIdx}{0}                    & \imda{cifar}{VAE}{mcar}{\cifarIdx}{50}                     & \imda{cifar}{VAE}{mcar}{\cifarIdx}{80}
& \imda{celeba}{VAE}{mcar}{\celebaIdx}{0}                  & \imda{celeba}{VAE}{mcar}{\celebaIdx}{50}                   & \imda{celeba}{VAE}{mcar}{\celebaIdx}{80} \\
\rowlabel{VAE}
& \im{svhn-extra}{VAE}{mcar}{\svhnIdx}{0}                & \im{svhn-extra}{VAE}{mcar}{\svhnIdx}{50}                 & \im{svhn-extra}{VAE}{mcar}{\svhnIdx}{80}
& \im{cifar}{VAE}{mcar}{\cifarIdx}{0}                   & \im{cifar}{VAE}{mcar}{\cifarIdx}{50}                    & \im{cifar}{VAE}{mcar}{\cifarIdx}{80}
& \im{celeba}{VAE}{mcar}{\celebaIdx}{0}                 & \im{celeba}{VAE}{mcar}{\celebaIdx}{50}                  & \im{celeba}{VAE}{mcar}{\celebaIdx}{80} \\
\rowlabel{MIWAE}
& \im{svhn-extra}{MIWAE}{mcar}{\svhnIdx}{0}                 & \im{svhn-extra}{MIWAE}{mcar}{\svhnIdx}{50}                  & \im{svhn-extra}{MIWAE}{mcar}{\svhnIdx}{80}
& \im{cifar}{MIWAE}{mcar}{\cifarIdx}{0}                    & \im{cifar}{MIWAE}{mcar}{\cifarIdx}{50}                     & \im{cifar}{MIWAE}{mcar}{\cifarIdx}{80}
& \im{celeba}{MIWAE}{mcar}{\celebaIdx}{0}                  & \im{celeba}{MIWAE}{mcar}{\celebaIdx}{50}                   & \im{celeba}{MIWAE}{mcar}{\celebaIdx}{80}\\
\rowlabel{APC}
& \im{svhn-extra}{APC_dec_nn}{mcar}{\svhnIdx}{0}         & \im{svhn-extra}{APC_dec_nn}{mcar}{\svhnIdx}{50}          & \im{svhn-extra}{APC_dec_nn}{mcar}{\svhnIdx}{80}
& \im{cifar}{APC_dec_nn}{mcar}{\cifarIdx}{0}            & \im{cifar}{APC_dec_nn}{mcar}{\cifarIdx}{50}             & \im{cifar}{APC_dec_nn}{mcar}{\cifarIdx}{80}
& \im{celeba}{APC_dec_nn}{mcar}{\celebaIdx}{0}          & \im{celeba}{APC_dec_nn}{mcar}{\celebaIdx}{50}           & \im{celeba}{APC_dec_nn}{mcar}{\celebaIdx}{80}
\end{tabular}

  \caption{%
    \textbf{\methods outperform VAE-based models in reconstruction quality under missing data.} For experiment settings, see
    \cref{sec:eval:reconstruction}.
  }
  \label{fig:rec-examples-first-page}
  \vspace{-1.0em}
\end{wrapfigure}

The ability to learn compact and expressive data representations is a fundamental aspect of modern machine learning.
Representation learning involves automatically discovering and encoding informative, low-dimensional features or
embeddings from raw data. These effectively capture underlying factors of variation to facilitate downstream tasks such
as classification, generation, or retrieval. Autoencoders have played a pivotal role in this domain by enabling the
discovery of low-dimensional embeddings that capture latent data features
\citep{Hinton2006ReducingTD,rifai2011learning}. Their impact extends across diverse applications, such as
self-supervised learning paradigms like joint embedding predictive architectures \citep{Assran2023SelfSupervisedLF},
natural language processing, where token embeddings are crucial for large language models \citep{Vaswani2017AttentionIA,
  Devlin2019BERTPO}, and computer vision, where embeddings facilitate efficient image retrieval in large databases
\citep{imagenet}. These advancements predominantly rely on neural network-based approaches, highlighting the critical
role of embeddings in contemporary representation learning. Neural networks have been extensively studied in the field
of representation learning, leading to their probabilistic extension, variational autoencoders~\citepacronym{VAEs}
{kingma2014auto,rezende2014stochasticba}, later improved and extended along several dimensions such as the introduction
of vector quantization~\citepacronym{VQ-VAE}{vanDenOord2017,razavi2019generating}, and
hierarchical~\citep{vahdat2020nvae,lievin2019towards} and deterministic~\citep{ghosh2020rae} formulations.

In recent years, a special kind of neural network has emerged for tractable probabilistic modeling: probabilistic
circuits~\citepacronym{PCs}{choi2020pc}. The computational graphs of PCs are constrained in the way they are formed, thus trading off
expressiveness for tractable inference \citep{vergari2019tractable}. For example, given certain structural properties that are easy to enforce, PCs
can allow the tractable marginalization of every subset of the input features in just a single feedforward pass of the
computational graph. These models provide a unified framework for tractable probabilistic
models~\citep{vergari2021compositional} and enable exact inference, addressing tractability challenges in deep
generative models~\citep{darwiche2003differential, poon2011spn, kisa2014probabilistic}. They have been successfully
applied in diverse areas, including lossless compression~\citep{liu2022lossless,severo2025lossless}, biomedical
modeling~\citep{dang2022tractable,mathur2023knowledge,mathur2024knowledge}, neuro-symbolic AI~\citep{ahmed2022semantic,
  loconte2023turn, karanam2025unified,kurscheid2025tprobabilistic}, multi-modal fusion \citep{sidheekh2025credibilityaware}, graph representation and
learning~\citep{errica2023gspn,zheng2018graph,loconte2023turn}, and constrained text generation~\citep{zhang2023tractable}. Recent
advancements have enhanced PCs in expressivity, learning, and scaling through sparsity-inducing
regularization~\citep{dang2022sparse}, structural improvements through subtractive mixtures and squared
circuits~\citep{loconte2024subtractive,loconte2024sum,wang2024relationship}, as well as hybrid approaches like
HyperSPNs~\citep{shih2021hyperspn}, neural conditioning in conditional sum-product networks~\citep{shao2022cspn} and
probabilistic neural circuits~\citep{martires2024pnc}, and latent variable distillation~\citep{liu2023scaling,
  liu2023understanding}. However, in contrast to neural networks, the use of PCs for representation learning has
remained largely unexplored.

All the above works focus on learning or querying PCs as black-box probability models, i.e., functions that
receive some input configuration and output a scalar probabilistic quantity of interest: a marginal or conditional
probability or the value of an expectation. However, since PCs are neural networks, one could use them to extract
representations and use these for downstream predictive tasks.
To date, research into representation learning with PCs has been limited, with two notable exceptions.
The first is the investigation of internal PC representations by \cite{vergari_2019}. The second is
sum-product autoencoding~\citepacronym{SPAE}{vergari2018spae}, which defines an autoencoding scheme using these
representations. However, these prior approaches face severe limitations: it is often unclear how to extract general
representations decoupled from the specific circuit structure, and the autoencoding scheme cannot be trained end-to-end.
This inherently prevents seamless integration with modern neural architectures.

\begin{figure}
  \centering
  \input{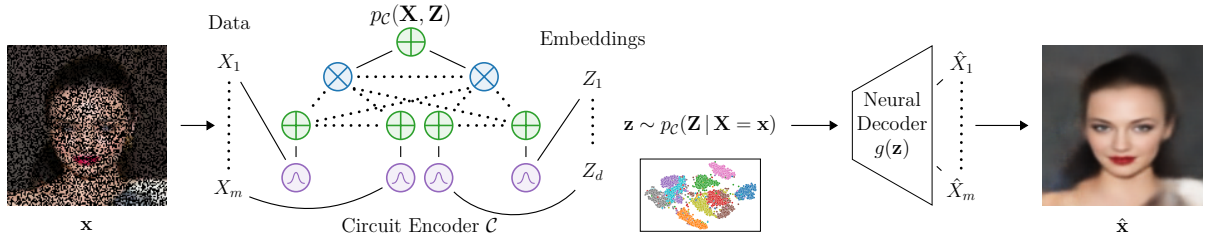}
  \caption{\textbf{Autoencoding probabilistic circuits (\methods) architecture}. An input data point, possibly with only partial
    information, is probabilistically encoded into a compact embedding space $\Z$ using a PC
    $\fn{p_{\gC}}{\X, \Z}$. The embedding $\z$ is sampled from the data-conditional distribution
    $\fn{p_{\gC}}{\Z \cbar \X}$. A neural decoder then reconstructs the complete data point $\hat{\X}$ from the
    embedding. The tractable nature of the PC encoder allows the intrinsic handling of missing data
    without the need for imputation.}
  \label{fig:arch}
\end{figure}

We present \emph{autoencoding probabilistic circuits} (\methods), a novel framework that introduces
end-to-end learning of representations in an autoencoding fashion using PCs. \methods leverage the
probabilistic capabilities of PCs by employing them as a tractable encoder, generating \textit{explicit}
probabilistic embeddings $\Z$. Critically, embeddings in \methods are obtained through tractable conditional state inference
(i.e. sampling or most probable explanation) from the joint distribution $\fn{p_{\gC}}{\X, \Z}$ over data $\X$ and
embedding random variables $\Z$ explicitly modeled by the PC: $\z \sim \fn{p_{\gC}}{\Z \cbar \X_{o} = \x_{o}}$, where $\X_{o} \subseteq \X$ can be partially observed, natively handling missing data.
In stark contrast, this is tough to do with fully neural-based autoencoders, where people often resort to unprincipled heuristics or computationally intensive imputation methods~\citep{nazabal2020handling,mattei2019miwae,simkus2023conditional,simkus2024improving} to deal with incomplete inputs (see \cref{sec:related-work} and \cref{fig:rec-examples-first-page}). \methods thus offer an elegant and principled solution by leveraging their ability to tractably model arbitrary marginal input distributions.
\methods then bridge circuit and neural architectures by allowing for an arbitrary neural network decoder, extending the frameworks capacity. \cref{fig:arch} illustrates the APC pipeline.

In summary, our key contributions are: (a) autoencoding probabilistic circuits (\methods), a novel framework for
tractable representation learning with explicit embeddings modeled with PCs; (b) integrating a hybrid architecture that improves
the learning of the encoder circuit, combining the strengths of tractable probabilistic and neural methods;
(c) enabling end-to-end training between the encoder circuit and neural decoder by an improved differentiable sampling procedure for PCs, leveraging novel advances in gradient estimation; and (d) extensively
demonstrating \methods' ability to reconstruct inputs, produce useful embeddings for downstream tasks, and
knowledge distillation without access to original training data,
all in the presence of missing inputs, a scenario that neural autoencoders cannot natively
handle.

\section{Background}
\label{sec:background}
To properly introduce \methods later in \cref{sec:apc}, we need to briefly provide the necessary background about the mechanism of autoencoding and the formal definitions of PCs.

\paragraph{Autoencoding.}
The autoencoding mechanism, fundamental to representation learning and central to \methods, is a form of self-supervised learning~\citep{Hinton2006ReducingTD}. 
An autoencoder consists of two primary components: an \emph{encoder} and a \emph{decoder}. The encoder, $f: \X \mapsto \Z$, maps an input data point $\x \in \X$ to a typically lower-dimensional ($|\Z| \ll |\X|$) latent representation, also called \emph{embedding}, $\z \in \Z$. The decoder, $g: \Z \mapsto \X$, then attempts to reconstruct the original input from this embedding, producing $\hat{\x} = g(\z) = g(f(\x))$. The model is trained by minimizing a possibly regularized reconstruction loss, $\mathcal{L}(\x, \hat{\x})$, which measures the dissimilarity between the original input and its reconstruction. This process encourages the latent space $\Z$ to capture salient representations and variations within the data \citep{vincent2008extracting,rifai2011learning}.
Several variants of autoencoders (AEs) have been proposed in the literature, sometimes tailored for specific data modality such as text \citep{sutskever2014sequence}, images \citep{xu2014deep} and graph data such as molecules \citep{kusner2017grammar}. 

\textit{Variational autoencoders} \citepacronym{VAEs}{rezende2014stochasticba,kingma2014auto} have been introduced as probabilistic AEs whose encoders and decoders realize \textit{stochastic} maps.
Specifically, the encoder in a VAE draws embedding samples from a conditional distribution $\z\sim p(\Z\mid \x)$, which is commonly realized as a simple distribution (most of the time, a diagonal Gaussian) parameterized by a neural backbone that receives $\x$ as input.
 Analogously, its decoder, draws reconstructions from a distribution $\hat{\x}\sim p(\X\mid \z)$ conditioned on the latent code $\z$.
Again, the distribution is assumed to have a simple form parameterized by a neural network \citep{ghosh2020rae}.
Neural parameterizations for both the encoder and decoder increase the expressiveness of these models. However, this increased expressiveness comes at the cost of reduced flexibility, especially when the models are required to perform advanced probabilistic inference tasks, such as computing marginals.

A classical example is that of missing input values \citep{colliervaes,williams2018autoencoders}.
That is, when only some variables $\X_{o}\subset\X$ are observed, a neural parameterization of the encoder entails that computing $p(\Z\mid \x_{o})$ is intractable, as it would require computing the marginal distribution $p(\X_{o})=\int p(\X_{o}, \X_{m})d\X_{m}$, where $\X_{m}=\X\setminus\X_{o}$ are the missing input features.
As such, dealing with missing values for both AEs and VAEs is a challenging task, for which several heuristics and approximations have been proposed \citep{nazabal2020handling,mattei2019miwae,simkus2023conditional,simkus2024improving}, see \cref{sec:related-work} for a detailed discussion.

\paragraph{Probabilistic circuits.}
A \emph{probabilistic circuit} $\gC$, is a computational graph representing a function
$\gC(\rmX)$ over its inputs $\rmX$ (this is just a placeholder which can e.g. include data \textit{and} embedding variables), where $\gC(\rvx) \geq 0$ for any input $\x$, making $\gC(\X)$ an unnormalized distribution.
 The computational graph, parameterized by $\theta$, consists of three types of computational units: \emph{input} \inputunit, \emph{sum} $\bigoplus$, and \emph{product} $\bigotimes$.
 An input unit $c$ encapsulates a parameterized function $f_c(\rmX_c)$, with $\rmX_c \subseteq \rmX$ denoting its scope.
 The scopes for product and sum units, $\X_c$, are the union of their input units' scopes.
 Specifically, a product unit computes the product of its inputs $\prod_{d \in \inscope(c)} d(\X_d)$, and a sum unit computes the weighted sum $\sum_{d \in \inscope(c)} \theta_{d}^{c} \, d(\X_d)$.
 We assume sum unit weights to be normalized, i.e. $\sum_{d\in \inscope(c)} \theta_{d}^{c} = 1$.

PCs can tractably (in time $\fn{\mathcal{O}}{|\gC|}$) and exactly compute evidence, marginal, conditional, and moment queries, which are essential to our proposed circuit-based encoding procedure, if it fulfills two structural properties: \emph{smoothness} and \emph{decomposability} \citep{darwiche_2002,choi2020pc,vergari2021compositional}.

\begin{figure}[!t]
  \centering
  \setlength{\minnodesize}{0.7cm}

\setlength{\shortenlength}{0.15cm}

\setlength{\nodedist}{1.75cm}

\setlength{\nodethickness}{1pt} %

\newsavebox{\inputunitboxnocirc}
\savebox{\inputunitboxnocirc}{%
	\tikz[x=.75em,y=1.5ex,baseline=1.ex]{%
		\draw[line width=\nodethickness, tab10purple] plot[domain=-.66:.66] (\x,{1.2*exp(-14*\x*\x)});
	}%
}
\newcommand{\inputunitnocirc}{\usebox\inputunitboxnocirc\xspace}

\centering
\scalebox{0.75}{
  \begin{tikzpicture}[
        node distance=1cm,
        font=\Large,
        prod/.style={
            circle, draw=tab10blue, fill=tab10blue!10, line width=0.75pt, minimum size=7mm,
            path picture={
                \draw[tab10blue, line width=0.75pt]
                (path picture bounding box.south east) -- (path picture bounding box.north west)
                (path picture bounding box.south west) -- (path picture bounding box.north east);
            }
        },
        sum/.style={
            circle, draw=tab10green, fill=tab10green!10, line width=0.75pt, minimum size=7mm,
            path picture={
                \draw[tab10green, line width=0.75pt]
                (path picture bounding box.south) -- (path picture bounding box.north)
                (path picture bounding box.west) -- (path picture bounding box.east);
            }
        },
        leaf/.style={circle, draw, line width=0.5\nodethickness, minimum
      size=\minnodesize},
        pcedge/.style={
            thick,
            shorten >=1.5mm,
            shorten <=1.5mm,
            {Triangle[scale=1.25]}-,
            font=\fontsize{16}{0}\selectfont
          },
        sampleedge/.style={
            thick,
            shorten >=1.5mm,
            shorten <=4.0mm,
            tab10orange,
            line width=1pt,
            {-Triangle[scale=1.25]},
            font=\fontsize{16}{0}\selectfont
          },
          gauss/.style={
            circle,
            tab10purple,
            draw,
            inner sep=0,outer sep=0,
            fill=tab10purple!10,
            line width=0.75\nodethickness,
            minimum size=\minnodesize,
            },
    ]
        \node[sum, label={[yshift=0.5ex]above:$p_\gC(\X, Z_1, Z_2)$}] (root) {};

        \node[prod] (p1_left) [below left=0.8cm and 3cm of root] {};
        \node[prod] (p1_right) [below right=0.8cm and 3cm of root] {};

        \node[gauss] (l_z1) [below left=1.25cm and 0.75cm of p1_left, label={[yshift=-0.5ex]below:{$Z_1$}}] {\inputunitnocirc};
        \node[sum]  (s_left) [below right=1.25cm and 0.75cm of p1_left] {};
        \node[gauss] (l_z2_a) [below left=1.25cm and 0.75cm of p1_right, label={[yshift=-0.5ex]below:{$Z_2$}}] {\inputunitnocirc};
        \node[sum]  (s_right) [below right=1.25cm and 0.75cm of p1_right] {};

        \node[prod] (p2_left_1) [below left=1.25cm and 0.75cm of s_left] {};
        \node[prod] (p2_left_2) [below right=1.25cm and 0.75cm of s_left] {};
        \node[prod] (p2_right_1) [below left=1.25cm and 0.75cm of s_right] {};
        \node[prod] (p2_right_2) [below right=1.25cm and 0.75cm of s_right] {};

        \node[gauss] (l_X1) [below left=1.25cm and 0.3cm of p2_left_1, label={[yshift=-0.5ex]below:{$\X$}}] {\inputunitnocirc};
        \node[gauss] (l_X2) [below right=1.25cm and 0.3cm of p2_left_2, label={[yshift=-0.5ex]below:{$\X$}}] {\inputunitnocirc};
        \node[gauss] (l_z2_b) [at={($(l_X1)!0.5!(l_X2)$ |- l_X1)}, label={[yshift=-0.5ex]below:{$Z_2$}}] {\inputunitnocirc};
        \node[gauss] (l_z1_a) [below left=1.25cm and 0.3cm of p2_right_1, label={[yshift=-0.5ex]below:{$Z_1$}}] {\inputunitnocirc};
        \node[gauss] (l_z1_b) [below right=1.25cm and 0.3cm of p2_right_2, label={[yshift=-0.5ex]below:{$Z_1$}}] {\inputunitnocirc};
        \node[gauss] (l_X3) [at={($(l_z1_a)!0.5!(l_z1_b)$ |- l_X1)}, label={[yshift=-0.5ex]below:{$\X$}}] {\inputunitnocirc};

        \draw[pcedge, shorten <=5mm, shorten >=8mm] ($(root)!0.1cm!90:(p1_left)$) -- ($(p1_left)!0.1cm!-90:(root)$);
        \draw[sampleedge, shorten <=8mm, shorten >=5mm] ($(root)!0.1cm!-90:(p1_left)$) -- ($(p1_left)!0.1cm!90:(root)$);
        \draw[pcedge] (root) -- (p1_right);

        \draw[pcedge, shorten <=5mm, shorten >=8mm] ($(p1_left)!0.1cm!90:(l_z1)$) -- ($(l_z1)!0.1cm!-90:(p1_left)$);
        \draw[sampleedge, shorten <=8mm, shorten >=5mm] ($(p1_left)!0.1cm!-90:(l_z1)$) -- ($(l_z1)!0.1cm!90:(p1_left)$);

        \draw[pcedge, shorten <=5mm, shorten >=8mm] ($(p1_left)!0.1cm!90:(s_left)$) -- ($(s_left)!0.1cm!-90:(p1_left)$);
        \draw[sampleedge, shorten <=8mm, shorten >=5mm] ($(p1_left)!0.1cm!-90:(s_left)$) -- ($(s_left)!0.1cm!90:(p1_left)$);

        \draw[pcedge] (p1_right) -- (l_z2_a);
        \draw[pcedge] (p1_right) -- (s_right);

        \draw[pcedge] (s_left) -- (p2_left_1);

        \draw[pcedge, shorten <=5mm, shorten >=8mm] ($(s_left)!0.1cm!90:(p2_left_2)$) -- ($(p2_left_2)!0.1cm!-90:(s_left)$);
        \draw[sampleedge, shorten <=8mm, shorten >=5mm] ($(s_left)!0.1cm!-90:(p2_left_2)$) -- ($(p2_left_2)!0.1cm!90:(s_left)$);

        \draw[pcedge] (s_right) -- (p2_right_1);
        \draw[pcedge] (s_right) -- (p2_right_2);

        \draw[pcedge] (p2_left_1) -- (l_X1);
        \draw[pcedge] (p2_left_1) -- (l_z2_b);

        \draw[pcedge, shorten <=5mm, shorten >=8mm] ($(p2_left_2)!0.1cm!90:(l_z2_b)$) -- ($(l_z2_b)!0.1cm!-90:(p2_left_2)$);
        \draw[sampleedge, shorten <=8mm, shorten >=5mm] ($(p2_left_2)!0.1cm!-90:(l_z2_b)$) -- ($(l_z2_b)!0.1cm!90:(p2_left_2)$);

        \draw[pcedge, shorten <=5mm, shorten >=8mm] ($(p2_left_2)!0.1cm!90:(l_X2)$) -- ($(l_X2)!0.1cm!-90:(p2_left_2)$);
        \draw[sampleedge, shorten <=8mm, shorten >=5mm] ($(p2_left_2)!0.1cm!-90:(l_X2)$) -- ($(l_X2)!0.1cm!90:(p2_left_2)$);

        \draw[pcedge] (p2_right_1) -- (l_z1_a);
        \draw[pcedge] (p2_right_1) -- (l_X3);
        \draw[pcedge] (p2_right_2) -- (l_X3);
        \draw[pcedge] (p2_right_2) -- (l_z1_b);

    \end{tikzpicture}
}
  \caption{A PC over a set of data random variables $\X$ and two embedding variable $Z_1$ and $Z_2$. The black connections illustrate the flow of computation during a forward pass for inference. In contrast, the orange subgraph highlights a sampling-induced tree.}
  \label{fig:pc}
\end{figure}

\paragraph{Structural properties for tractable marginals and sampling.}
A sum unit $n$ is \emph{smooth}, if all input units $c \in \inscope(n)$ have the same scope, i.e., $\X_{c} = \X_{n}$.
A product unit $n$ is \emph{decomposable}, if all input units $c_{i},c_{j} \in \inscope(n)$ have pairwirse disjoint
scopes, i.e. $\X_{c_{i}} \cap \X_{c_{j}} = \emptyset$ for $c_{i} \neq c_{j}$.
A PC is smooth and decomposable, if all of its sum units are smooth and all of its product units are decomposable.
\cref{fig:pc} presents an example of a smooth and decomposable PC.
Smooth and decomposable PCs can be interpreted as hierarchical latent variable models, where a discrete latent variable $H_c$ is associated to each sum unit $c$.
Such a latent variable  has as many states as the number of input connections of its corresponding sum unit \citep{peharz2017latent}.
Under this light, the weights $\theta_{d}^{c}$ of a sum unit $c$ represent the probabilities of selecting a specific mixture component, i.e., one of its input connections.
This latent variable interpretation enables exact and efficient
ancestral sampling from the distribution $\fn{p_{\gC}}{\X, \mathbf{H}}$, where $\mathbf{H}$ denote all the discrete
latent variables associated to all sum units in the PC.\footnote{Usually, samples from $\X$ are retained and those from
  $\mathbf{H}$ discarded \citep{vergari2018spae}.}

Sampling proceeds as follows.
From the output unit of a PC, one traverses the computational graph backwards, outputs before inputs, until the input distribution units are reached.
This process constructs a \textit{sampling-induced tree} in the PC.
\cref{fig:pc} illustrates this process.
For a product unit, all input connections are followed, while for a sum unit, a single connection is sampled with probability proportional to its corresponding weight.
Once an input distribution is reached, one can sample a feature corresponding to its scope according to the parametric distribution encoded in the unit.
A complete sample ($\mathbf{x}, \mathbf{h}$) is retrieved by concatenating all features sampled from input units, and all latent variables $\mathbf{H}$ encountered in the backward traversal.
However, this sampling procedure is inherently non-differentiable due to the discrete choices made at sum units.
While methods to make PC sampling differentiable have been explored \citep{lang2022diff-sampling-spns}, their practical limitations, such as the poor sample quality and the gradient instability, motivate the need for more robust ways to sample PCs and integrate them in larger neural pipelines.

\section{Autoencoding Probabilistic Circuits}
\label{sec:apc}

We now introduce \emph{autoencoding probabilistic circuits} (\methods), a framework to learn end-to-end representations in an autoencoding fashion using PCs, leveraging their tractable inference capabilities.
Prior attempts to extract representations from PCs have explored two main avenues, notably within sum-product autoencoding \citepacronym{SPAE}{vergari2018spae}.
First, as PCs can be viewed as a specialized type of neural network, representations (termed ACT embeddings in SPAE) have been derived from the activations of selected circuit units.
However, unlike typical neural networks, PCs are sparse computational graphs that often lack a clear, inherent notion of a ``bottleneck'' layer, making it challenging to identify and select which units provide compact and informative representations.
Second, given their interpretation as hierarchical latent variable models with internal discrete latent variables (as discussed in \cref{sec:background}), representations (termed CAT embeddings in SPAE) have been formed by inferring the most probable assignments to these internal latent variables.
Yet, modern PCs can involve an enormous number of such internal latent variables, making it non-trivial to select a concise and globally meaningful subset for representation.
Crucially, these prior approaches primarily derive embeddings post-hoc, without explicitly integrating the learning of embeddings into the training process end-to-end.

In contrast, \methods introduce a fundamentally different approach by treating embeddings as first-class random
variables within the PC itself.
Instead of learning a separate conditional encoder as in VAEs (i.e., an inference model $p(\Z \cbar \x)$), we design a single PC, $\gC$, that explicitly models the \emph{joint probability distribution} $p_\gC(\X, \Z)$ over both data $\X$ and embeddings $\Z$.
This is achieved by representing the embedding variables with their own dedicated, parametric input units within the circuit structure.
The primary advantage of this formulation is that the encoder \emph{is} the joint distribution itself, which remains fully tractable.
Consequently, we can obtain embedding representations through exact probabilistic inference, such as by sampling from
the true conditional posterior $\z \sim p_\gC(\Z | \x)$ \textit{even for partially-observed inputs $\x$} or performing
most probable explanation (MPE) inference.
This provides a more principled and direct foundation for probabilistic representation learning.
Note that these \textit{explicit} input latent embeddings $\Z$ are different from the \emph{internal} discrete latent variables $\mathbf{H}$ of the PC (see previous section). While the latter are discrete latent variables, the former can be continuous.

\subsection{Encoding with Tractable Joint Data-Embedding Distributions}
\label{sec:apc:encoder}

As the encoder model, we construct a smooth and decomposable PC $\gC$ that models the joint distribution $\fn{p_{\gC}}{\X, \Z}$ of data and embedding random variables.
The encoding process $f_{\gC}\!: \X \mapsto \Z$ maps a data sample $\x$ to its embedding representation $\z$.
We define the encoding $\fn{f_{\gC}}{\x}$ as performing probabilistic state inference in the PC by conditionally sampling an embedding, given the data $\x$.
In contrast to the VAE-based formulation, when $\gC$ is a smooth and decomposable PC, we can perform arbitrary marginalization and conditioning, allowing us to directly sample latent variables from the data-conditional distribution $p_\gC(\Z \cbar \x)$.
Conditional sampling in PCs consists of two passes through the network.
First, a forward pass of the marginal $\fn{p_{\gC}}{\x}$ for a given data sample $\x$ caches log-likelihoods at each input edge.
Subsequently, a sampling pass with reweighted sum unit weights according to their cached log-likelihoods samples from the on $\X = \x$ conditioned marginal $\fn{p_{\left.\gC\right\rvert_{\X=\x}}}{\Z} = \fn{p_{\gC}}{\Z \cbar \X = \x}$.
We outline the full encoding algorithm in \cref{alg:encoding}.

Moreover, and in stark contrast to neural encoders, a PC-based encoding scheme allows us to infer the full embedding state $\z$ from \textit{partial evidence}, where only a subset $\X_{o} \subseteq \X$ of the random variables are observed, and the rest is missing $ \X_{m} = \X \setminus \X_{o}$.
This is achieved through marginalizing the missing variables $\int \fn{p_{\gC}}{\X_{o}, \X_{m}} \, d\X_{m}$, without the need for any data imputation.
On the other hand, neural networks assume all input values to be given.
Consequently, data with missing observations must be manually preprocessed before being fed to a neural network.
When not natively accounted for in a specific method, this preprocessing typically involves employing various imputation techniques such as mean imputation, multiple imputation~\citep{Rubin1989MultipleIF}, or advanced methods like expectation-maximization algorithms and deep learning approaches~\citep{Yoon2018GAINMD,Gondara2017MIDAMI,Luo2018MultivariateTS}.
As a side effect, for neural networks these imputation methods usually introduce biases or assumptions that can affect model performance.

While not explicitly explored in this work, we emphasize that using a circuit encoder enables leveraging data-specific input or even embedding distributions to model random variables accurately.
This is particularly beneficial in scenarios where we have prior knowledge about the specific data or embedding distributions or when each variable can be effectively represented as a mixture of multiple distributions of different types.

\paragraph{Encoder Structure.}
A key design choice within the \methods framework is the architecture of the PC encoder.
While our approach is agnostic to the specific topology beyond smoothness and decomposability, the chosen structure dictates how the joint distribution $p_\gC(\X, \Z)$ is decomposed and influences the learned representations.
In our experiments, we employ architectures tailored to the data modality.
For tabular data, we use EinsumNetworks~\citep{peharz2020einet} with a RAT~\citep{peharz20a-rat-spn} structure.
Here, inputs are modeled in random order in multiple random permutations without any quantification as to which split between data variables is a ``good fit'', given the data.
Therefore, we simply choose to model embedding variables as input units alongside the data variables at the circuit's input layer.
For image data, we construct a convolutional-style layerwise PC that builds a feature hierarchy through alternating product and sum layers.
Each product layer reduces the height and width of the input by a factor of 2, building the product over all scopes in disjoint neighboring windows, similar to non-overlapping convolution kernels, when the stride is equal to the kernel size.
After each product layer, a sum layer maps all input units of the same scope to a vector of sum units, similar to convolution in- and out-channels in traditional neural convolution layers.
In this structure, each embedding input is randomly coupled with a random data variable unit via a local product unit.
Further implementation details for these architectures are provided in
\cref{app:eval-protocol:model-architectures}.
While these are simplistic choices for how we incorporate the embedding variables in our experiments, one could also move beyond the designs used here by leveraging structure learning algorithms or more principled heuristics.
This could e.g. include the strategic placement of embedding variables at various depths within the circuit, allowing for hierarchical embeddings capable of distinguishing between low-level features and high-level abstractions.

\subsection{Differentiable Tractable Probabilistic Encoding}

\begin{wrapfigure}{r}{0.5\textwidth}
  \vspace{-1.5em}
  \centering
  \includegraphics[width=0.5\textwidth]{ 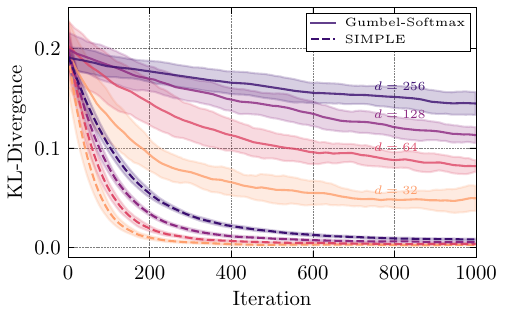 }
  \caption{\textbf{SIMPLE outperforms Gumbel-Softmax in gradient estimation} as measured by the KLD between ground-truth and learned sum units for different input dimensions $d$.}
  \label{fig:simple-vs-gumbel}
  \vspace{-2.0em}
\end{wrapfigure}

As discussed in \cref{sec:background}, while exact (conditional) sampling from a smooth and decomposable PC is possible, it presents a challenge for gradient-based optimization: it is not differentiable.
This non-differentiability arises from the inherent discreteness of the sampling procedure, specifically drawing from categorical distributions defined by sum unit weights and deterministically choosing successor units.

Here, we make use of the concept of differentiable sampling for PCs introduced in \cite{lang2022diff-sampling-spns}.
This allows us to replace the discrete operations present in the sampling procedure of sum units with continuous, and thus differentiable, reparametrizations.
We further improve the method proposed in \cite{lang2022diff-sampling-spns} and replace the Gumbel-Softmax trick with SIMPLE~\citep{ahmed2022simple}.
SIMPLE is a gradient estimator for $k$-subset sampling that uses discrete sampling in the forward pass and exact conditional marginals in the backward pass.
When we set $k=1$, SIMPLE performs sampling from a mixture equal to what we require in sum units and can be used in-place of the Gumbel-Softmax trick.
We outline how SIMPLE is applied to a sum unit in \cref{alg:simple}.
By incorporating SIMPLE into the differentiable sampling procedure for PCs, we can improve gradient estimation and overall performance of the autoencoding pipeline, overcoming the initial difficulties of differentiable sampling in PCs reported in \cite{lang2022diff-sampling-spns}, which we show in \cref{fig:simple-vs-gumbel}.
We designed a synthetic task where the objective is to learn sum unit parameters by differentiably sampling from this sum unit and minimizing the sample mean squared error to samples drawn from a fixed ground-truth categorical distribution.
In this setting, SIMPLE consistently converged faster, achieved higher accuracy (as measured by the Kullback–Leibler divergence between the sum unit and the true categorical distribution), and exhibited lower variance compared to Gumbel-Softmax.
These findings validate SIMPLE for more accurate and stable end-to-end training of \methods.
We provide full details on this synthetic task in \cref{app:simple}.

\subsection{Hybrid Decoding with Neural Networks}
\label{sec:apc:decoder}

In \methods, while the encoding phase leverages the principled and tractable nature of PCs, the decoding process employs a neural network to reconstruct data from the learned embeddings.
This design choice is motivated by the complementary strengths of neural networks, particularly their capacity for modeling complex, non-linear mappings.
Therefore, in \methods, the decoder $g_\theta\!: \Z \mapsto \X$ is implemented as a separate neural network, distinct from the PC encoder $\gC$.
This neural decoder maps the embedding representation $\z$ back to the original data space $\X$, effectively complementing the probabilistic encoding with the modeling capacity and computational efficiency of neural networks during decoding.
The specific architecture of the neural decoder can be tailored to the data type.
For instance, CNNs are well-suited for image data, enabling the incorporation of implicit biases beneficial for capturing spatial relationships not explicitly modeled in the encoding circuit.
Similarly, Transformer-based architectures can be leveraged for sequential or high-dimensional data due to their ability to model long-range dependencies and complex interactions.
Beyond these, for achieving high-fidelity reconstructions and sample quality, advanced generative models such as diffusion models could also be integrated as decoders.

Importantly, this neural decoding strategy remains robust even when dealing with partial evidence.
The tractable encoder $f_{\gC}$'s ability to infer a complete embedding $\z$ from a partially observed input $\x_{o}$ ensures that the neural decoder always receives a full embedding vector.
Consequently, the decoding process is unaffected by missing data in the input.
We empirically explore the benefits of this hybrid approach by comparing \methods with neural decoders to \methods with self-decoding PCs in \cref{sec:eval:ablation}.

\subsection{End-to-End Training of Tractable Encoder and Neural Decoder}
\label{sec:apc:training}

\methods are trained end-to-end using objectives that encourage both accurate reconstruction and meaningful and robust embedding representations.
We derive our training objective from a standard regularized autoencoding mechansim.
The loss function is a sum of three crucial components: a reconstruction term, an embedding prior regularization term, and a joint data-embedding likelihood regularization term:
\begin{equation}
  \label{eq:loss-full}
  \mathcal{L} = \lambda_{\text{REC}} \cdot \mathcal{L}_{\text{REC}} + \lambda_{\text{KLD}} \cdot \mathcal{L}_{\text{KLD}} + \lambda_{\text{NLL}} \cdot \mathcal{L}_{\text{NLL}} \quad .
\end{equation}
The factors $\lambda$ allow us to assign specific weights to different objectives.
Such weighting can, e.g., facilitate disentanglement \citep{higgins2017betavae}.
A detailed investigation of alternative configurations for $\lambda$ is left to future work.
For all experiments in \cref{sec:eval}, we set $\lambda_{\text{REC}} = \lambda_{\text{KLD}} = \lambda_{\text{NLL}} = 1$, as this configuration already worked well in our scenario.
In the following, we will describe each objective in detail.

\paragraph{Reconstruction.}
We use a reconstruction term, $\mathcal{L}_{\text{REC}}$, to ensure that embeddings represent sufficient information to reconstruct the original input data accurately.
This objective, analogous to the one used in conventional autoencoders~\citep{Hinton2006ReducingTD}, encourages the model to learn a meaningful and compact embedding space that captures essential features of the data-generating distribution:
\begin{equation}
  \label{eq:loss-rec}
   \mathcal{L}_{\text{REC}} = - \frac{1}{B} \sum_{i=1}^B \log p_\theta(\x_i \mid \z_i), \quad \text{where } \z_i \sim p_{\gC}(\Z \mid \x_i) \quad ,
\end{equation}
where $B$ is the batch size, $D$ is the dimension of the input data, $\x_i$ is the $i$-th input sample, and
$\hat{\x}_i = \fn{g_{\theta}}{\fn{f_{\gC}}{\x_{i}}}$ is its corresponding reconstruction.

\paragraph{Embedding Regularization.} 
To ensure the learned embeddings are meaningful and to avoid overfitting, we regularize the embedding distribution to a chosen prior.
While simple $L_p$ regularization could be applied to the embedding vectors $\z$, PCs allow for a more principled approach.
Sampling in a PC instantiates the hidden sum unit  variables $\mathbf{h} \in \mathbf{H}$ and induces a product unit tree with independent input units.
This is highlighted as the orange subgraph in \cref{fig:pc}.
Consequently, sampling embeddings from the conditional distribution $\fn{p_{\gC}}{\Z \cbar \x}$, given some data point $\x$, induces a tree $\gC'$ in which embedding variables are mutually conditionally independent, given $\x$ and $\mathbf{h}$:
\begin{equation}
  Z_j \perp Z_k \cbar \X=\x, \mathbf{H} = \mathbf{h}, \quad \forall j, k \in {1, \ldots, |\Z|}, j \neq k \quad .
\end{equation}
This independence allows us to apply statistical distances between the induced embedding distribution and a chosen prior $q$.
Specifically, we use the Kullback-Leibler divergence (KLD):
\begin{equation}
  \label{eq:loss-kld}
  \mathcal{L}_{\text{KLD}} = \sum_{i=1}^{B} \sum_{j=1}^{|\Z|}\fn{\text{KLD}}{\fn{p_{\gC'}}{Z_j \cbar \x_{i}, \mathbf{h}_i} \ccbar \fn{q}{Z_j}}[1]
\end{equation}
where $B$ is the batch size and $\x_i$ represents the $i$-th data sample.
When choosing distributions of particular parametric forms, e.g. the exponential family, we can compute \cref{eq:loss-kld} analytically in closed form. Embedding distributions in $\gC$ are chosen as Gaussian input units in all cases.

\paragraph{Joint Data and Embedding Likelihood.}

A key advantage of the PC encoder is its capacity to model the full joint distribution $p_\gC(\X, \Z)$.
We leverage this by introducing a third loss component that maximizes the joint log-likelihood of the data and their inferred embeddings.
This objective steers the encoder's parameters towards a maximum likelihood estimate (MLE) of the joint distribution, regularizing the reconstruction-focused training.
Our ablation study in \cref{sec:eval:ablation} empirically validates that this term is crucial for achieving high performance.
We formalize this objective as the negative log-likelihood (NLL):
\begin{equation}
  \label{eq:loss-nll}
  \mathcal{L}_{\text{NLL}} = -\frac{1}{B} \sum_{i=1}^{B} \log \fn{p_\gC}{\x_i, \z_i} \quad \text{where } \z_i \sim p_{\gC}(\Z \mid \x_i) \quad .
\end{equation}

\paragraph{Training Via Knowledge Distillation.}
While the \methods framework can be trained from scratch on raw data using the aforementioned objectives, its tractable marginals also enable data-free knowledge distillation from pre-trained generative latent variable models like VAEs.
In this scenario, the \method model acts as a student, iteratively learning to capture the teacher's generative distribution without direct access to the original training data.
The process involves sampling an embedding $\z$ from the \method's prior $\fn{p_{\gC}}{\Z}$, which the teacher VAE's decoder $\fn{g_{\text{VAE}}}{\z}$ uses to generate synthetic data $\hat{\x}_{\text{VAE}}$.
The corresponding latent code $\z_{\text{VAE}}$ is then obtained by encoding $\hat{\x}_{\text{VAE}}$ with the VAE's encoder $\fn{f_{\text{VAE}}}{\hat{\x}_{\text{VAE}}}$.
For training the \method student, these synthetic data $\hat{\x}_{\text{VAE}}$ and teacher latent codes $\z_{\text{VAE}}$ replace the ``ground truth'' data points $\x_i$ and their corresponding latent codes $\z_i$ in the loss terms.
Specifically, we get
\begin{align}
  \mathcal{L}_{\text{KLD,KD}} &= \sum_{i=1}^{B} \sum_{j=1}^{|\Z|}\fn{\text{KLD}}{\fn{p_{\gC'}}{Z_j \cbar \x_{\text{VAE},i}, \mathbf{h}_i} \ccbar \fn{q}{Z_j}}[1] \\
  \mathcal{L}_{\text{NLL,KD}} &= -\frac{1}{B} \sum_{i=1}^{B} \log \fn{p_\gC}{\x_{\text{VAE},i}, \z_{\text{VAE},i}} \\
   \mathcal{L}_{\text{REC,KD}} &= - \frac{1}{B} \sum_{i=1}^B \log p_\theta(\x_{\text{VAE},i} \mid \z_i), \quad \text{where } \z_i \sim p_{\gC}(\Z \mid \x_{\text{VAE},i}) \quad .
\end{align}
A detailed data-free knowledge-distillation algorithm is outlined in \cref{alg:knowledge-distillation}.
While our main focus in \cref{sec:eval} is on training \methods from scratch, we additionally explore the alternative perspective of training via data-free knowledge distillation from pretrained VAEs in \cref{sec:eval:knowledge-distillation}.

\section{Related Work}
\label{sec:related-work}

Having introduced \methods in the previous section, we now contextualize our framework by discussing prior work in representation learning using both PCs and neural networks and prior hybrid approaches between these two classes of models.

\paragraph{Representation Learning with Probabilistic Circuits.}
PCs have seen significant advancements in expressivity and learning algorithms including learning tensorized representations \citep{mari2023unifying, liu2023scaling, liu2023understanding,ventola2023probabilistic, gala2024probabilistic,gala2024scaling,loconte2025relationship}, sparse structures \citep{di2015learning,di2017fast, trapp2019bayesian,vergari2019automatic,dang2022sparse,yang23bayesian}, and extending them to have negative parameters \citep{loconte2024subtractive,loconte2024sum, wang2024relationship}.
Nevertheless, their application to representation learning in an autoencoding fashion has been comparatively limited.
The main work in this area is sum-product autoencoding \citepacronym{SPAE}{vergari2018spae}, which proposed two encoding strategies for Sum-Product Networks: CAT embeddings, derived from MPE inference over categorical latent variables, and ACT embeddings, using circuit activations as continuous representations.
However, SPAE derives these embeddings post-hoc from PCs trained for density estimation via maximum likelihood estimation.
Crucially, there is no explicit loss function during SPAE's training that directly optimizes the quality of these embeddings, for example, based on their ability to generate accurate reconstructions.
In \cref{app:sec:eval:variants}, we additionally evaluate the APC framework with an SPAE encoder and decoder and a neural decoder.
Other research has explored integrating PCs with deep generative models (DGMs).
This includes using neural networks to guide PC structure learning or act as conditional components~\citep{shih2021hyperspn, shao2022cspn, martires2024pnc}, incorporating deep learning inductive biases into PCs~\citep{ventola2020residual, butz2019deep, yu2022sumproductattention}, or using PCs within hybrid DGMs for tasks like knowledge distillation or in combination with normalizing flows and VAEs \citep{liu2023scaling, liu2023understanding, tom2020sptn, sidheekh2023probabilistic, correia2023continuous, gala2024probabilistic}.
\citet{aaron2017aespns} proposed a hybrid model using two separate circuits for data and embeddings alongside a neural autoencoder to refine samples.
In contrast to these approaches, \methods introduce a framework where probabilistic embeddings are \emph{explicitly} modeled as random variables within a single PC encoder and are learned end-to-end.

\paragraph{Neural Autoencoders for Representation Learning.}
Neural autoencoders (AEs) form the foundation of many modern representation learning techniques \citep{Hinton2006ReducingTD}.
Early extensions like denoising autoencoders~\citepacronym{DAEs}{vincent2008extracting} and contractive autoencoders \citepacronym{CAEs}{salah2011contractive} focused on learning robust and invariant features.
The advent of variational autoencoders~\citepacronym{VAEs}{kingma2014auto,rezende2014stochasticba} introduced a principled probabilistic approach, enabling generative modeling by learning a stochastic mapping to a latent space, typically regularized towards a simple prior.
Despite their success, VAEs often rely on approximate inference and simplified posterior distributions (e.g., Gaussians), which can limit their expressiveness.
Significant research has aimed to overcome these limitations through richer posterior families like normalizing flows~\citep{Rezende2015VariationalFlows, Kingma2016improved, Louizos2016Structured}, hierarchical latent structures~\citep{Sonderby2016, vahdat2020nvae}, discrete representations like VQ-VAEs~\citep{vanDenOord2017, razavi2019generating}, and improved training objectives~\citep{Burda2016, higgins2017betavae, Tomczak2018}.
Deterministic variants like regularized autoencoders~\citepacronym{RAEs}{ghosh2020rae} have also been explored to structure the latent space without the sampling complexities of VAEs, often relying on post-hoc density estimation for generation.
Furthermore various neural AE adaptations have been proposed to address the real-world scenario of missing data.
These include techniques such as importance-weighted training (MIWAE, \citealp{mattei2019miwae}; not-MIWAE, \citealp{ipsen2021not}; supMIWAE, \citealp{ipsen2022deal}), tailored likelihoods for heterogeneous missing data \citepacronym{HI-VAE}{nazabal2020handling}, modified losses in DAEs \citepacronym{mDAE}{dupuy2024mdae}, and advanced inference or sampling strategies in hierarchical or iterative models~\citep{peis2022hhvaem, kuang2024variational}.
While powerful, these neural autoencoding paradigms, even those specialized for missing data, often involve approximate inference for their probabilistic embeddings or necessitate imputation strategies for incomplete data.
Another self-supervised paradigm is the masked autoencoder \citepacronym{MAE}{he2022masked}, which is a vision transformer-based~\citepacronym{ViT}{dosovitskiy2020vit} approach that learns representations by encoding only a selected subset of visible input patches to reconstruct the masked ones.
MAEs are trained on specifically chosen patch masking patterns and map each unmasked input patch to its own patch-embedding. This is distinct from models like VAEs and \methods, which generate a single embedding for the entire input.
In contrast to these fully neural based models, \methods leverage a PC encoder for exact and tractable probabilistic inference over embeddings and the capability to handle missing observations via marginalization.
The empirical comparison in the following section will further substantiate these distinctions.

\section{Empirical Evaluation}
\label{sec:eval}

To measure the effectiveness of \methods, we conduct a comprehensive series of experiments designed to evaluate their
performance and robustness across multiple tests against an array of model choices. Our evaluation aims to answer the
following four key research questions. \textbf{RQ1} (\cref{sec:eval:reconstruction}): How effectively can \methods
reconstruct data, particularly in the presence of missing values, compared to established PC-based and neural
approaches? \textbf{RQ2} (\cref{sec:eval:embedding-quality,sec:eval:probabilistic-properties}): Do the explicit
probabilistic embeddings learned by \methods yield high-quality, informative representations that are beneficial for
downstream classification tasks, especially when inputs are incomplete? \textbf{RQ3} (\cref{sec:eval:sota-vaes}): Do
modern autoencoder architectures, training procedures, and higher model capacities already improve robustness
against missing data, or does this remain a challenge? \textbf{RQ4} (\cref{sec:eval:knowledge-distillation}):
Can \methods act as effective student models for data-free knowledge distillation from pre-trained generative latent
variable models and improve robustness to missing data compared to the teacher?

To address these questions, we include the existing autoencoding-mechanism SPAE introduced in \cite{vergari2018spae} as
a PC-based reference, a vanilla variational autoencoder~\citep{kingma2014auto} as a neural autoencoding baseline,
MIWAE~\citep{mattei2019miwae} as a VAE variant specifically designed for handling missing data through multiple
imputation rounds, and missForest~\citep{stekhoven2011missForest}, an iterative imputation method based on random
forests as a strong classic machine learning baseline. All models are evaluated on common image benchmark datasets, as well as the 20
DEBD binary tabular datasets~\citep{Lowd2010LearningMN,Haaren2012MarkovNS,Bekker2015TractableLF,Larochelle2011TheNA}.

\subsection{Reconstructions: Comparing Circuit and Neural Encoders}
\label{sec:eval:reconstruction}

We assess reconstruction robustness against missing data by evaluating models under two primary missing data mechanisms:
missing completely at random (MCAR) and missing at random (MAR). For MCAR, we introduce increasing levels of input
corruption by randomly dropping pixels chosen from a uniform distribution, with the percentage of missing data ranging
from 0\% to 95\% in 5\% steps. This is the default corruption method unless otherwise specified. For MAR, we evaluate
specific structured missingness patterns, such as missing entire regions of an image (e.g., left-to-right,
center-to-border). All models, with the exception of MIWAE, were trained on the complete datasets without missing
values. For MIWAE, training was performed using MCAR-style data in which 50\% of entries were missing. For each metric,
we report the missing-input MSE averaged over the different levels of corruptions.
We repeat all experiments five times with different seeds and report their mean and standard deviation. We refer the
reader to \cref{app:eval-protocol} for additional details. Notably, missForest is an imputation algorithm that operates
directly in the input space, unlike the autoencoding methods, which learn a compressed representation in an embedding
space. Consequently, when no data is missing (0\% corruption), missForest can achieve a perfect reconstruction error of
$0.0$. In contrast, autoencoder-based approaches are inherently limited by their embedding bottleneck, resulting in
non-zero reconstruction error even with complete inputs.

\begin{figure}[t]
  \centering
  \includegraphics[width=\textwidth]{ 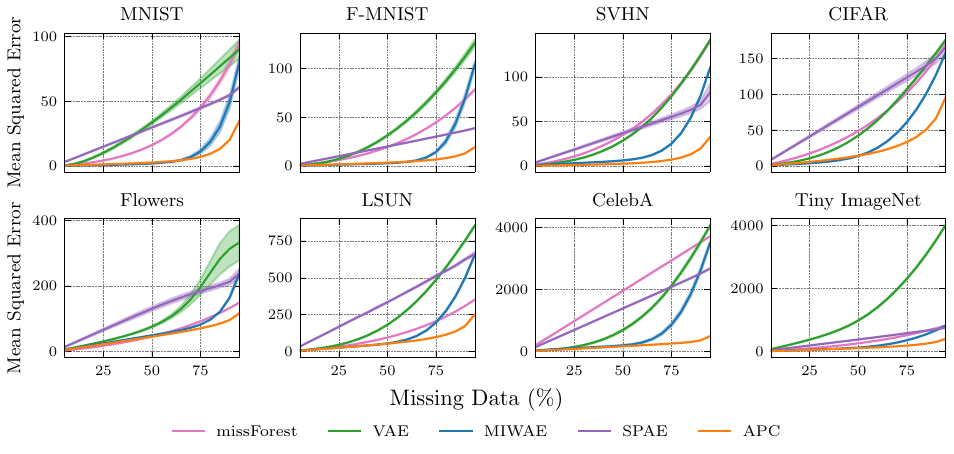 }
  \caption{%
    \textbf{\methods can deliver lower reconstruction errors than PC, VAE, and missForest baselines in the asymptotic regime of MCAR-style randomly missing data}. Reconstruction mean squared error is reported across all image datasets as the degree of MCAR corruption increases from 0\% to 95\%.
  }
  \label{fig:rec-missing-graph}
\end{figure}

We first evaluate the reconstruction capabilities of \methods and examine performance on image data. In
\cref{fig:rec-missing-graph}, we show MSE reconstruction error trends (SSIM in \cref{app:fig:rec-missing-graph-ssim}) as
the percentage of MCAR-style missing pixels increases from 0\% to 95\%. While all models perform comparably at low
corruption levels, their results diverge quickly as corruption increases: neural autoencoders show rapid performance
degradation. At the same time, \methods maintain lower reconstruction error even at high corruption levels. It is
important to highlight that APC, VAE, and MIWAE utilize \emph{the same neural decoder architecture}. Therefore, the
observed differences in performance can be solely attributed to the choice of the encoder model and the corresponding
encoding scheme. \methods demonstrate superior performance across all datasets, with lower MSE and higher SSIM compared
to both neural and PC-based models. As an aggregate, we additionally provide the average reconstruction error over all
corruption levels for MSE in \cref{tab:reconstruction-mse} and for SSIM in \cref{app:tab:reconstruction-ssim}.

\begin{figure}[t]
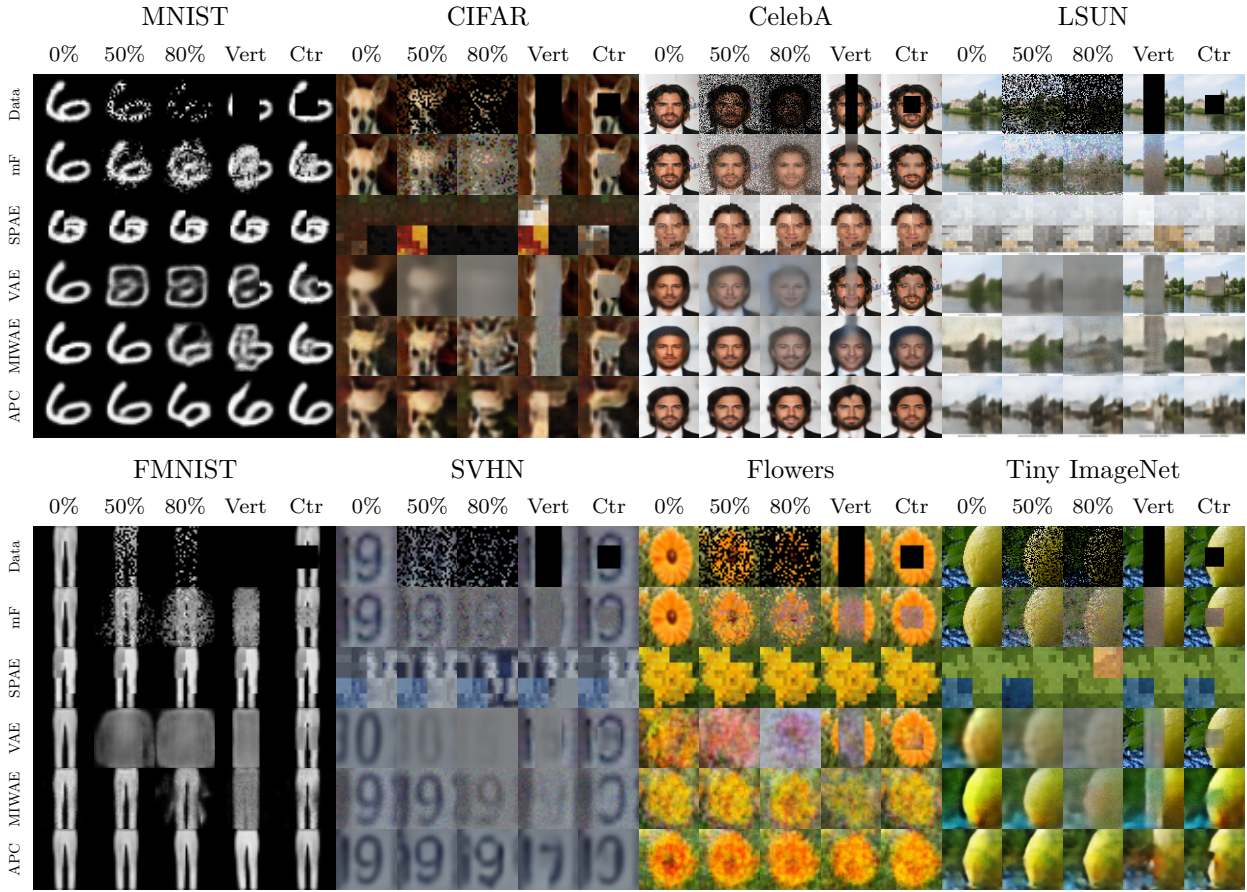

  \tablesize \newcommand{\rbox}[1]{\rotatebox{90}{\small #1}}
\centering

\setlength{\tabcolsep}{0.0pt}  %
\renewcommand{\arraystretch}{0.0}  %

\newcommand{\tabularvertspace}{0.75em}

\newcommand{\imwidth}{1.0\linewidth}
\newcommand{\colwidth}{0.04875\linewidth}
\newcommand{\imda}[5]{%
  \includegraphics[width=\imwidth]{./res/rec-missing-images/#1/#2/#3/data/#4_#5.png }
}
\newcommand{\im}[5]{\includegraphics[width=\imwidth]{./res/rec-missing-images/#1/#2/#3/recs/#4_#5.png }}
\newcommand{\mc}[1]{\multicolumn{1}{c}{#1}}
\newcommand{\mnistIdx}{6}
\newcommand{\cifarIdx}{5}
\newcommand{\celebaIdx}{1}
\newcommand{\lsunIdx}{11}
\newcommand{\p}[1]{\mc{\footnotesize #1}}
\newcommand{\rowlabel}[1]{\rbox{\scalebox{0.625}{#1}}}  %

\newcommand{\vertS}{14} %
\newcommand{\vertM}{22} %
\newcommand{\vertL}{26} %
\newcommand{\centS}{6} %
\newcommand{\centM}{10} %
\newcommand{\centL}{18} %

\newcommand{\datasetrow}[6]{%
 & #1{#2}{#3}{mcar}{#4}{0}      & #1{#2}{#3}{mcar}{#4}{50}      & #1{#2}{#3}{mcar}{#4}{80}      & #1{#2}{#3}{vertical-band}{#4}{#5}      & #1{#2}{#3}{center-to-border}{#4}{#6}%
}
\newcommand{\fullrow}[3]{%
 \rowlabel{#1}
 \datasetrow{#2}{mnist-32}{#3}{\mnistIdx}{\vertS}{\centS}
 \datasetrow{#2}{cifar}{#3}{\cifarIdx}{\vertS}{\centS}
 \datasetrow{#2}{celeba}{#3}{\celebaIdx}{\vertL}{\centL}
 \datasetrow{#2}{lsun}{#3}{\lsunIdx}{\vertM}{\centM} \\
}

\begin{tabular}{m{0.02\linewidth} m{\colwidth} m{\colwidth} m{\colwidth} m{\colwidth} m{\colwidth} m{\colwidth} m{\colwidth} m{\colwidth} m{\colwidth} m{\colwidth} m{\colwidth} m{\colwidth} m{\colwidth} m{\colwidth} m{\colwidth} m{\colwidth} m{\colwidth} m{\colwidth} m{\colwidth} m{\colwidth}}
& \multicolumn{5}{c}{MNIST} & \multicolumn{5}{c}{CIFAR} & \multicolumn{5}{c}{CelebA} & \multicolumn{5}{c}{LSUN} \\[0.75em]
& \p{0\%}                            & \p{50\%}                           & \p{80\%}                            & \p{Vert} & \p{Ctr}
& \p{0\%}                            & \p{50\%}                           & \p{80\%}                            & \p{Vert} & \p{Ctr}
& \p{0\%}                            & \p{50\%}                           & \p{80\%}                            & \p{Vert} & \p{Ctr}
& \p{0\%}                            & \p{50\%}                           & \p{80\%}                            & \p{Vert} & \p{Ctr}                        \\[0.5em]

\fullrow{Data}{\imda}{VAE}
\fullrow{mF}{\im}{missForest}
\fullrow{SPAE}{\im}{ACT_PC}
\fullrow{VAE}{\im}{VAE}
\fullrow{MIWAE}{\im}{MIWAE}
\fullrow{APC}{\im}{APC_dec_nn}
\end{tabular}

\vspace{\tabularvertspace}

\newcommand{\fmnistIdx}{1}
\newcommand{\svhnIdx}{9}
\newcommand{\flowersIdx}{4}
\newcommand{\tinyIdx}{14}

\renewcommand{\fullrow}[3]{%
 \rowlabel{#1}
 \datasetrow{#2}{fmnist-32}{#3}{\fmnistIdx}{\vertS}{\centS}
 \datasetrow{#2}{svhn-extra}{#3}{\svhnIdx}{\vertS}{\centS}
 \datasetrow{#2}{flowers}{#3}{\flowersIdx}{\vertS}{\centS}
 \datasetrow{#2}{tiny-imagenet}{#3}{\tinyIdx}{\vertM}{\centM} \\
}

\begin{tabular}{m{0.02\linewidth} m{\colwidth} m{\colwidth} m{\colwidth} m{\colwidth} m{\colwidth} m{\colwidth} m{\colwidth} m{\colwidth} m{\colwidth} m{\colwidth} m{\colwidth} m{\colwidth} m{\colwidth} m{\colwidth} m{\colwidth} m{\colwidth} m{\colwidth} m{\colwidth} m{\colwidth} m{\colwidth}}
& \multicolumn{5}{c}{FMNIST} & \multicolumn{5}{c}{SVHN} & \multicolumn{5}{c}{Flowers} & \multicolumn{5}{c}{Tiny ImageNet} \\[0.75em]
& \p{0\%}                            & \p{50\%}                           & \p{80\%}                            & \p{Vert} & \p{Ctr}
& \p{0\%}                            & \p{50\%}                           & \p{80\%}                            & \p{Vert} & \p{Ctr}
& \p{0\%}                            & \p{50\%}                           & \p{80\%}                            & \p{Vert} & \p{Ctr}
& \p{0\%}                            & \p{50\%}                           & \p{80\%}                            & \p{Vert} & \p{Ctr}                        \\[0.5em]

\fullrow{Data}{\imda}{VAE}
\fullrow{mF}{\im}{missForest}
\fullrow{SPAE}{\im}{ACT_PC}
\fullrow{VAE}{\im}{VAE}
\fullrow{MIWAE}{\im}{MIWAE}
\fullrow{APC}{\im}{APC_dec_nn}
\end{tabular}

  \caption{
    \textbf{While neural encoder models quickly collapse with an increase in missing data ratio and mostly fail on MAR corruptions, \methods
    are able to maintain a stable reconstruction, even at high degrees of MAR and MCAR corruption.} The first row (Data) shows
    inputs with different corruptions, and subsequent rows represent the respective model reconstructions.
  }
  \label{fig:rec-examples}
\end{figure}

In addition, qualitative examples of reconstructions with missing data in \cref{fig:rec-examples}
visually confirm our quantitative findings. We show reconstructions for 0\%, 50\%, and 80\% MCAR-style missing data and
vertical-band as well as center-square MAR-style missing data. As missing data increases, neural autoencoders produce
increasingly blurry and eventually unrecognizable reconstructions. In contrast, \methods maintain recognizable
reconstructions even at high levels of missing data or images missing complete patches, preserving key structural
elements and details across all datasets. We attribute this robustness to the encoder circuit's ability to handle
partial observations through tractable marginalization natively.

While image data inherently contains spatial structure and correlations that neural architectures can exploit through
layers with implicit biases such as convolutions, tabular data typically lacks such inherent structure. We therefore
additionally investigate whether \methods' strong performance on image datasets generalizes to tabular data.
\cref{tab:reconstruction-debd} presents reconstruction performance on the 20 DEBD binary tabular datasets under
increasing percentages of missing values. \methods outperform all other models on 18 out of 20 datasets, demonstrating
that their advantages are not limited to image data but generalize to tabular datasets. This is particularly significant
for tabular data applications where missing values are common in real-world scenarios, such as medical records, survey
and financial data, or environmental modeling.

\subsection{Embedding Quality and Downstream Task Performance}
\label{sec:eval:embedding-quality}

\begin{figure}[t]
  \centering
  \includegraphics[width=\textwidth]{ 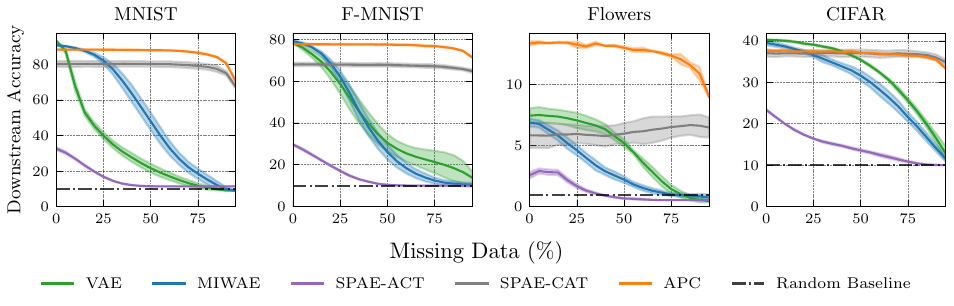 }
  \caption{%
    \textbf{\methods produce linearly seperable embeddings better than VAEs.}
    Downstream task accuracy using a Logistic Regression model under different MCAR-style corruption levels (0\%
    to 95\%) for MNIST, F-MNIST, Flowers, and CIFAR. We observe that \methods generate embeddings even under high
    corruption levels that are still linearly separable. Neural encoders (VAE/MIWAE), on the other hand, quickly
    collapse their embedding representation and thus drop in downstream task performance, when data is missing.%
  }
  \label{fig:downstream-acc-missing-graph}
\end{figure}

While quantitative and qualitative experiments in \cref{sec:eval:reconstruction} so far have investigated reconstruction
performance, a key indicator of representation learning quality is how well the embeddings can be leveraged for downstream task
applications after \emph{unsupervised pre-training} of the autoencoding model. Following a common practice in representation
learning for evaluating the quality of unsupervisedly learned features \citep{Bengio2012RepresentationLA,alain2017understanding-il},
we assess the utility of our models' embeddings by treating them as fixed inputs for a supervised classifier.
For this, a logistic regression model was trained iteratively using stochastic gradient descent (SGD) on these extracted embeddings.
For further details on the training protocol of this downstream task, we refer the reader to \cref{app:eval-protocol}.
We specifically use a logistic regression model to evaluate the learned embeddings' ability to capture
independent and useful data representations. The downstream task performance is measured by the downstream task
classification accuracy (DS-Acc. $\uparrow$), providing insights into the quality and linear separability of the learned
representations. \cref{fig:downstream-acc-missing-graph} presents downstream classification accuracy of the logistic
regression model trained on embeddings extracted from four image datasets under increasing levels of MCAR-style
corruption. The results reveal a stark contrast between \methods and neural autoencoder variants: while neural encoders
achieve slightly higher accuracy with complete data (0\% missing), \method embeddings maintain their classification
performance even as corruption increases to severe levels (90\%), whereas neural encoder embeddings show immediate
degradation in downstream performance, quickly converging towards a baseline random guessing estimator, when data is
missing. This pattern is consistent across all datasets. The robustness to increasing proportions of missing data can
again be attributed to the circuit's fundamental capability for tractable marginalization, allowing inference of
embeddings even with incomplete observations without additional imputation methods.

\begin{figure}[t]
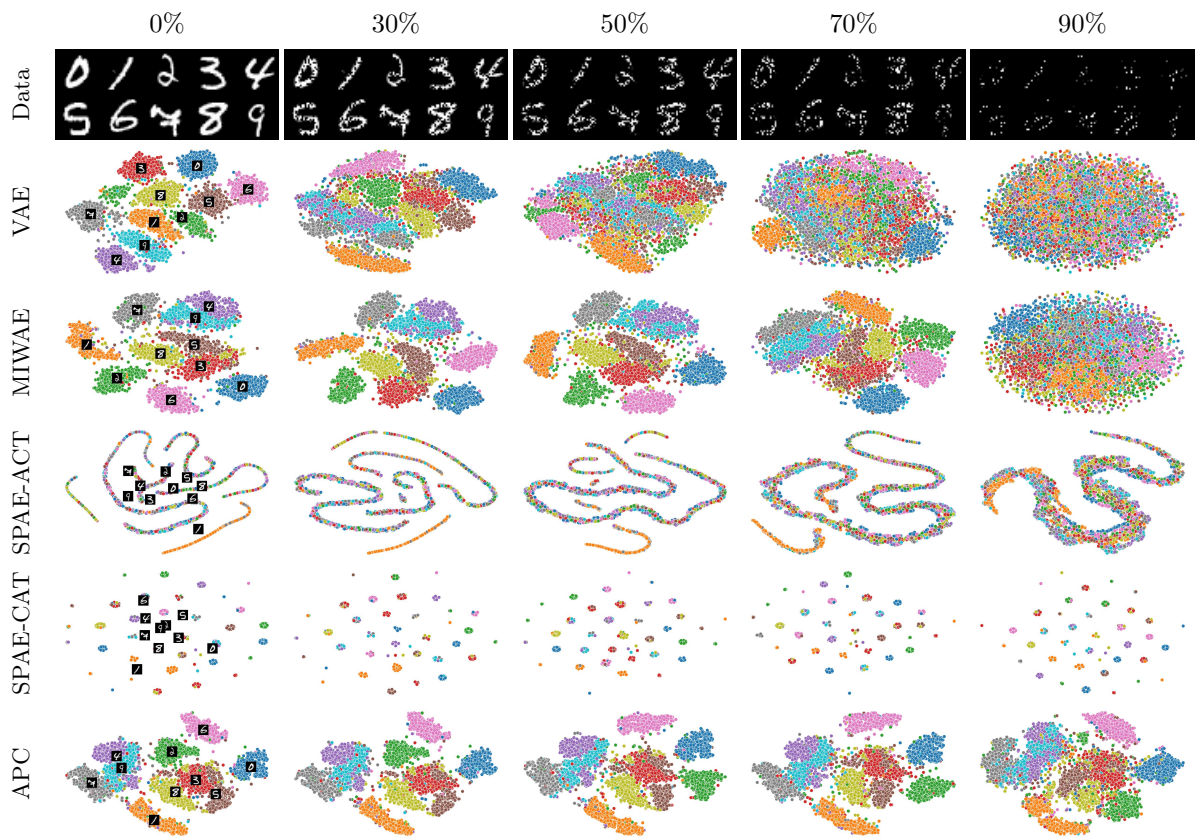

  \tablesize   \setlength{\tabcolsep}{1pt}  %
  \renewcommand{\arraystretch}{0.5}  %

  \newcommand{\imwidth}{1.0\linewidth}
  \newcommand{\colwidth}{0.18\linewidth}
  \newcommand{\im}[2]{\includegraphics[width=\imwidth]{ ./res/tsne-embeddings/#1/#2.jpg }}
  \newcommand{\da}[1]{\includegraphics[width=\imwidth]{ ./res/tsne-embeddings/corruption/#1.png }}
  \newcommand{\mc}[1]{\multicolumn{1}{c}{#1}}
  \newcommand{\p}[1]{\mc{#1}}
  \newcommand{\rbox}[1]{\rotatebox{90}{\small #1}}
  \newcommand{\rowlabel}[1]{\rbox{#1}}

  \centering
  \begin{tabular}{m{0.03\linewidth} m{\colwidth} m{\colwidth} m{\colwidth} m{\colwidth} m{\colwidth}}
                      & \p{0\%}            & \p{30\%}            & \p{50\%}            & \p{70\%}            & \p{90\%}            \\[0.5em]
    \rowlabel{Data}   & \da{0}             & \da{30}             & \da{50}             & \da{70}             & \da{90}             \\
    \rowlabel{VAE}    & \im{VAE}{0}        & \im{VAE}{30}        & \im{VAE}{50}        & \im{VAE}{70}        & \im{VAE}{90}        \\
    \rowlabel{MIWAE}    & \im{MIWAE}{0}        & \im{MIWAE}{30}        & \im{MIWAE}{50}        & \im{MIWAE}{70}        & \im{MIWAE}{90}        \\
    \rowlabel{SPAE-ACT} & \im{ACT_PC}{0}     & \im{ACT_PC}{30}     & \im{ACT_PC}{50}     & \im{ACT_PC}{70}     & \im{ACT_PC}{90}     \\
    \rowlabel{SPAE-CAT} & \im{CAT_PC}{0}     & \im{CAT_PC}{30}     & \im{CAT_PC}{50}     & \im{CAT_PC}{70}     & \im{CAT_PC}{90}     \\
    \rowlabel{APC}    & \im{APC_dec_nn}{0} & \im{APC_dec_nn}{30} & \im{APC_dec_nn}{50} & \im{APC_dec_nn}{70} & \im{APC_dec_nn}{90} \\
  \end{tabular}

  \caption{\textbf{\methods keep a stable embedding space even when corruption is high}. T-SNE projections of MNIST model embeddings for different corruption levels of missing data. Each dot is an
    MNIST test data point and colored according to its class. While neural encoders (*AE) initially (0\%) can separate
    different classes in the embedding space, their representation starts to heavily degrade with increasing corruption
    levels until no separation is possible. }
  \label{fig:tsne-embeddings}
\end{figure}

To provide visual insight into these performance differences, \cref{fig:tsne-embeddings} shows t-SNE projections of
embedding spaces for the MNIST test dataset split at different corruption levels. Because SPAE-ACT embeddings represent
circuit activations, which are log-likelihoods and can have large negative values, we normalize them before t-SNE to
ensure a comparable input range with other models. With complete data (0\% corruption), both neural autoencoders and
\methods produce well-structured embeddings with clear class separation. However, as corruption increases, neural
encoder, even MIWAE, which is specifically trained on missing data, embeddings progressively collapse into an
unstructured mass with no class separation. In contrast, \method embeddings maintain their structure with distinct class
clusters even at 90\% corruption, visually confirming our quantitative results of
\cref{fig:downstream-acc-missing-graph} and their robustness to missing data.

\subsection{Embedding Space Analysis}
\label{sec:eval:probabilistic-properties}

A key advantage of \methods is the fact that the encoder models the joint data-embedding distribution
$\fn{p_{\gC}}{\X, \Z}$ explicitly and can tractable compute marginals, enabling additional probabilistic capabilities
compared to standard and variational autoencoders. In this section, we briefly examine \methods from a probabilistic
perspective by investigating the embedding space and samples decoded from the embedding space.

\begin{figure}[t]
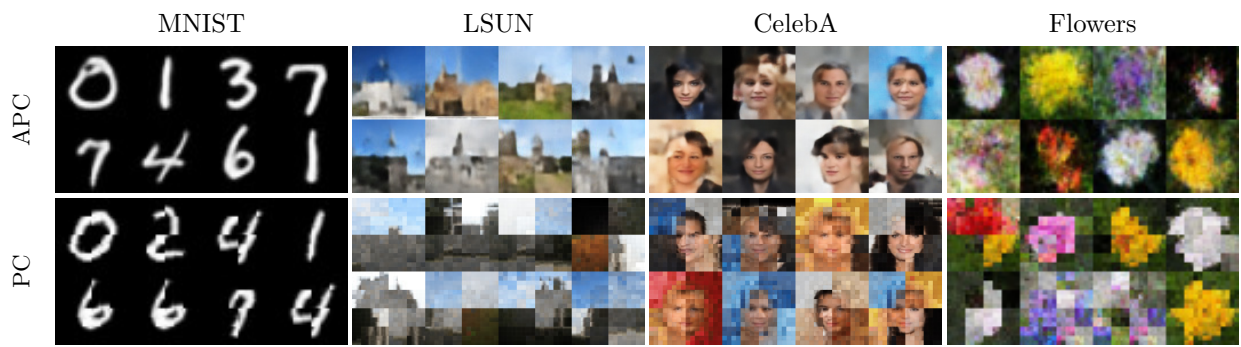

  \setlength{\tabcolsep}{1pt} %
  \newcommand{\imwidth}{1.0\linewidth} \newcommand{\imapc}[1]{\includegraphics[width=\imwidth]{
      ./res/samples/#1/APC_dec_nn/samples-4x2.jpg }} \newcommand{\imspae}[1]{\includegraphics[width=\imwidth]{
      ./res/samples/#1/ACT_PC/samples-4x2.jpg }}

  \newcommand{\rbox}[1]{\rotatebox{90}{\small #1}} \newcommand{\rowlabel}[1]{\rbox{#1}}
  \newcommand{\colwidth}{0.235\linewidth} \newcommand{\mc}[1]{\multicolumn{1}{c}{#1}}
  \begin{center}
    \begin{tabular}{m{0.03\linewidth}m{\colwidth}m{\colwidth}m{\colwidth}m{\colwidth}}
      & \mc{MNIST}         & \mc{LSUN}      & \mc{CelebA}      & \mc{Flowers} \\[0.2em]
      \rowlabel{\method} & \imapc{mnist-32} & \imapc{lsun} & \imapc{celeba} & \imapc{flowers} \\[-0.15em]
      \rowlabel{PC} & \imspae{mnist-32} & \imspae{lsun} & \imspae{celeba} & \imspae{flowers} \\
    \end{tabular}
  \end{center}
  \caption{%
    \textbf{\methods circumvent known visual sampling artifacts of traditional PCs}.
    \method and vanilla PC samples from models trained on MNIST, LSUN (Church), CelebA, and Flowers. \methods
    successfully learn the data-generating distribution and can produce novel samples.
  }
  \label{fig:apc-samples}
\end{figure}

Unlike VAEs, which implicitly model the approximate posterior distribution $\fn{q}{\Z \cbar \X}$ using a neural network
and rely on approximate inference, \methods directly model the joint distribution $\fn{p_{\gC}}{\X, \Z}$ and allow for
exact and tractable probabilistic queries. VAEs are trained to encourage the encoder's approximate posterior
$\fn{q}{\Z \cbar \X}$ to approach a prior distribution, typically a standard normal distribution
$\fn{\mathcal{N}}{0, 1}$. This is achieved by incorporating a KLD term in the loss function that measures the
discrepancy between $\fn{q}{\Z \cbar \X}$ and $\fn{\mathcal{N}}{0, 1}$. Samples are drawn from $\fn{\mathcal{N}}{0, 1}$
after training, assuming the learned approximate posterior is a good approximation of the true posterior
$\fn{p}{\Z \cbar \X}$. However, the encoder rarely perfectly matches the true posterior, leading to a mismatch between
the true posterior and the distribution used for sampling. In contrast, \methods allow us to sample embeddings directly
from the exact marginal distribution $\z \sim \fn{p_{\gC}}{\Z}$, obtained tractably through marginalization of the
PC: $\int \fn{p_{\gC}}{\X, \Z}d\X$. These embeddings are then decoded to generate new data samples.
\cref{fig:apc-samples} showcases these decoded embedding samples across datasets of varying complexity and contrasts
them with a PC solely trained with MLE. The generated samples exhibit diversity and maintain the structural
characteristics of their respective datasets, ranging from clear digit forms in MNIST to more complex architectural
features in LSUN and facial attributes in CelebA. Vanilla PC samples exhibit well-known visual artifacts related to the
circuit structure. \methods circumvent this by deferring the decoding to a neural network. This demonstrates that
\methods successfully capture the underlying data distribution through their tractable encoder, generating higher
quality samples by decoding from the embedding distribution and natively avoiding posterior approximation discrepancies
inherent in VAEs due to the intractable nature of their posterior distributions.

\subsection{Model Capacity Is Insufficient for Missing Data Robustness Even in State-Of-The-Art VAEs}
\label{sec:eval:sota-vaes}

\begin{figure}[t]
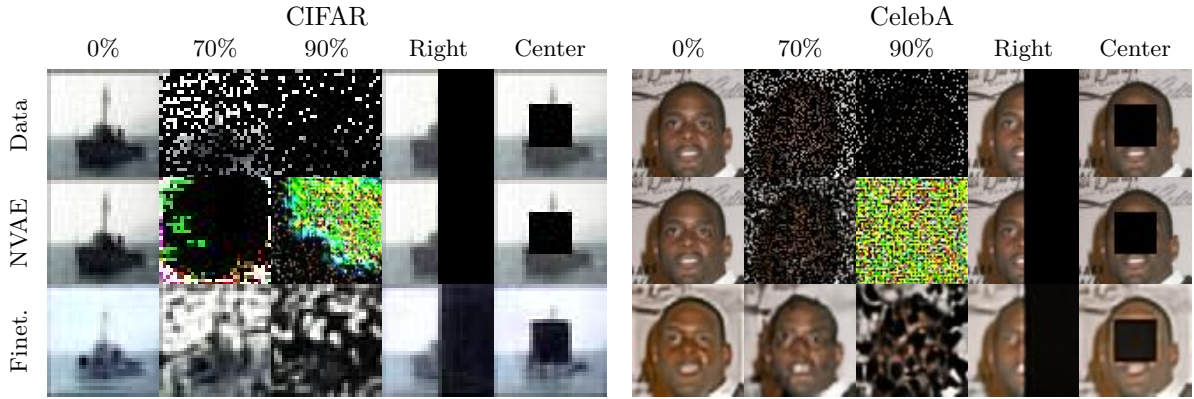

  \setlength{\tabcolsep}{0.0pt} %
  \renewcommand{\arraystretch}{0.0} %
  \newcommand{\imwidth}{1.0\linewidth}
  \newcommand{\imdata}[2]{\includegraphics[width=\imwidth]{res/NVAE/#1/base/data_#2_p_0.jpg }}
  \newcommand{\immask}[3]{\includegraphics[width=\imwidth]{res/NVAE/#1/base/data_#2_masked_p_#3.jpg }}
  \newcommand{\immaskright}[2]{\includegraphics[width=\imwidth]{res/NVAE/#1/base/data_#2_masked_right.jpg }}
  \newcommand{\immaskcenter}[2]{\includegraphics[width=\imwidth]{res/NVAE/#1/base/data_#2_masked_center.jpg }}
  \newcommand{\imbase}[3]{\includegraphics[width=\imwidth]{res/NVAE/#1/base/data_#2_rec_p_#3.jpg }}
  \newcommand{\imbaseright}[2]{\includegraphics[width=\imwidth]{res/NVAE/#1/base/data_#2_rec_right.jpg }}
  \newcommand{\imbasecenter}[2]{\includegraphics[width=\imwidth]{res/NVAE/#1/base/data_#2_rec_center.jpg }}
  \newcommand{\imfine}[4]{\includegraphics[width=\imwidth]{res/NVAE/#1/finetune-p_0.5-ep_#4/data_#2_rec_p_#3.jpg }}
  \newcommand{\imfineright}[3]{\includegraphics[width=\imwidth]{res/NVAE/#1/finetune-p_0.5-ep_#3/data_#2_rec_right.jpg
    }}
  \newcommand{\imfinecenter}[3]{\includegraphics[width=\imwidth]{res/NVAE/#1/finetune-p_0.5-ep_#3/data_#2_rec_center.jpg
    }} \newcommand{\rbox}[1]{\rotatebox{90}{\small #1}} \newcommand{\rowlabel}[1]{\rbox{#1}}
  \newcommand{\colwidth}{0.09\linewidth} \newcommand{\mcdata}[1]{\multicolumn{5}{c}{#1}}
  \newcommand{\mcdisc}[1]{\multicolumn{1}{c}{\small #1}} \newcommand{\cifarIdx}{11} \newcommand{\celebIdx}{7}
  \begin{center}
    \begin{tabular}{m{0.03\linewidth}m{\colwidth}m{\colwidth}m{\colwidth}m{\colwidth}m{\colwidth} p{1.0em}
        m{\colwidth}m{\colwidth}m{\colwidth}m{\colwidth}m{\colwidth}}
      & \mcdata{CIFAR} &
      & \mcdata{CelebA} \\[0.5em]
      & \mcdisc{0\%}                          & \mcdisc{70\%}                          & \mcdisc{90\%}                          & \mcdisc{Right}                          & \mcdisc{Center} &
                                                                                                                                & \mcdisc{0\%}                          & \mcdisc{70\%}                          & \mcdisc{90\%}                          & \mcdisc{Right}                          & \mcdisc{Center}                          \\[0.5em]
      \rowlabel{Data}   & \immask{cifar10}{\cifarIdx}{0}        & \immask{cifar10}{\cifarIdx}{70}        & \immask{cifar10}{\cifarIdx}{90}        & \immaskright{cifar10}{\cifarIdx}        & \immaskcenter{cifar10}{\cifarIdx} &
                                                                                                                                & \immask{celeba_64}{\celebIdx}{0}      & \immask{celeba_64}{\celebIdx}{70}      & \immask{celeba_64}{\celebIdx}{90}      & \immaskright{celeba_64}{\celebIdx}      & \immaskcenter{celeba_64}{\celebIdx}      \\[-0.15em]
      \rowlabel{NVAE}   & \imbase{cifar10}{\cifarIdx}{0}        & \imbase{cifar10}{\cifarIdx}{70}        & \imbase{celeba_64}{\cifarIdx}{90}      & \imbaseright{cifar10}{\cifarIdx}        & \imbasecenter{cifar10}{\cifarIdx} &
                                                                                                                                & \imbase{celeba_64}{\celebIdx}{0}      & \imbase{celeba_64}{\celebIdx}{70}      & \imbase{celeba_64}{\celebIdx}{90}      & \imbaseright{celeba_64}{\celebIdx}      & \imbasecenter{celeba_64}{\celebIdx}      \\[-0.15em]
      \rowlabel{Finet.} & \imfine{cifar10}{\cifarIdx}{0}{500}   & \imfine{cifar10}{\cifarIdx}{70}{500}   & \imfine{cifar10}{\cifarIdx}{90}{500}   & \imfineright{cifar10}{\cifarIdx}{500}   & \imfinecenter{cifar10}{\cifarIdx}{500} &
                                                                                                                                & \imfine{celeba_64}{\celebIdx}{0}{110} & \imfine{celeba_64}{\celebIdx}{70}{110} & \imfine{celeba_64}{\celebIdx}{90}{110} & \imfineright{celeba_64}{\celebIdx}{110} & \imfinecenter{celeba_64}{\celebIdx}{110} \\[-0.15em]
    \end{tabular}
  \end{center}
  \caption{
    \textbf{Even modern VAEs cannot robustly handle data corruption.}
    Example reconstructions from NVAE and MCAR-finetuned NVAE on CIFAR and CelebA samples with MCAR/MAR corruption.
    Standard NVAE retains zero-imputed values, collapsing at high corruption (e.g., 90\%). MCAR-finetuning offers
    marginal improvement for MCAR patterns but fails to generalize to other MAR patterns (e.g., missing right half or square in the center).}
  \label{fig:nvae-recs}
\end{figure}

Our preceding analyses have primarily focused on neural encoder and decoder architectures. However, the autoencoding,
particularly the VAE landscape, has witnessed significant advancements, leading to more sophisticated and complex
architectures and training procedures. Notable examples include Vector Quantized VAEs (VQ-VAE)~\citep{vanDenOord2017}
and VQ-VAE2~\citep{razavi2019generating}, as well as very deep models like VDVAE~\citep{child2020vdvae} and hierarchical
ones like NVAE~\citep{vahdat2020nvae}. To determine whether robust reconstruction under missing data is predominantly a
function of model scale, complexity, and capacity, we evaluate NVAE under MAR and MCAR corruptions as a representative
for modern VAE architectures.

We employed the official publicly available pretrained NVAE models for CIFAR and CelebA from \cite{vahdat2020nvae}.
These models were evaluated on image reconstruction tasks with varying levels of data missingness: 0\%, 70\%, and 90\%
using MCAR-style patterns and two MAR-style patterns (missing right-half, missing center-square), as illustrated in
\cref{fig:nvae-recs}. Initial experiments revealed that the base NVAE model tended to either keep the zero-imputed
values or exhibit failure modes such as oversaturation, which were not seen in previous VAE experiments. To address
this, we finetuned each NVAE model (labeled ``Finet.'' in \cref{fig:nvae-recs}) on its respective dataset with MCAR
patterns of $p=50\%$ uniformly missing pixels for an additional 25\% of its original training epochs. This finetuning
process requires careful calibration, as finetuning degrades the model's original reconstruction capabilities or can
lead to complete collapse. As depicted in \cref{fig:nvae-recs}, the base NVAE achieves nearly perfect reconstructions on
complete data, whereas the finetuned model exhibits significantly poorer performance. This observation is supported by a
substantial increase in bits-per-dimension (BPD) on the full test sets: from 2.91 to 16.93 for CIFAR and from 2.27 to
104.26 for CelebA, for the base to finetuned models, respectively. Although the finetuned model demonstrates some
ability to reconstruct images with 70\% missing pixels, it remains susceptible to similar failure modes as the base
model at 90\% missingness and cannot reconstruct MAR-style corruptions at all.

\subsection{Application: Data-Free Knowledge Distillation}
\label{sec:eval:knowledge-distillation}

Pre-trained autoencoding models are increasingly shared online and offer powerful generative capabilities. However, a
significant limitation, as demonstrated in our earlier evaluations
(\cref{sec:eval:reconstruction,sec:eval:embedding-quality,sec:eval:probabilistic-properties}), is their often fragile
performance when confronted with MCAR and MAR corrupted input data. This raises a practical question: can we transfer
the generative knowledge of a pretrained VAE to our \method framework when the original training data is unavailable?
Knowledge distillation (KD) has emerged as a powerful technique for model compression, adaptation, and continual
learning by transferring knowledge from a teacher model to a student model~\citep{hinton2015distill}. KD has seen an
increasing amount of application in natural language
processing~\citep{tang2019distilling-bert,jiao2019tiny-bert,mou2016word-emb}, computer
vision~\citep{liu2017learning-efficient-cnn,zhou2018dorefanet-tl,yim2017agf}, and speech
recognition~\citep{hinton2015distill,lu2016highway,ramsay2018low-dimensional-bf}. Traditionally, this process requires
access to the original training data, which may not always be available due to privacy concerns, proprietary
restrictions, or storage constraints. Data-free knowledge distillation seeks to overcome these challenges by extracting
transferable knowledge solely from the teacher's learned representations without requiring access to original data.
Approaches range from generative adversarial network~\citep{goodfellow2014gan} based and inspired
methods~\citep{chen2019dafl,han2020robustness-ad} to those that first generate synthetic data through specific
distillation objectives and then train students on this synthetic
data~\citep{mordvintsev2015deepdream,hongxu2019deepinversion,braun2023cake}, align intermediate features between teacher
and student~\citep{Wang2021DataFreeKD} to black-box hard-label based methods~\citep{wang2021zskdb}.

In this section, we explore the capabilities of \methods as robust data-free knowledge
distillers. Given the explicit probabilistic nature and tractable inference in \methods, we investigate their potential
as student models to distill knowledge from generative latent variable models, such as VAEs, while circumventing the
need for the original dataset. This approach allows us to examine whether \methods can learn the data distribution
captured by a VAE teacher, while additionally benefiting from \methods robustness shown in the previous sections. We
outline the data-free knowledge distillation procedure in \cref{alg:knowledge-distillation}. In short, we sample an
embedding from the \method prior $\fn{p_{\gC}}{\Z}$ which is used by the teacher VAE to generate synthetic data
samples. We then train the \method according to the same objectives outlined in \cref{sec:apc} by treating the synthetic
samples as reconstruction ground truth.

\begin{figure}[t]
  \begin{center}
    \input{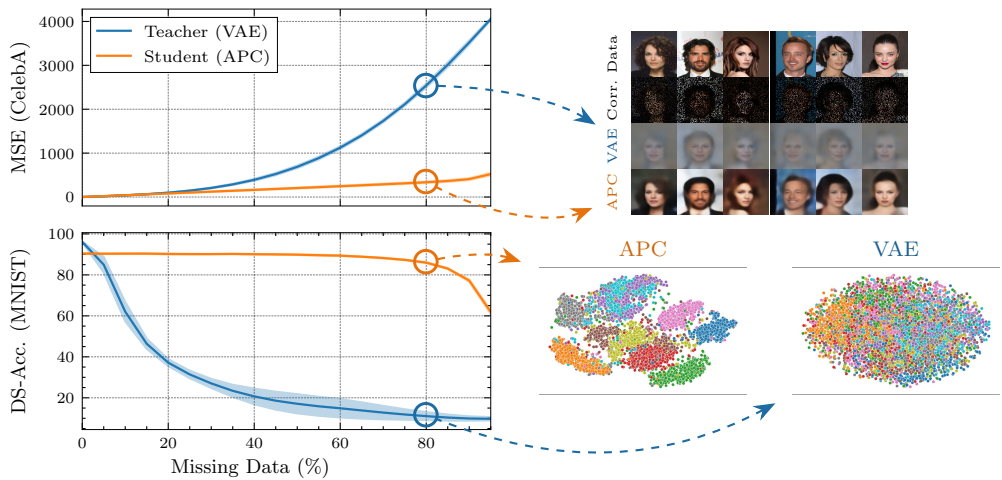}
  \end{center}
  \caption{\textbf{\methods are robust data-free knowledge distillers.} Data-free Knowledge Distillation from VAE to \method. The distilled \method model can learn the VAE's
    distribution and improves in reconstruction and downstream performance under data corruptions.}
  \label{fig:knowledge-distillation}
\end{figure}

In \cref{fig:knowledge-distillation}, we show the results of knowledge distillation from a VAE to an \method of similar
capacity and the same decoder architecture illustratively on two example datasets. The top graph quantifies
reconstruction error on CelebA, measured as the MSE, as a function of MCAR-style missing data
percentage. The teacher VAE exhibits a sharp increase in MSE as data corruption intensifies. In contrast, the student
\method maintains lower reconstruction error, demonstrating its effectiveness in knowledge distillation and showing the
same characteristics of \methods observed in the previous sections. In addition, we highlight the qualitative results of reconstructions with 80\% randomly missing data on the
right, where the \method student model more faithfully reconstructs corrupted images compared to its teacher VAE model.

The bottom graph evaluates downstream task performance on MNIST, measured by logistic regression classification accuracy
using embeddings obtained from the models. As missing data increases, as observed earlier, the teacher VAE's performance
rapidly deteriorates, whereas the student \method maintains high accuracy, indicating superior robustness of its learned
embeddings towards corruptions. We once more highlight these results qualitatively on the right as t-SNE visualizations
of the embedding spaces at 80\% missing data, which further reinforces our findings: While the teacher VAE embeddings
collapse into a dispersed and unstructured distribution, the student \method embeddings maintain well-separated
clusters, suggesting better downstream utility. To provide further evidence, we repeat these experiments for every image
and tabular dataset explored in \cref{sec:eval} and show results for teacher and student reconstructions and downstream
task accuracy under full evidence (Full Evi.) and average MSE over an increasing level of
corruption (0\% to 95\% in 5\% steps) in \cref{tab:latent-kd}. The results confirm that \methods closely match the VAE's
performance under full evidence while surpassing it in robustness to missing data, achieving lower reconstruction errors
and higher downstream accuracy across most datasets. 
This highlights \methods' ability to effectively distill 
knowledge from VAEs without access to original training data while enhancing robustness to data corruption.

\subsection{Ablation Studies}
\label{sec:eval:ablation}

\begin{table}[t]
  \caption{\textbf{Every APC component makes a distinct and complementary contribution to the framework.} Ablation study examining the contribution of key \method components: differentiable sampling for
    end-to-end training (Diff.), neural decoder (NN-Dec.), KLD embedding regularization ($\Z$-Reg.), and joint
    data-embedding log-likelihood regularization ($\fn{p}{\x, \z}$). Performance is evaluated using MSE ($\downarrow$), downstream classification accuracy (DS-Acc. $\uparrow$) under full evidence
    conditions (Full Evi.), and average MSE metrics measuring robustness.}
  \label{tab:ablation}

  \begin{center}
    \tablesize \newcommand{\cm}{\Cmark}
\newcommand{\xm}{\Xmark}
\newcolumntype{x}[1]{>{\centering\arraybackslash\hspace{0pt}}p{#1}}
\begin{tabular}{ccccc p{0.25em} rrrr}
    \maybetoprule
 &       &         &           &                  &  & \multicolumn{2}{c}{MSE ($\downarrow$)} & \multicolumn{2}{c}{DS-Acc ($\uparrow$)}                      \\
    \cmidrule(lr){7-8} \cmidrule(lr){9-10}
 & Diff. & NN-Dec. & $\Z$-Reg. & $\fn{p}{\x, \z}$ &  & \header{Full Evi.}                     & \header{MCAR}       & \header{Full Evi.} & \header{MCAR}       \\
  \cmidrule(lr){2-10}
  \multirow{ 6}{*}{\rotatebox{90}{MNIST}}
 & \cm   & \xm     & \xm       & \xm              &  & \res{79.98}{10.3}                      & \res{63.84}{7.36}  & \res{69.28}{2.69}  & \res{39.16}{5.43}  \\
 & \xm   & \cm     & \cm       & \cm              &  & \res{55.21}{1.24}                      & \res{26.57}{0.48}  & \res{59.69}{3.05}  & \res{58.01}{3.11}  \\
 & \cm   & \xm     & \cm       & \cm              &  & \res{20.73}{0.66}                      & \res{17.12}{0.30}  & \res{50.70}{3.50}  & \res{45.79}{3.24}  \\
 & \cm   & \cm     & \xm       & \cm              &  & \res{8.18}{0.07}                       & \res{8.43}{0.02}   & \res{88.13}{0.22}  & \res{82.94}{0.19}  \\
 & \cm   & \cm     & \cm       & \xm              &  & \res{7.45}{0.19}                       & \res{19.87}{1.14}  & \res{85.98}{1.77}  & \res{65.58}{3.65}  \\
 & \cm   & \cm     & \cm       & \cm              &  & \resB{4.13}{0.10}                      & \resB{5.06}{0.05}  & \resB{88.32}{0.14} & \resB{86.85}{0.11} \\
  \cmidrule(lr){2-10}
  \multirow{ 6}{*}{\rotatebox{90}{CIFAR}}
 & \cm   & \xm     & \xm       & \xm              &  & \res{441.32}{17.5}                     & \res{291.15}{7.43} & \res{34.76}{0.51}  & \res{24.32}{0.51}  \\
 & \xm   & \cm     & \cm       & \cm              &  & \res{193.77}{0.00}                     & \res{92.07}{0.00}  & \res{29.80}{0.00}  & \res{29.51}{0.00}  \\
 & \cm   & \xm     & \cm       & \cm              &  & \res{60.55}{0.86}                      & \res{68.31}{0.93}  & \res{33.09}{0.58}  & \res{29.24}{0.64}  \\
 & \cm   & \cm     & \xm       & \cm              &  & \res{25.45}{0.20}                      & \res{22.52}{0.11}  & \res{36.39}{0.44}  & \res{31.11}{0.15}  \\
 & \cm   & \cm     & \cm       & \xm              &  & \res{35.70}{0.13}                      & \res{67.52}{0.56}  & \res{33.71}{0.27}  & \res{25.79}{0.50}  \\
 & \cm   & \cm     & \cm       & \cm              &  & \resB{16.29}{0.14}                     & \resB{20.64}{0.08} & \resB{37.61}{0.27} & \resB{36.90}{0.14} \\
    \maybebottomrule
\end{tabular}

  \end{center}
\end{table}

Following an extensive evaluation of \methods across a diverse range of datasets and against circuit-based, neural, and
classic models in the previous sections, we now conduct an ablation study to isolate and quantify the contribution of
each component within our framework, thereby strengthening the understanding of \methods. As shown in
\cref{tab:ablation}, we analyze both the removal of all components simultaneously and the exclusion of individual
components while maintaining others. We evaluate four key architectural elements: (1) differentiable sampling (Diff.),
which enables end-to-end gradient flow between the encoder circuit and decoder network; (2) the neural decoder
(NN-Dec.), as opposed to using the circuit itself for reconstruction; (3) embedding regularization via KLD against a
standard Gaussian prior ($\Z$-Reg.); and (4) joint data-embedding log-likelihood regularization ($\fn{p}{\x, \z}$). For each
configuration, \cref{tab:ablation} reports MSE and downstream classification accuracy (DS-Acc.)
under full evidence (Full Evi.) conditions, along with the mean MCAR-style corruptions (MCAR) for both MNIST and CIFAR datasets. We
deliberately selected these two datasets to evaluate our components across different complexity levels: MNIST serves as
a relatively simple benchmark with clear digit structures, while CIFAR represents a more challenging natural image
dataset with complex objects, backgrounds, and color variations. We maintain consistent model capacity throughout the
experiments to ensure fair comparison across all configurations. Specifically for configurations where NN-Dec. =
\XmarkBlack (indicating the absence of a neural decoder), we double the encoder size to match the total parameter count
of a complete \method model, where capacity is evenly distributed between encoder and decoder.

The ablation results in \cref{tab:ablation} highlight the indispensable role of each component in the \method framework.
The baseline configuration (first row) is a circuit that performs both encoding and decoding while being optimized
solely through reconstruction error minimization, mimicking a vanilla autoencoding scheme without additional objectives.
Differentiable sampling (Diff.) is fundamental, as it ensures end-to-end gradient flow from the neural decoder back to
the tractable encoder, optimizing the learned representations for reconstruction. Without this connectivity,
training fails to propagate meaningful gradients, resulting in catastrophic degradation: MSE increases by factors of
13.4$\times$ on MNIST and 11.9$\times$ on CIFAR. Similarly, the neural decoder (NN-Dec.) plays a pivotal role in
reconstruction and downstream performance, as its absence leads to severe drops in reconstruction quality, increasing
MSE by 5$\times$ and 3.7$\times$ on MNIST and CIFAR, respectively. This impact is even more pronounced in corruption
MCAR metrics, demonstrating the decoder's importance for robustness against missing data. Embedding regularization
($\Z$-Reg.) further refines representation learning, nearly halving the MSE on MNIST (from 8.18 to 4.13) and
significantly improving the corruption MCAR MSE from 8.43 to 5.06. These results confirm that each component makes a
distinct and complementary contribution to the \method framework, with their combination yielding better performance
than any partial implementation.

\section{Conclusion}
\label{sec:conclusion}

In this work, we introduced autoencoding probabilistic circuits, a novel framework for representation learning in PCs that leverages their tractability to model explicit probabilistic embeddings.
Autoencoding approaches such as VAEs typically rely on neural networks, which often yield implicit and intractable probabilistic formulations.
In contrast, \methods model the joint distribution of data and embeddings using a PC encoder.
This enables tractable conditional sampling for encoding, yielding embeddings that are samples from a well-defined conditional probability distribution.
Our hybrid architecture combines a PC encoder with a neural network decoder, leveraging the complementary strengths of PCs for tractable probabilistic encoding and the modeling capacity of neural networks for decoding.
Our empirical evaluations across a wide range of image and tabular datasets demonstrate the effectiveness of \methods.
Crucially, \methods show superior reconstruction performance compared to existing circuit-based autoencoding schemes, and to neural encoders in the face of increasing data corruption and missing values.
Furthermore, embeddings learned by \methods prove to be robust and meaningful, maintaining downstream task performance even under high data corruption.

\paragraph{Future Work.} While our study has thoroughly examined the \method framework, numerous avenues remain for further research.
Using a circuit-based encoder grants explicit control over both data and embedding distributions, lifting the typical restriction to independent Gaussians, as seen in VAEs.
Moreover, the circuit structures employed in this work were selected heuristically.
However, the broader literature on PCs offers various approaches for learning circuit structures.
Incorporating these methods could enable more strategic placement of embedding input units, potentially distributing them across different circuit levels.
This could facilitate the formation of low-level and high-level representations, introducing hierarchical structures in the learned embedding space.
Furthermore, \cref{sec:eval:sota-vaes} shows that while modern VAEs achieve high-quality reconstructions on complete data using novel scaling and training tricks, they remain vulnerable to corruption.
We could thus apply analogous methodologies to scale PC encoders, aiming for NVAE-level reconstruction fidelity on complete data while preserving APCs' inherent robustness to data corruptions.

\bibliography{refs}

\begin{thebibliography}{131}
\providecommand{\natexlab}[1]{#1}
\providecommand{\url}[1]{\texttt{#1}}
\expandafter\ifx\csname urlstyle\endcsname\relax
  \providecommand{\doi}[1]{doi: #1}\else
  \providecommand{\doi}{doi: \begingroup \urlstyle{rm}\Url}\fi

\bibitem[Ahmed et~al.(2022)Ahmed, Teso, Chang, den Broeck, and
  Vergari]{ahmed2022semantic}
Kareem Ahmed, Stefano Teso, Kai-Wei Chang, Guy~Van den Broeck, and Antonio
  Vergari.
\newblock Semantic probabilistic layers for neuro-symbolic learning.
\newblock In \emph{Advances in Neural Information Processing Systems
  ({NeurIPS})}, 2022.

\bibitem[Ahmed et~al.(2023)Ahmed, Zeng, Niepert, and den
  Broeck]{ahmed2022simple}
Kareem Ahmed, Zhe Zeng, Mathias Niepert, and Guy~Van den Broeck.
\newblock Simple: A gradient estimator for k-subset sampling.
\newblock In \emph{International Conference on Learning Representations
  ({ICLR})}, 2023.

\bibitem[Alain \& Bengio(2017)Alain and Bengio]{alain2017understanding-il}
Guillaume Alain and Yoshua Bengio.
\newblock Understanding intermediate layers using linear classifier probes.
\newblock \emph{International Conference on Learning Representations ({ICLR})},
  2017.

\bibitem[Ansel et~al.(2024)Ansel, Yang, He, Gimelshein, Jain, Voznesensky, Bao,
  Bell, Berard, Burovski, Chauhan, Chourdia, Constable, Desmaison, DeVito,
  Ellison, Feng, Gong, Gschwind, Hirsh, Huang, Kalambarkar, Kirsch, Lazos,
  Lezcano, Liang, Liang, Lu, Luk, Maher, Pan, Puhrsch, Reso, Saroufim,
  Siraichi, Suk, Suo, Tillet, Wang, Wang, Wen, Zhang, Zhao, Zhou, Zou, Mathews,
  Chanan, Wu, and Chintala]{Ansel_PyTorch_2_Faster_2024}
Jason Ansel, Edward Yang, Horace He, Natalia Gimelshein, Animesh Jain, Michael
  Voznesensky, Bin Bao, Peter Bell, David Berard, Evgeni Burovski, Geeta
  Chauhan, Anjali Chourdia, Will Constable, Alban Desmaison, Zachary DeVito,
  Elias Ellison, Will Feng, Jiong Gong, Michael Gschwind, Brian Hirsh, Sherlock
  Huang, Kshiteej Kalambarkar, Laurent Kirsch, Michael Lazos, Mario Lezcano,
  Yanbo Liang, Jason Liang, Yinghai Lu, CK~Luk, Bert Maher, Yunjie Pan,
  Christian Puhrsch, Matthias Reso, Mark Saroufim, Marcos~Yukio Siraichi, Helen
  Suk, Michael Suo, Phil Tillet, Eikan Wang, Xiaodong Wang, William Wen,
  Shunting Zhang, Xu~Zhao, Keren Zhou, Richard Zou, Ajit Mathews, Gregory
  Chanan, Peng Wu, and Soumith Chintala.
\newblock {PyTorch 2: Faster Machine Learning Through Dynamic Python Bytecode
  Transformation and Graph Compilation}.
\newblock In \emph{29th ACM International Conference on Architectural Support
  for Programming Languages and Operating Systems, Volume 2 (ASPLOS '24)}. ACM,
  2024.
\newblock URL \url{https://pytorch.org/assets/pytorch2-2.pdf}.

\bibitem[Assran et~al.(2023)Assran, Duval, Misra, Bojanowski, Vincent, Rabbat,
  LeCun, and Ballas]{Assran2023SelfSupervisedLF}
Mahmoud Assran, Quentin Duval, Ishan Misra, Piotr Bojanowski, Pascal Vincent,
  Michael~G. Rabbat, Yann LeCun, and Nicolas Ballas.
\newblock Self-supervised learning from images with a joint-embedding
  predictive architecture.
\newblock \emph{Conference on Computer Vision and Pattern Recognition
  ({CVPR})}, 2023.

\bibitem[Bekker et~al.(2015)Bekker, Davis, Choi, Darwiche, and den
  Broeck]{Bekker2015TractableLF}
Jessa Bekker, Jesse Davis, Arthur Choi, Adnan Darwiche, and Guy~Van den Broeck.
\newblock Tractable learning for complex probability queries.
\newblock In \emph{Advances in Neural Information Processing Systems
  ({NeurIPS})}, 2015.

\bibitem[Bengio et~al.(2012)Bengio, Courville, and
  Vincent]{Bengio2012RepresentationLA}
Yoshua Bengio, Aaron~C. Courville, and Pascal Vincent.
\newblock Representation learning: A review and new perspectives.
\newblock \emph{Transactions on Pattern Analysis and Machine Intelligence
  ({TPAMI})}, 2012.

\bibitem[Braun et~al.(2023)Braun, Mundt, and Kersting]{braun2023cake}
Steven Braun, Martin Mundt, and Kristian Kersting.
\newblock Deep classifier mimicry without data access.
\newblock In \emph{International Conference on Artificial Intelligence and
  Statistics ({AISTATS})}, 2023.

\bibitem[Burda et~al.(2016)Burda, Grosse, and Salakhutdinov]{Burda2016}
Yuri Burda, Roger~B. Grosse, and Ruslan Salakhutdinov.
\newblock Importance weighted autoencoders.
\newblock In \emph{International Conference on Learning Representations
  ({ICLR})}, 2016.

\bibitem[Butz et~al.(2019)Butz, de~S.~Oliveira, dos Santos, and
  Teixeira]{butz2019deep}
Cory~J. Butz, Jhonatan de~S.~Oliveira, Andr{\'{e}}~E. dos Santos, and
  Andr{\'{e}}~L. Teixeira.
\newblock Deep convolutional sum-product networks.
\newblock In \emph{Association for the Advancement of Artificial Intelligence
  ({AAAI})}, 2019.

\bibitem[Chen et~al.(2019)Chen, Wang, Xu, Yang, Liu, Shi, Xu, Xu, and
  Tian]{chen2019dafl}
Hanting Chen, Yunhe Wang, Chang Xu, Zhaohui Yang, Chuanjian Liu, Boxin Shi,
  Chunjing Xu, Chao Xu, and Qi~Tian.
\newblock {DAFL}: Data-free learning of student networks.
\newblock In \emph{International Conference on Computer Vision ({ICCV})}, 2019.

\bibitem[Child(2020)]{child2020vdvae}
Rewon Child.
\newblock Very deep vaes generalize autoregressive models and can outperform
  them on images.
\newblock \emph{arXiv preprint arXiv:2011.10650}, abs/2011.10650, 2020.

\bibitem[Choi et~al.(2020)Choi, Vergari, and den Broeck]{choi2020pc}
YooJung Choi, Antonio Vergari, and Guy~Van den Broeck.
\newblock Probabilistic circuits: A unifying framework for tractable
  probabilistic models.
\newblock Technical report, University of California, Los Angeles (UCLA), 2020.

\bibitem[Collier et~al.(2020)Collier, Nazabal, and Williams]{colliervaes}
Mark Collier, Alfredo Nazabal, and Chris Williams.
\newblock Vaes in the presence of missing data.
\newblock In \emph{ICML Workshop on the Art of Learning with Missing Values
  (Artemiss)}, 2020.

\bibitem[Correia et~al.(2023)Correia, Gala, Quaeghebeur, de~Campos, and
  Peharz]{correia2023continuous}
Alvaro H.~C. Correia, Gennaro Gala, Erik Quaeghebeur, Cassio de~Campos, and
  Robert Peharz.
\newblock Continuous mixtures of tractable probabilistic models.
\newblock In \emph{Association for the Advancement of Artificial Intelligence
  ({AAAI})}, 2023.

\bibitem[Dang et~al.(2022{\natexlab{a}})Dang, Liu, and den
  Broeck]{dang2022sparse}
Meihua Dang, Anji Liu, and Guy~Van den Broeck.
\newblock Sparse probabilistic circuits via pruning and growing.
\newblock In \emph{Advances in Neural Information Processing Systems
  ({NeurIPS})}, 2022{\natexlab{a}}.

\bibitem[Dang et~al.(2022{\natexlab{b}})Dang, Liu, Wei, Sankararaman, and den
  Broeck]{dang2022tractable}
Meihua Dang, Anji Liu, Xinzhu Wei, Sriram Sankararaman, and Guy~Van den Broeck.
\newblock Tractable and expressive generative models of genetic variation data.
\newblock In \emph{Research in Computational Molecular Biology},
  2022{\natexlab{b}}.

\bibitem[Darwiche(2003)]{darwiche2003differential}
Adnan Darwiche.
\newblock A differential approach to inference in bayesian networks.
\newblock \emph{Journal of ACM ({JACM})}, 2003.

\bibitem[Darwiche \& Marquis(2002)Darwiche and Marquis]{darwiche_2002}
Adnan Darwiche and Pierre Marquis.
\newblock A knowledge compilation map.
\newblock \emph{JAIR}, 2002.

\bibitem[Deng et~al.(2009)Deng, Dong, Socher, Li, Li, and Fei-Fei]{imagenet}
Jia Deng, Wei Dong, Richard Socher, Li-Jia Li, Kai Li, and Li~Fei-Fei.
\newblock Imagenet: A large-scale hierarchical image database.
\newblock In \emph{Conference on Computer Vision and Pattern Recognition
  ({CVPR})}, 2009.

\bibitem[Dennis \& Ventura(2017)Dennis and Ventura]{aaron2017aespns}
Aaron Dennis and Dan Ventura.
\newblock Autoencoder-enhanced sum-product networks.
\newblock In \emph{IEEE International Conference on Machine Learning and
  Applications ({ICMLA})}, 2017.

\bibitem[Devlin et~al.(2019)Devlin, Chang, Lee, and
  Toutanova]{Devlin2019BERTPO}
Jacob Devlin, Ming-Wei Chang, Kenton Lee, and Kristina Toutanova.
\newblock Bert: Pre-training of deep bidirectional transformers for language
  understanding.
\newblock In \emph{North American Chapter of the Association for Computational
  Linguistics ({NAACL})}, 2019.

\bibitem[{Di Mauro} et~al.(2015){Di Mauro}, Vergari, and
  Esposito]{di2015learning}
Nicola {Di Mauro}, Antonio Vergari, and Floriana Esposito.
\newblock Learning accurate cutset networks by exploiting decomposability.
\newblock In \emph{International Conference of the Italian Association for
  Artificial Intelligence ({AIXIA})}. Springer, 2015.

\bibitem[{Di Mauro} et~al.(2017){Di Mauro}, Vergari, Basile, Esposito,
  et~al.]{di2017fast}
Nicola {Di Mauro}, Antonio Vergari, Teresa~MA Basile, Floriana Esposito, et~al.
\newblock Fast and accurate density estimation with extremely randomized cutset
  networks.
\newblock In \emph{European Conference on Machine Learning and Principles and
  Practice of Knowledge Discovery in Databases ({ECML PKDD})}, 2017.

\bibitem[Dosovitskiy et~al.(2020)Dosovitskiy, Beyer, Kolesnikov, Weissenborn,
  Zhai, Unterthiner, Dehghani, Minderer, Heigold, Gelly, Uszkoreit, and
  Houlsby]{dosovitskiy2020vit}
Alexey Dosovitskiy, Lucas Beyer, Alexander Kolesnikov, Dirk Weissenborn,
  Xiaohua Zhai, Thomas Unterthiner, Mostafa Dehghani, Matthias Minderer, Georg
  Heigold, Sylvain Gelly, Jakob Uszkoreit, and Neil Houlsby.
\newblock An image is worth 16x16 words: Transformers for image recognition at
  scale.
\newblock 2020.

\bibitem[Dupuy et~al.(2024)Dupuy, Chavent, and Dubois]{dupuy2024mdae}
Mariette Dupuy, Marie Chavent, and Remi Dubois.
\newblock mdae : modified denoising autoencoder for missing data imputation.
\newblock \emph{arXiv preprint arXiv:2411.12847}, abs/2411.12847, 2024.

\bibitem[Errica \& Niepert(2023)Errica and Niepert]{errica2023gspn}
Federico Errica and Mathias Niepert.
\newblock Tractable probabilistic graph representation learning with
  graph-induced sum-product networks.
\newblock \emph{arXiv preprint arXiv:2305.10544}, 2023.

\bibitem[Falcon \& {The PyTorch Lightning team}(2019)Falcon and {The PyTorch
  Lightning team}]{Falcon_PyTorch_Lightning_2019}
William Falcon and {The PyTorch Lightning team}.
\newblock {PyTorch Lightning}, March 2019.
\newblock URL \url{https://github.com/Lightning-AI/lightning}.

\bibitem[Gala et~al.(2024{\natexlab{a}})Gala, de~Campos, Peharz, Vergari, and
  Quaeghebeur]{gala2024probabilistic}
Gennaro Gala, Cassio de~Campos, Robert Peharz, Antonio Vergari, and Erik
  Quaeghebeur.
\newblock Probabilistic integral circuits.
\newblock In \emph{International Conference on Artificial Intelligence and
  Statistics ({AISTATS})}, 2024{\natexlab{a}}.

\bibitem[Gala et~al.(2024{\natexlab{b}})Gala, de~Campos, Vergari, and
  Quaeghebeur]{gala2024scaling}
Gennaro Gala, Cassio~P de~Campos, Antonio Vergari, and Erik Quaeghebeur.
\newblock Scaling continuous latent variable models as probabilistic integral
  circuits.
\newblock \emph{Advances in Neural Information Processing Systems ({NeurIPS})},
  2024{\natexlab{b}}.

\bibitem[Ghosh et~al.(2020)Ghosh, Sajjadi, Vergari, Black, and
  Scholkopf]{ghosh2020rae}
Partha Ghosh, Mehdi S.~M. Sajjadi, Antonio Vergari, Michael Black, and Bernhard
  Scholkopf.
\newblock From variational to deterministic autoencoders.
\newblock In \emph{International Conference on Learning Representations
  ({ICLR})}, 2020.

\bibitem[Gondara \& Wang(2018)Gondara and Wang]{Gondara2017MIDAMI}
Lovedeep Gondara and Ke~Wang.
\newblock Mida: Multiple imputation using denoising autoencoders.
\newblock In \emph{Pacific-Asia Conference on Knowledge Discovery and Data
  Mining}, 2018.

\bibitem[Goodfellow et~al.(2014)Goodfellow, Pouget-Abadie, Mirza, Xu,
  Warde-Farley, Ozair, Courville, and Bengio]{goodfellow2014gan}
Ian~J. Goodfellow, Jean Pouget-Abadie, Mehdi Mirza, Bing Xu, David
  Warde-Farley, Sherjil Ozair, Aaron Courville, and Yoshua Bengio.
\newblock Generative adversarial nets.
\newblock In \emph{Advances in Neural Information Processing Systems
  ({NeurIPS})}, 2014.

\bibitem[Gurnani et~al.(2017)Gurnani, Mavani, Gajjar, and
  Khandhediya]{gurnani2017flowers}
Ayesha Gurnani, Viraj Mavani, Vandit Gajjar, and Yash Khandhediya.
\newblock Flower categorization using deep convolutional neural networks, 2017.

\bibitem[Haaren \& Davis(2012)Haaren and Davis]{Haaren2012MarkovNS}
Jan~Van Haaren and Jesse Davis.
\newblock Markov network structure learning: A randomized feature generation
  approach.
\newblock In \emph{Association for the Advancement of Artificial Intelligence
  ({AAAI})}, 2012.

\bibitem[Han et~al.(2021)Han, Park, Wang, and Liu]{han2020robustness-ad}
Pengchao Han, Jihong Park, Shiqiang Wang, and Yejun Liu.
\newblock Robustness and diversity seeking data-free knowledge distillation.
\newblock In \emph{International Conference on Acoustics, Speech, and Signal
  Processing ({ICASSP})}, 2021.

\bibitem[He et~al.(2022)He, Chen, Xie, Li, Doll{\'a}r, and
  Girshick]{he2022masked}
Kaiming He, Xinlei Chen, Saining Xie, Yanghao Li, Piotr Doll{\'a}r, and Ross
  Girshick.
\newblock Masked autoencoders are scalable vision learners.
\newblock In \emph{Conference on Computer Vision and Pattern Recognition
  ({CVPR})}, 2022.

\bibitem[Higgins et~al.(2017)Higgins, Matthey, Pal, Burgess, Glorot, Botvinick,
  Mohamed, and Lerchner]{higgins2017betavae}
Irina Higgins, Lo\"{i}c Matthey, Arka Pal, Christopher~P. Burgess, Xavier
  Glorot, Matthew Botvinick, Shakir Mohamed, and Alexander Lerchner.
\newblock beta-vae: Learning basic visual concepts with a constrained
  variational framework.
\newblock In \emph{International Conference on Learning Representations
  ({ICLR})}, 2017.

\bibitem[Hinton et~al.(2015)Hinton, Vinyals, and Dean]{hinton2015distill}
Geoffrey Hinton, Oriol Vinyals, and Jeffrey Dean.
\newblock Distilling the knowledge in a neural network.
\newblock In \emph{Deep Learning and Representation Learning Workshop
  ({NIPS})}, 2015.

\bibitem[Hinton \& Salakhutdinov(2006)Hinton and
  Salakhutdinov]{Hinton2006ReducingTD}
Geoffrey~E. Hinton and Ruslan Salakhutdinov.
\newblock Reducing the dimensionality of data with neural networks.
\newblock \emph{Science}, 2006.

\bibitem[Ipsen et~al.(2021)Ipsen, Mattei, and Frellsen]{ipsen2021not}
Niels~Bruun Ipsen, Pierre-Alexandre Mattei, and Jes Frellsen.
\newblock not-miwae: Deep generative modelling with missing not at random data.
\newblock In \emph{ICLR 2021-International Conference on Learning
  Representations}, 2021.

\bibitem[Ipsen et~al.(2022)Ipsen, Mattei, and Frellsen]{ipsen2022deal}
Niels~Bruun Ipsen, Pierre-Alexandre Mattei, and Jes Frellsen.
\newblock How to deal with missing data in supervised deep learning?
\newblock In \emph{10th International Conference on Learning Representations},
  2022.

\bibitem[Jiao et~al.(2020)Jiao, Yin, Shang, Jiang, Chen, Li, Wang, and
  Liu]{jiao2019tiny-bert}
Xiaoqi Jiao, Yichun Yin, Lifeng Shang, Xin Jiang, Xiao Chen, Linlin Li, Fang
  Wang, and Qun Liu.
\newblock Tinybert: Distilling {BERT} for natural language understanding.
\newblock In \emph{Conference on Empirical Methods in Natural Language
  Processing ({EMNLP})}, 2020.

\bibitem[Karanam et~al.(2025)Karanam, Mathur, Sidheekh, and
  Natarajan]{karanam2025unified}
Athresh Karanam, Saurabh Mathur, Sahil Sidheekh, and Sriraam Natarajan.
\newblock A unified framework for human-allied learning of probabilistic
  circuits.
\newblock In \emph{Association for the Advancement of Artificial Intelligence
  ({AAAI})}, 2025.

\bibitem[Kingma \& Ba(2015)Kingma and Ba]{kingma2015adam}
Diederik~P. Kingma and Jimmy Ba.
\newblock Adam: A method for stochastic optimization.
\newblock In \emph{International Conference on Learning Representations
  ({ICLR})}, 2015.

\bibitem[Kingma \& Welling(2014)Kingma and Welling]{kingma2014auto}
Diederik~P. Kingma and Max Welling.
\newblock Auto-encoding variational bayes.
\newblock In \emph{International Conference on Learning Representations
  ({ICLR})}, 2014.

\bibitem[Kingma et~al.(2016)Kingma, Salimans, Jozefowicz, Chen, Sutskever, and
  Welling]{Kingma2016improved}
Diederik~P. Kingma, Tim Salimans, Rafal Jozefowicz, Xi~Chen, Ilya Sutskever,
  and Max Welling.
\newblock Improved variational inference with inverse autoregressive flow.
\newblock In \emph{Advances in Neural Information Processing Systems
  ({NeurIPS})}, 2016.

\bibitem[Kisa et~al.(2014)Kisa, den Broeck, Choi, and
  Darwiche]{kisa2014probabilistic}
Doga Kisa, Guy~Van den Broeck, Arthur Choi, and Adnan Darwiche.
\newblock Probabilistic sentential decision diagrams.
\newblock In \emph{International Conference on Principles of Knowledge
  Representation and Reasoning ({KR})}, 2014.

\bibitem[Krizhevsky(2009)]{cifar100}
Alex Krizhevsky.
\newblock Learning multiple layers of features from tiny images.
\newblock Technical report, U. of Toronto, 2009.

\bibitem[Kuang et~al.(2024)Kuang, Song, Wang, and Zhu]{kuang2024variational}
Shenfen Kuang, Jie Song, Shangjiu Wang, and Huafeng Zhu.
\newblock Variational autoencoding with conditional iterative sampling for
  missing data imputation.
\newblock \emph{Mathematics}, 2024.

\bibitem[Kurscheidt et~al.(2025)Kurscheidt, Morettin, Sebastiani, Passerini,
  and Vergari]{kurscheid2025tprobabilistic}
Leander Kurscheidt, Paolo Morettin, Roberto Sebastiani, Andrea Passerini, and
  Antonio Vergari.
\newblock A probabilistic neuro-symbolic layer for algebraic constraint
  satisfaction.
\newblock In \emph{The 41st Conference on Uncertainty in Artificial
  Intelligence}, 2025.

\bibitem[Kusner et~al.(2017)Kusner, Paige, and
  Hern{\'a}ndez-Lobato]{kusner2017grammar}
Matt~J Kusner, Brooks Paige, and Jos{\'e}~Miguel Hern{\'a}ndez-Lobato.
\newblock Grammar variational autoencoder.
\newblock In \emph{International conference on machine learning}, pp.\
  1945--1954. PMLR, 2017.

\bibitem[Lang et~al.(2022)Lang, Mundt, Ventola, Peharz, and
  Kersting]{lang2022diff-sampling-spns}
Steven Lang, Martin Mundt, Fabrizio Ventola, Robert Peharz, and Kristian
  Kersting.
\newblock Elevating perceptual sample quality in probabilistic circuits through
  differentiable sampling.
\newblock In \emph{Proceedings of Machine Learning Research ({PMLR})}, 2022.

\bibitem[Larochelle \& Murray(2011)Larochelle and Murray]{Larochelle2011TheNA}
H.~Larochelle and Iain Murray.
\newblock The neural autoregressive distribution estimator.
\newblock In \emph{International Joint Conference on Artificial Intelligence
  ({IJCAI})}, 2011.

\bibitem[LeCun et~al.(1998)LeCun, Bottou, Bengio, and Haffner]{lecun1998mnist}
Yann LeCun, L{\'{e}}on Bottou, Yoshua Bengio, and Patrick Haffner.
\newblock Gradient-based learning applied to document recognition.
\newblock \emph{Proc. {IEEE}}, 1998.

\bibitem[Li{\'e}vin et~al.(2019)Li{\'e}vin, Dittadi, Maal{\o}e, and
  Winther]{lievin2019towards}
Valentin Li{\'e}vin, Andrea Dittadi, Lars Maal{\o}e, and Ole Winther.
\newblock Towards hierarchical discrete variational autoencoders.
\newblock In \emph{Proceedings of the 2nd Symposium on Advances in Approximate
  Bayesian Inference}, 2019.

\bibitem[Liu et~al.(2022)Liu, Mandt, and den Broeck]{liu2022lossless}
Anji Liu, Stephan Mandt, and Guy~Van den Broeck.
\newblock Lossless compression with probabilistic circuits.
\newblock In \emph{International Conference on Learning Representations
  ({ICLR})}, 2022.

\bibitem[Liu et~al.(2023{\natexlab{a}})Liu, Zhang, and den
  Broeck]{liu2023scaling}
Anji Liu, Honghua Zhang, and Guy~Van den Broeck.
\newblock Scaling up probabilistic circuits by latent variable distillation.
\newblock In \emph{International Conference on Learning Representations
  ({ICLR})}, 2023{\natexlab{a}}.

\bibitem[Liu et~al.(2023{\natexlab{b}})Liu, Liu, Van~den Broeck, and
  Liang]{liu2023understanding}
Xuejie Liu, Anji Liu, Guy Van~den Broeck, and Yitao Liang.
\newblock Understanding the distillation process from deep generative models to
  tractable probabilistic circuits.
\newblock In \emph{International Conference on Machine Learning ({ICML})},
  2023{\natexlab{b}}.

\bibitem[Liu et~al.(2017)Liu, Li, Shen, Huang, Yan, and
  Zhang]{liu2017learning-efficient-cnn}
Zhuang Liu, Jianguo Li, Zhiqiang Shen, Gao Huang, Shoumeng Yan, and Changshui
  Zhang.
\newblock Learning efficient convolutional networks through network slimming.
\newblock In \emph{International Conference on Computer Vision ({ICCV})}, 2017.

\bibitem[Liu et~al.(2015)Liu, Luo, Wang, and Tang]{liu2015faceattributes}
Ziwei Liu, Ping Luo, Xiaogang Wang, and Xiaoou Tang.
\newblock Deep learning face attributes in the wild.
\newblock In \emph{International Conference on Computer Vision ({ICCV})},
  December 2015.

\bibitem[Loconte et~al.(2023)Loconte, Mauro, Peharz, and
  Vergari]{loconte2023turn}
Lorenzo Loconte, Nicola~Di Mauro, Robert Peharz, and Antonio Vergari.
\newblock How to turn your knowledge graph embeddings into generative models
  via probabilistic circuits.
\newblock In \emph{Advances in Neural Information Processing Systems
  ({NeurIPS})}, 2023.

\bibitem[Loconte et~al.(2024)Loconte, Sladek, Mengel, Trapp, Solin, Gillis, and
  Vergari]{loconte2024subtractive}
Lorenzo Loconte, Aleksanteri~Mikulus Sladek, Stefan Mengel, Martin Trapp, Arno
  Solin, Nicolas Gillis, and Antonio Vergari.
\newblock Subtractive mixture models via squaring: Representation and learning.
\newblock In \emph{International Conference on Learning Representations
  ({ICLR})}, 2024.

\bibitem[Loconte et~al.(2025{\natexlab{a}})Loconte, Mari, Gala, Peharz,
  de~Campos, Quaeghebeur, Vessio, and Vergari]{loconte2025relationship}
Lorenzo Loconte, Antonio Mari, Gennaro Gala, Robert Peharz, Cassio de~Campos,
  Erik Quaeghebeur, Gennaro Vessio, and Antonio Vergari.
\newblock What is the relationship between tensor factorizations and circuits
  (and how can we exploit it)?
\newblock \emph{Transactions on Machine Learning Research ({TMLR})},
  2025{\natexlab{a}}.

\bibitem[Loconte et~al.(2025{\natexlab{b}})Loconte, Mengel, and
  Vergari]{loconte2024sum}
Lorenzo Loconte, Stefan Mengel, and Antonio Vergari.
\newblock Sum of squares circuits.
\newblock \emph{Association for the Advancement of Artificial Intelligence
  ({AAAI})}, 2025{\natexlab{b}}.

\bibitem[Loshchilov \& Hutter(2017)Loshchilov and Hutter]{loshchilov2017adamw}
Ilya Loshchilov and Frank Hutter.
\newblock Fixing weight decay regularization in adam.
\newblock \emph{arXiv preprint arXiv:1711.05101}, 2017.

\bibitem[Louizos \& Welling(2016)Louizos and Welling]{Louizos2016Structured}
Christos Louizos and Max Welling.
\newblock Structured and efficient variational deep learning with matrix
  gaussian posteriors.
\newblock In \emph{International Conference on Machine Learning ({ICML})},
  2016.

\bibitem[Lowd \& Davis(2010)Lowd and Davis]{Lowd2010LearningMN}
Daniel Lowd and Jesse Davis.
\newblock Learning markov network structure with decision trees.
\newblock \emph{2010 IEEE International Conference on Data Mining}, 2010.

\bibitem[Lu et~al.(2017)Lu, Guo, and Renals]{lu2016highway}
Liang Lu, Michelle Guo, and Steve Renals.
\newblock Knowledge distillation for small-footprint highway networks.
\newblock In \emph{International Conference on Acoustics, Speech, and Signal
  Processing ({ICASSP})}, 2017.

\bibitem[Luo et~al.(2018)Luo, Cai, Zhang, Xu, and Yuan]{Luo2018MultivariateTS}
Yonghong Luo, Xiangrui Cai, Y.~Zhang, Jun Xu, and Xiaojie Yuan.
\newblock Multivariate time series imputation with generative adversarial
  networks.
\newblock In \emph{Advances in Neural Information Processing Systems
  ({NeurIPS})}, 2018.

\bibitem[Mari et~al.(2023)Mari, Vessio, and Vergari]{mari2023unifying}
Antonio Mari, Gennaro Vessio, and Antonio Vergari.
\newblock Unifying and understanding overparameterized circuit representations
  via low-rank tensor decompositions.
\newblock In \emph{The 6th Workshop on Tractable Probabilistic Modeling}, 2023.

\bibitem[Martires(2024)]{martires2024pnc}
Pedro Zuidberg~Dos Martires.
\newblock Probabilistic neural circuits.
\newblock In \emph{Association for the Advancement of Artificial Intelligence
  ({AAAI})}, 2024.

\bibitem[Mathur et~al.(2023)Mathur, Gogate, and Natarajan]{mathur2023knowledge}
Saurabh Mathur, Vibhav Gogate, and Sriraam Natarajan.
\newblock Knowledge intensive learning of cutset networks.
\newblock In \emph{Uncertainty in Artificial Intelligence ({UAI})}, 2023.

\bibitem[Mathur et~al.(2024)Mathur, Antonucci, and
  Natarajan]{mathur2024knowledge}
Saurabh Mathur, Alessandro Antonucci, and Sriraam Natarajan.
\newblock Knowledge intensive learning of credal networks.
\newblock In \emph{Uncertainty in Artificial Intelligence ({UAI})}, 2024.

\bibitem[Mattei \& Frellsen(2019)Mattei and Frellsen]{mattei2019miwae}
Pierre-Alexandre Mattei and Jes Frellsen.
\newblock Miwae: Deep generative modelling and imputation of incomplete data
  sets.
\newblock In \emph{International Conference on Machine Learning ({ICML})},
  2019.

\bibitem[Mordvintsev et~al.(2015)Mordvintsev, Olah, and
  Tyka]{mordvintsev2015deepdream}
Alexander Mordvintsev, Christopher Olah, and Mike Tyka.
\newblock Inceptionism: Going deeper into neural networks, 2015.
\newblock URL
  \url{https://ai.googleblog.com/2015/06/inceptionism-going-deeper-into-neural.html}.

\bibitem[Mou et~al.(2016)Mou, Jia, Xu, Li, Zhang, and Jin]{mou2016word-emb}
Lili Mou, Ran Jia, Yan Xu, Ge~Li, Lu~Zhang, and Zhi Jin.
\newblock Distilling word embeddings: An encoding approach.
\newblock In \emph{International Conference on Information and Knowledge
  Management ({CIKM})}, 2016.

\bibitem[Nazabal et~al.(2020)Nazabal, Olmos, Ghahramani, and
  Valera]{nazabal2020handling}
Alfredo Nazabal, Pablo~M Olmos, Zoubin Ghahramani, and Isabel Valera.
\newblock Handling incomplete heterogeneous data using vaes.
\newblock \emph{Pattern Recognition}, 2020.

\bibitem[Netzer et~al.(2011)Netzer, Wang, Coates, Bissacco, Wu, and Ng]{svhn}
Yuval Netzer, Tao Wang, Adam Coates, Alessandro Bissacco, Bo~Wu, and Andrew~Y.
  Ng.
\newblock Reading digits in natural images with unsupervised feature learning.
\newblock In \emph{NIPS Workshop on Deep Learning and Unsupervised Feature
  Learning}, 2011.

\bibitem[Peharz et~al.(2017)Peharz, Gens, Pernkopf, and
  Domingos]{peharz2017latent}
Robert Peharz, Robert Gens, Franz Pernkopf, and Pedro~M. Domingos.
\newblock On the latent variable interpretation in sum-product networks.
\newblock \emph{Transactions on Pattern Analysis and Machine Intelligence
  ({TPAMI})}, 2017.

\bibitem[Peharz et~al.(2020{\natexlab{a}})Peharz, Lang, Vergari, Stelzner,
  Molina, Trapp, den Broeck, Kersting, and Ghahramani]{peharz2020einet}
Robert Peharz, Steven Lang, Antonio Vergari, Karl Stelzner, Alejandro Molina,
  Martin Trapp, Guy~Van den Broeck, Kristian Kersting, and Zoubin Ghahramani.
\newblock Einsum networks: Fast and scalable learning of tractable
  probabilistic circuits.
\newblock In \emph{International Conference on Machine Learning ({ICML})},
  2020{\natexlab{a}}.

\bibitem[Peharz et~al.(2020{\natexlab{b}})Peharz, Vergari, Stelzner, Molina,
  Shao, Trapp, Kersting, and Ghahramani]{peharz20a-rat-spn}
Robert Peharz, Antonio Vergari, Karl Stelzner, Alejandro Molina, Xiaoting Shao,
  Martin Trapp, Kristian Kersting, and Zoubin Ghahramani.
\newblock Random sum-product networks: A simple and effective approach to
  probabilistic deep learning.
\newblock In \emph{Uncertainty in Artificial Intelligence ({UAI})},
  2020{\natexlab{b}}.

\bibitem[Peis et~al.(2022)Peis, Ma, and Hern\'{a}ndez-Lobato]{peis2022hhvaem}
Ignacio Peis, Chao Ma, and Jos\'{e}~Miguel Hern\'{a}ndez-Lobato.
\newblock Missing data imputation and acquisition with deep hierarchical models
  and hamiltonian monte carlo.
\newblock In \emph{Advances in Neural Information Processing Systems
  ({NeurIPS})}, 2022.

\bibitem[Pevn\'{y} et~al.(2020)Pevn\'{y}, Sm\'{i}dl, Trapp, Pol\'{a}\v{c}ek,
  and Oberhuber]{tom2020sptn}
Tom\'{a}\v{s} Pevn\'{y}, V\'{a}clav Sm\'{i}dl, Martin Trapp, Ond\v{r}ej
  Pol\'{a}\v{c}ek, and Tom\'{a}\v{s} Oberhuber.
\newblock Sum-product-transform networks: Exploiting symmetries using
  invertible transformations.
\newblock In \emph{International Conference on Probabilistic Graphical Models
  ({PGM})}, 2020.

\bibitem[Poon \& Domingos(2011)Poon and Domingos]{poon2011spn}
Hoifung Poon and Pedro~M. Domingos.
\newblock Sum-product networks: {A} new deep architecture.
\newblock In \emph{Uncertainty in Artificial Intelligence ({UAI})}, 2011.

\bibitem[Ramsay et~al.(2019)Ramsay, Kilgour, Roblek, and
  Sharifi]{ramsay2018low-dimensional-bf}
David~B. Ramsay, Kevin Kilgour, Dominik Roblek, and Matthew Sharifi.
\newblock Low-dimensional bottleneck features for on-device continuous speech
  recognition.
\newblock In \emph{Conference of the International Speech Communication
  Association ({ISCA})}, 2019.

\bibitem[Razavi et~al.(2019)Razavi, van~den Oord, and
  Vinyals]{razavi2019generating}
Ali Razavi, A{\"a}ron van~den Oord, and Oriol Vinyals.
\newblock Generating diverse high-fidelity images with vq-vae-2.
\newblock In \emph{Advances in Neural Information Processing Systems
  ({NeurIPS})}, 2019.

\bibitem[Rezende \& Mohamed(2015)Rezende and
  Mohamed]{Rezende2015VariationalFlows}
Danilo~Jimenez Rezende and Shakir Mohamed.
\newblock Variational inference with normalizing flows.
\newblock In \emph{International Conference on Machine Learning ({ICML})},
  2015.

\bibitem[Rezende et~al.(2014)Rezende, Mohamed, and
  Wierstra]{rezende2014stochasticba}
Danilo~Jimenez Rezende, Shakir Mohamed, and Daan Wierstra.
\newblock Stochastic backpropagation and approximate inference in deep
  generative models.
\newblock In \emph{International Conference on Machine Learning ({ICML})},
  2014.

\bibitem[Rifai et~al.(2011)Rifai, Muller, Glorot, Mesnil, Bengio, and
  Vincent]{rifai2011learning}
Salah Rifai, Xavier Muller, Xavier Glorot, Gr{\'e}goire Mesnil, Yoshua Bengio,
  and Pascal Vincent.
\newblock Learning invariant features through local space contraction.
\newblock \emph{arXiv preprint arXiv:1104.4153}, 2011.

\bibitem[Rubin(1989)]{Rubin1989MultipleIF}
Donald~B. Rubin.
\newblock Multiple imputation for nonresponse in surveys.
\newblock In \emph{International Biometric Society}, 1989.

\bibitem[Salah et~al.(2011)Salah, Vincent, Muller, Gloro, and
  Bengio]{salah2011contractive}
Rifai Salah, P~Vincent, X~Muller, X~Gloro, and Y~Bengio.
\newblock Contractive auto-encoders: Explicit invariance during feature
  extraction.
\newblock In \emph{International Conference on Machine Learning ({ICML})},
  2011.

\bibitem[Severo et~al.(2025)Severo, Su, Liu, Johnson, Karrer, Van Den~Broeck,
  Muckley, and Ullrich]{severo2025lossless}
Daniel Severo, Jingtong Su, Anji Liu, Jeff Johnson, Brian Karrer, Guy Van
  Den~Broeck, Matthew~J. Muckley, and Karen Ullrich.
\newblock Enhancing and evaluating probabilistic circuits for high-resolution
  lossless image compression.
\newblock In \emph{Data Compression Conference (DCC)}, 2025.

\bibitem[Shao et~al.(2022)Shao, Molina, Vergari, Stelzner, Peharz, Liebig, and
  Kersting]{shao2022cspn}
Xiaoting Shao, Alejandro Molina, Antonio Vergari, Karl Stelzner, Robert Peharz,
  Thomas Liebig, and Kristian Kersting.
\newblock Conditional sum-product networks: Modular probabilistic circuits via
  gate functions.
\newblock \emph{International Journal of Approximate Reasoning ({IJAR})}, 2022.

\bibitem[Shih et~al.(2021)Shih, Sadigh, and Ermon]{shih2021hyperspn}
Andy Shih, Dorsa Sadigh, and Stefano Ermon.
\newblock Hyperspns: Compact and expressive probabilistic circuits.
\newblock In \emph{Advances in Neural Information Processing Systems
  ({NeurIPS})}, 2021.

\bibitem[Sidheekh et~al.(2023)Sidheekh, Kersting, and
  Natarajan]{sidheekh2023probabilistic}
Sahil Sidheekh, Kristian Kersting, and Sriraam Natarajan.
\newblock Probabilistic flow circuits: Towards unified deep models for
  tractable probabilistic inference.
\newblock In \emph{Uncertainty in Artificial Intelligence ({UAI})}, 2023.

\bibitem[Sidheekh et~al.(2025)Sidheekh, Tenali, Mathur, Blasch, Kersting, and
  Natarajan]{sidheekh2025credibilityaware}
Sahil Sidheekh, Pranuthi Tenali, Saurabh Mathur, Erik Blasch, Kristian
  Kersting, and Sriraam Natarajan.
\newblock Credibility-aware multimodal fusion using probabilistic circuits.
\newblock In \emph{International Conference on Artificial Intelligence and
  Statistics ({AISTATS})}, 2025.

\bibitem[Simkus \& Gutmann(2023)Simkus and Gutmann]{simkus2023conditional}
Vaidotas Simkus and Michael~U Gutmann.
\newblock Conditional sampling of variational autoencoders via iterated
  approximate ancestral sampling.
\newblock \emph{Transactions on Machine Learning Research ({TMLR})}, 2023.

\bibitem[Simkus \& Gutmann(2024)Simkus and Gutmann]{simkus2024improving}
Vaidotas Simkus and Michael~U Gutmann.
\newblock Improving variational autoencoder estimation from incomplete data
  with mixture variational families.
\newblock \emph{Transactions on Machine Learning Research ({TMLR})}, 2024.

\bibitem[S{\o}nderby et~al.(2016)S{\o}nderby, Raiko, Maal{\o}e, S{\o}nderby,
  and Winther]{Sonderby2016}
Casper~Kaae S{\o}nderby, Tapani Raiko, Lars Maal{\o}e, S{\o}ren~Kaae
  S{\o}nderby, and Ole Winther.
\newblock Ladder variational autoencoders.
\newblock In \emph{Advances in Neural Information Processing Systems
  ({NeurIPS})}, 2016.

\bibitem[Stekhoven \& B{\"u}hlmann(2011)Stekhoven and
  B{\"u}hlmann]{stekhoven2011missForest}
Daniel~J. Stekhoven and Peter B{\"u}hlmann.
\newblock Missforest - non-parametric missing value imputation for mixed-type
  data.
\newblock \emph{Bioinformatics}, 2011.

\bibitem[Sutskever et~al.(2014)Sutskever, Vinyals, and
  Le]{sutskever2014sequence}
Ilya Sutskever, Oriol Vinyals, and Quoc~V Le.
\newblock Sequence to sequence learning with neural networks.
\newblock \emph{Advances in neural information processing systems}, 27, 2014.

\bibitem[Tang et~al.(2019)Tang, Lu, Liu, Mou, Vechtomova, and
  Lin]{tang2019distilling-bert}
Raphael Tang, Yao Lu, Linqing Liu, Lili Mou, Olga Vechtomova, and Jimmy Lin.
\newblock Distilling task-specific knowledge from bert into simple neural
  networks.
\newblock \emph{arXiv preprint arXiv:1903.12136}, 2019.

\bibitem[Tomczak \& Welling(2018)Tomczak and Welling]{Tomczak2018}
Jakub~M. Tomczak and Max Welling.
\newblock Vae with a vampprior.
\newblock In \emph{International Conference on Artificial Intelligence and
  Statistics ({AISTATS})}, 2018.

\bibitem[Trapp et~al.(2019)Trapp, Peharz, Ge, Pernkopf, and
  Ghahramani]{trapp2019bayesian}
Martin Trapp, Robert Peharz, Hong Ge, Franz Pernkopf, and Zoubin Ghahramani.
\newblock Bayesian learning of sum-product networks.
\newblock In \emph{Advances in Neural Information Processing Systems
  ({NeurIPS})}, 2019.

\bibitem[Vahdat \& Kautz(2020)Vahdat and Kautz]{vahdat2020nvae}
Arash Vahdat and Jan Kautz.
\newblock {NVAE}: A deep hierarchical variational autoencoder.
\newblock In \emph{Advances in Neural Information Processing Systems
  ({NeurIPS})}, 2020.

\bibitem[van~den Oord et~al.(2017)van~den Oord, Vinyals, and
  Kavukcuoglu]{vanDenOord2017}
Aaron van~den Oord, Oriol Vinyals, and Koray Kavukcuoglu.
\newblock Neural discrete representation learning.
\newblock In \emph{Advances in Neural Information Processing Systems
  ({NeurIPS})}, 2017.

\bibitem[Vaswani et~al.(2017)Vaswani, Shazeer, Parmar, Uszkoreit, Jones, Gomez,
  Kaiser, and Polosukhin]{Vaswani2017AttentionIA}
Ashish Vaswani, Noam~M. Shazeer, Niki Parmar, Jakob Uszkoreit, Llion Jones,
  Aidan~N. Gomez, Lukasz Kaiser, and Illia Polosukhin.
\newblock Attention is all you need.
\newblock In \emph{Advances in Neural Information Processing Systems
  ({NeurIPS})}, 2017.

\bibitem[Ventola et~al.(2020)Ventola, Stelzner, Molina, and
  Kersting]{ventola2020residual}
Fabrizio Ventola, Karl Stelzner, Alejandro Molina, and Kristian Kersting.
\newblock Residual sum-product networks.
\newblock In \emph{European Workshop on Probabilistic Graphical Models}, 2020.

\bibitem[Ventola et~al.(2023)Ventola, Braun, Yu, Mundt, and
  Kersting]{ventola2023probabilistic}
Fabrizio Ventola, Steven Braun, Zhongjie Yu, Martin Mundt, and Kristian
  Kersting.
\newblock Probabilistic circuits that know what they don't know.
\newblock In \emph{Uncertainty in Artificial Intelligence ({UAI})}, 2023.

\bibitem[Vergari et~al.(2018)Vergari, Peharz, Mauro, Molina, Kersting, and
  Esposito]{vergari2018spae}
Antonio Vergari, Robert Peharz, Nicola~Di Mauro, Alejandro Molina, Kristian
  Kersting, and Floriana Esposito.
\newblock Sum-product autoencoding: Encoding and decoding representations using
  sum-product networks.
\newblock In \emph{Association for the Advancement of Artificial Intelligence
  ({AAAI})}, 2018.

\bibitem[Vergari et~al.(2019{\natexlab{a}})Vergari, Di~Mauro, and
  Esposito]{vergari_2019}
Antonio Vergari, Nicola Di~Mauro, and Floriana Esposito.
\newblock Visualizing and understanding sum-product networks.
\newblock \emph{Machine Learning}, 2019{\natexlab{a}}.

\bibitem[Vergari et~al.(2019{\natexlab{b}})Vergari, Di~Mauro, and Van~den
  Broeck]{vergari2019tractable}
Antonio Vergari, Nicola Di~Mauro, and G~Van~den Broeck.
\newblock Tractable probabilistic models: Representations, algorithms,
  learning, and applications.
\newblock \emph{Tutorial at UAI}, 2019{\natexlab{b}}.

\bibitem[Vergari et~al.(2019{\natexlab{c}})Vergari, Molina, Peharz, Ghahramani,
  Kersting, and Valera]{vergari2019automatic}
Antonio Vergari, Alejandro Molina, Robert Peharz, Zoubin Ghahramani, Kristian
  Kersting, and Isabel Valera.
\newblock Automatic bayesian density analysis.
\newblock In \emph{Association for the Advancement of Artificial Intelligence
  ({AAAI})}, 2019{\natexlab{c}}.

\bibitem[Vergari et~al.(2021)Vergari, Choi, Liu, Teso, and den
  Broeck]{vergari2021compositional}
Antonio Vergari, YooJung Choi, Anji Liu, Stefano Teso, and Guy~Van den Broeck.
\newblock A compositional atlas of tractable circuit operations for
  probabilistic inference.
\newblock In \emph{Advances in Neural Information Processing Systems
  ({NeurIPS})}, 2021.

\bibitem[Vincent et~al.(2008)Vincent, Larochelle, Bengio, and
  Manzagol]{vincent2008extracting}
Pascal Vincent, Hugo Larochelle, Yoshua Bengio, and Pierre-Antoine Manzagol.
\newblock Extracting and composing robust features with denoising autoencoders.
\newblock In \emph{International Conference on Machine Learning ({ICML})},
  2008.

\bibitem[Wang \& Broeck(2025)Wang and Broeck]{wang2024relationship}
Benjie Wang and Guy Van~den Broeck.
\newblock On the relationship between monotone and squared probabilistic
  circuits.
\newblock \emph{Association for the Advancement of Artificial Intelligence
  ({AAAI})}, 2025.

\bibitem[Wang(2021{\natexlab{a}})]{Wang2021DataFreeKD}
Z.~Wang.
\newblock Data-free knowledge distillation with soft targeted transfer set
  synthesis.
\newblock In \emph{Association for the Advancement of Artificial Intelligence
  ({AAAI})}, 2021{\natexlab{a}}.

\bibitem[Wang(2021{\natexlab{b}})]{wang2021zskdb}
Zi~Wang.
\newblock Zero-shot knowledge distillation from a decision-based black-box
  model.
\newblock In \emph{International Conference on Machine Learning ({ICML})},
  2021{\natexlab{b}}.

\bibitem[Williams et~al.(2018)Williams, Nash, and
  Naz{\'a}bal]{williams2018autoencoders}
Christopher~KI Williams, Charlie Nash, and Alfredo Naz{\'a}bal.
\newblock Autoencoders and probabilistic inference with missing data: An exact
  solution for the factor analysis case.
\newblock \emph{arXiv preprint arXiv:1801.03851}, 2018.

\bibitem[Xiao et~al.(2017)Xiao, Rasul, and Vollgraf]{xiao2017fmnist}
Han Xiao, Kashif Rasul, and Roland Vollgraf.
\newblock Fashion-mnist: a novel image dataset for benchmarking machine
  learning algorithms.
\newblock \emph{arXiv preprint arXiv:1708.07747}, 2017.

\bibitem[Xu et~al.(2014)Xu, Ren, Liu, and Jia]{xu2014deep}
Li~Xu, Jimmy~S Ren, Ce~Liu, and Jiaya Jia.
\newblock Deep convolutional neural network for image deconvolution.
\newblock \emph{Advances in neural information processing systems}, 27, 2014.

\bibitem[Yang et~al.(2023)Yang, Gala, and Peharz]{yang23bayesian}
Yang Yang, Gennaro Gala, and Robert Peharz.
\newblock Bayesian structure scores for probabilistic circuits.
\newblock In \emph{International Conference on Artificial Intelligence and
  Statistics ({AISTATS})}, 2023.

\bibitem[Yim et~al.(2017)Yim, Joo, Bae, and Kim]{yim2017agf}
Junho Yim, Donggyu Joo, Ji{-}Hoon Bae, and Junmo Kim.
\newblock A gift from knowledge distillation: Fast optimization, network
  minimization and transfer learning.
\newblock In \emph{Conference on Computer Vision and Pattern Recognition
  ({CVPR})}, 2017.

\bibitem[Yin et~al.(2020)Yin, Molchanov, Alvarez, Li, Mallya, Hoiem, Jha, and
  Kautz]{hongxu2019deepinversion}
Hongxu Yin, Pavlo Molchanov, Jose~M. Alvarez, Zhizhong Li, Arun Mallya, Derek
  Hoiem, Niraj~K. Jha, and Jan Kautz.
\newblock Dreaming to distill: Data-free knowledge transfer via deepinversion.
\newblock In \emph{Conference on Computer Vision and Pattern Recognition
  ({CVPR})}, 2020.

\bibitem[Yoon et~al.(2018)Yoon, Jordon, and van~der Schaar]{Yoon2018GAINMD}
Jinsung Yoon, James Jordon, and Mihaela van~der Schaar.
\newblock {GAIN}: Missing data imputation using generative adversarial nets.
\newblock In \emph{International Conference on Machine Learning ({ICML})},
  2018.

\bibitem[Yu et~al.(2015)Yu, Zhang, Song, Seff, and Xiao]{lsun}
Fisher Yu, Yinda Zhang, Shuran Song, Ari Seff, and Jianxiong Xiao.
\newblock {LSUN:} construction of a large-scale image dataset using deep
  learning with humans in the loop.
\newblock \emph{arXiv preprint arXiv:1506.03365}, 2015.

\bibitem[Yu et~al.(2022)Yu, Dhami, and Kersting]{yu2022sumproductattention}
Zhongjie Yu, Devendra~Singh Dhami, and Kristian Kersting.
\newblock Sum-product-attention networks: Leveraging self-attention in
  energy-based probabilistic circuits.
\newblock In \emph{The 5th Workshop on Tractable Probabilistic Modeling}, 2022.

\bibitem[Zhang et~al.(2023)Zhang, Dang, Peng, and den
  Broeck]{zhang2023tractable}
Honghua Zhang, Meihua Dang, Nanyun Peng, and Guy~Van den Broeck.
\newblock Tractable control for autoregressive language generation.
\newblock In \emph{International Conference on Machine Learning ({ICML})},
  2023.

\bibitem[Zheng et~al.(2018)Zheng, Pronobis, and Rao]{zheng2018graph}
Kaiyu Zheng, Andrzej Pronobis, and Rajesh P.~N. Rao.
\newblock Learning graph-structured sum-product networks for probabilistic
  semantic maps.
\newblock In \emph{Association for the Advancement of Artificial Intelligence
  ({AAAI})}, 2018.

\bibitem[Zhou et~al.(2018)Zhou, Wu, Ni, Zhou, Wen, and
  Zou]{zhou2018dorefanet-tl}
Shuchang Zhou, Yuxin Wu, Zekun Ni, Xinyu Zhou, He~Wen, and Yuheng Zou.
\newblock {DoReFa-Net}: Training low bitwidth convolutional neural networks
  with low bitwidth gradients.
\newblock \emph{arXiv preprint arXiv:1606.06160}, 2018.

\end{thebibliography}


\begin{thebibliography}{113}
\providecommand{\natexlab}[1]{#1}
\providecommand{\url}[1]{\texttt{#1}}
\expandafter\ifx\csname urlstyle\endcsname\relax
  \providecommand{\doi}[1]{doi: #1}\else
  \providecommand{\doi}{doi: \begingroup \urlstyle{rm}\Url}\fi

\bibitem[Adel et~al.(2015)Adel, Balduzzi, and Ghodsi]{adel2015learning}
Tameem Adel, David Balduzzi, and Ali Ghodsi.
\newblock Learning the structure of sum-product networks via an svd-based
  algorithm.
\newblock In \emph{Uncertainty in Artificial Intelligence ({UAI})}, 2015.

\bibitem[Ahmed et~al.(2022)Ahmed, Teso, Chang, den Broeck, and
  Vergari]{ahmed2022semantic}
Kareem Ahmed, Stefano Teso, Kai-Wei Chang, Guy~Van den Broeck, and Antonio
  Vergari.
\newblock Semantic probabilistic layers for neuro-symbolic learning.
\newblock In \emph{Advances in Neural Information Processing Systems
  ({NeurIPS})}, 2022.

\bibitem[Ahmed et~al.(2023)Ahmed, Zeng, Niepert, and den
  Broeck]{Ahmed2022SIMPLEAG}
Kareem Ahmed, Zhe Zeng, Mathias Niepert, and Guy~Van den Broeck.
\newblock Simple: A gradient estimator for k-subset sampling.
\newblock 2023.

\bibitem[Ansel et~al.(2024)Ansel, Yang, He, Gimelshein, Jain, Voznesensky, Bao,
  Bell, Berard, Burovski, Chauhan, Chourdia, Constable, Desmaison, DeVito,
  Ellison, Feng, Gong, Gschwind, Hirsh, Huang, Kalambarkar, Kirsch, Lazos,
  Lezcano, Liang, Liang, Lu, Luk, Maher, Pan, Puhrsch, Reso, Saroufim,
  Siraichi, Suk, Suo, Tillet, Wang, Wang, Wen, Zhang, Zhao, Zhou, Zou, Mathews,
  Chanan, Wu, and Chintala]{Ansel_PyTorch_2_Faster_2024}
Jason Ansel, Edward Yang, Horace He, Natalia Gimelshein, Animesh Jain, Michael
  Voznesensky, Bin Bao, Peter Bell, David Berard, Evgeni Burovski, Geeta
  Chauhan, Anjali Chourdia, Will Constable, Alban Desmaison, Zachary DeVito,
  Elias Ellison, Will Feng, Jiong Gong, Michael Gschwind, Brian Hirsh, Sherlock
  Huang, Kshiteej Kalambarkar, Laurent Kirsch, Michael Lazos, Mario Lezcano,
  Yanbo Liang, Jason Liang, Yinghai Lu, CK~Luk, Bert Maher, Yunjie Pan,
  Christian Puhrsch, Matthias Reso, Mark Saroufim, Marcos~Yukio Siraichi, Helen
  Suk, Michael Suo, Phil Tillet, Eikan Wang, Xiaodong Wang, William Wen,
  Shunting Zhang, Xu~Zhao, Keren Zhou, Richard Zou, Ajit Mathews, Gregory
  Chanan, Peng Wu, and Soumith Chintala.
\newblock {PyTorch 2: Faster Machine Learning Through Dynamic Python Bytecode
  Transformation and Graph Compilation}.
\newblock In \emph{29th ACM International Conference on Architectural Support
  for Programming Languages and Operating Systems, Volume 2 (ASPLOS '24)}. ACM,
  2024.
\newblock URL \url{https://pytorch.org/assets/pytorch2-2.pdf}.

\bibitem[Assran et~al.(2023)Assran, Duval, Misra, Bojanowski, Vincent, Rabbat,
  LeCun, and Ballas]{Assran2023SelfSupervisedLF}
Mahmoud Assran, Quentin Duval, Ishan Misra, Piotr Bojanowski, Pascal Vincent,
  Michael~G. Rabbat, Yann LeCun, and Nicolas Ballas.
\newblock Self-supervised learning from images with a joint-embedding
  predictive architecture.
\newblock \emph{Conference on Computer Vision and Pattern Recognition
  ({CVPR})}, 2023.

\bibitem[Bekker et~al.(2015)Bekker, Davis, Choi, Darwiche, and den
  Broeck]{Bekker2015TractableLF}
Jessa Bekker, Jesse Davis, Arthur Choi, Adnan Darwiche, and Guy~Van den Broeck.
\newblock Tractable learning for complex probability queries.
\newblock In \emph{Advances in Neural Information Processing Systems
  ({NeurIPS})}, 2015.

\bibitem[Braun et~al.(2023)Braun, Mundt, and Kersting]{braun2023cake}
Steven Braun, Martin Mundt, and Kristian Kersting.
\newblock Deep classifier mimicry without data access.
\newblock In \emph{International Conference on Artificial Intelligence and
  Statistics ({AISTATS})}, 2023.

\bibitem[Burda et~al.(2016)Burda, Grosse, and Salakhutdinov]{Burda2016}
Yuri Burda, Roger~B. Grosse, and Ruslan Salakhutdinov.
\newblock Importance weighted autoencoders.
\newblock In \emph{International Conference on Learning Representations
  ({ICLR})}, 2016.

\bibitem[Butz et~al.(2019)Butz, de~S.~Oliveira, dos Santos, and
  Teixeira]{butz2019deep}
Cory~J. Butz, Jhonatan de~S.~Oliveira, Andr{\'{e}}~E. dos Santos, and
  Andr{\'{e}}~L. Teixeira.
\newblock Deep convolutional sum-product networks.
\newblock In \emph{Association for the Advancement of Artificial Intelligence
  ({AAAI})}, 2019.

\bibitem[Chen et~al.(2019)Chen, Wang, Xu, Yang, Liu, Shi, Xu, Xu, and
  Tian]{chen2019dafl}
Hanting Chen, Yunhe Wang, Chang Xu, Zhaohui Yang, Chuanjian Liu, Boxin Shi,
  Chunjing Xu, Chao Xu, and Qi~Tian.
\newblock {DAFL}: Data-free learning of student networks.
\newblock In \emph{International Conference on Computer Vision ({ICCV})}, 2019.

\bibitem[Choi et~al.(2020)Choi, Vergari, and den Broeck]{choi2020pc}
YooJung Choi, Antonio Vergari, and Guy~Van den Broeck.
\newblock Probabilistic circuits: A unifying framework for tractable
  probabilistic models.
\newblock Technical report, University of California, Los Angeles (UCLA), 2020.

\bibitem[Correia et~al.(2023)Correia, Gala, Quaeghebeur, de~Campos, and
  Peharz]{correia2023continuous}
Alvaro H.~C. Correia, Gennaro Gala, Erik Quaeghebeur, Cassio de~Campos, and
  Robert Peharz.
\newblock Continuous mixtures of tractable probabilistic models.
\newblock In \emph{Association for the Advancement of Artificial Intelligence
  ({AAAI})}, 2023.

\bibitem[Dang et~al.(2022{\natexlab{a}})Dang, Liu, and den
  Broeck]{dang2022sparse}
Meihua Dang, Anji Liu, and Guy~Van den Broeck.
\newblock Sparse probabilistic circuits via pruning and growing.
\newblock In \emph{Advances in Neural Information Processing Systems
  ({NeurIPS})}, 2022{\natexlab{a}}.

\bibitem[Dang et~al.(2022{\natexlab{b}})Dang, Liu, Wei, Sankararaman, and den
  Broeck]{dang2022tractable}
Meihua Dang, Anji Liu, Xinzhu Wei, Sriram Sankararaman, and Guy~Van den Broeck.
\newblock Tractable and expressive generative models of genetic variation data.
\newblock In \emph{Research in Computational Molecular Biology},
  2022{\natexlab{b}}.

\bibitem[Darwiche(2003)]{darwiche2003differential}
Adnan Darwiche.
\newblock A differential approach to inference in bayesian networks.
\newblock \emph{Journal of ACM ({JACM})}, 2003.

\bibitem[Deng et~al.(2009)Deng, Dong, Socher, Li, Li, and Fei-Fei]{imagenet}
Jia Deng, Wei Dong, Richard Socher, Li-Jia Li, Kai Li, and Li~Fei-Fei.
\newblock Imagenet: A large-scale hierarchical image database.
\newblock In \emph{Conference on Computer Vision and Pattern Recognition
  ({CVPR})}, 2009.

\bibitem[Dennis \& Ventura(2017)Dennis and Ventura]{aaron2017aespns}
Aaron Dennis and Dan Ventura.
\newblock Autoencoder-enhanced sum-product networks.
\newblock In \emph{IEEE International Conference on Machine Learning and
  Applications ({ICMLA})}, 2017.

\bibitem[Dennis \& Ventura(2012)Dennis and Ventura]{ventura12buildspn}
Aaron~W. Dennis and Dan Ventura.
\newblock Learning the architecture of sum-product networks using clustering on
  variables.
\newblock In \emph{Advances in Neural Information Processing Systems
  ({NeurIPS})}, 2012.

\bibitem[Devlin et~al.(2019)Devlin, Chang, Lee, and
  Toutanova]{Devlin2019BERTPO}
Jacob Devlin, Ming-Wei Chang, Kenton Lee, and Kristina Toutanova.
\newblock Bert: Pre-training of deep bidirectional transformers for language
  understanding.
\newblock In \emph{North American Chapter of the Association for Computational
  Linguistics ({NAACL})}, 2019.

\bibitem[Di~Mauro et~al.(2017)Di~Mauro, Vergari, Basile, and
  Esposito]{di2017fast}
Nicola Di~Mauro, Antonio Vergari, Teresa~MA Basile, and Floriana Esposito.
\newblock Fast and accurate density estimation with extremely randomized cutset
  networks.
\newblock In \emph{European Conference on Machine Learning and Principles and
  Practice of Knowledge Discovery in Databases ({ECML PKDD})}, 2017.

\bibitem[Errica \& Niepert(2023)Errica and Niepert]{errica2023gspn}
Federico Errica and Mathias Niepert.
\newblock Tractable probabilistic graph representation learning with
  graph-induced sum-product networks.
\newblock \emph{arXiv preprint arXiv:2305.10544}, 2023.

\bibitem[Falcon \& {The PyTorch Lightning team}(2019)Falcon and {The PyTorch
  Lightning team}]{Falcon_PyTorch_Lightning_2019}
William Falcon and {The PyTorch Lightning team}.
\newblock {PyTorch Lightning}, March 2019.
\newblock URL \url{https://github.com/Lightning-AI/lightning}.

\bibitem[Gala et~al.(2024{\natexlab{a}})Gala, de~Campos, Peharz, Vergari, and
  Quaeghebeur]{gala2024probabilistic}
Gennaro Gala, Cassio de~Campos, Robert Peharz, Antonio Vergari, and Erik
  Quaeghebeur.
\newblock Probabilistic integral circuits.
\newblock In \emph{International Conference on Artificial Intelligence and
  Statistics ({AISTATS})}, 2024{\natexlab{a}}.

\bibitem[Gala et~al.(2024{\natexlab{b}})Gala, de~Campos, Vergari, and
  Quaeghebeur]{gala2024scaling}
Gennaro Gala, Cassio~P de~Campos, Antonio Vergari, and Erik Quaeghebeur.
\newblock Scaling continuous latent variable models as probabilistic integral
  circuits.
\newblock \emph{Advances in Neural Information Processing Systems ({NeurIPS})},
  2024{\natexlab{b}}.

\bibitem[Gens \& Domingos(2013)Gens and Domingos]{gens2013learnspn}
Robert Gens and Pedro~M. Domingos.
\newblock Learning the structure of sum-product networks.
\newblock In \emph{International Conference on Machine Learning ({ICML})},
  2013.

\bibitem[Ghosh et~al.(2020)Ghosh, Sajjadi, Vergari, Black, and
  Scholkopf]{ghosh2020rae}
Partha Ghosh, Mehdi S.~M. Sajjadi, Antonio Vergari, Michael Black, and Bernhard
  Scholkopf.
\newblock From variational to deterministic autoencoders.
\newblock In \emph{International Conference on Learning Representations
  ({ICLR})}, 2020.

\bibitem[Gondara \& Wang(2018)Gondara and Wang]{Gondara2017MIDAMI}
Lovedeep Gondara and Ke~Wang.
\newblock Mida: Multiple imputation using denoising autoencoders.
\newblock In \emph{Pacific-Asia Conference on Knowledge Discovery and Data
  Mining}, 2018.

\bibitem[Goodfellow et~al.(2014)Goodfellow, Pouget-Abadie, Mirza, Xu,
  Warde-Farley, Ozair, Courville, and Bengio]{goodfellow2014gan}
Ian~J. Goodfellow, Jean Pouget-Abadie, Mehdi Mirza, Bing Xu, David
  Warde-Farley, Sherjil Ozair, Aaron Courville, and Yoshua Bengio.
\newblock Generative adversarial nets.
\newblock In \emph{Advances in Neural Information Processing Systems
  ({NeurIPS})}, 2014.

\bibitem[Gurnani et~al.(2017)Gurnani, Mavani, Gajjar, and
  Khandhediya]{gurnani2017flowers}
Ayesha Gurnani, Viraj Mavani, Vandit Gajjar, and Yash Khandhediya.
\newblock Flower categorization using deep convolutional neural networks, 2017.

\bibitem[Haaren \& Davis(2012)Haaren and Davis]{Haaren2012MarkovNS}
Jan~Van Haaren and Jesse Davis.
\newblock Markov network structure learning: A randomized feature generation
  approach.
\newblock In \emph{Association for the Advancement of Artificial Intelligence
  ({AAAI})}, 2012.

\bibitem[Han et~al.(2021)Han, Park, Wang, and Liu]{han2020robustness-ad}
Pengchao Han, Jihong Park, Shiqiang Wang, and Yejun Liu.
\newblock Robustness and diversity seeking data-free knowledge distillation.
\newblock In \emph{International Conference on Acoustics, Speech, and Signal
  Processing ({ICASSP})}, 2021.

\bibitem[Higgins et~al.(2017)Higgins, Matthey, Pal, Burgess, Glorot, Botvinick,
  Mohamed, and Lerchner]{Higgins2017}
Irina Higgins, Lo\"{i}c Matthey, Arka Pal, Christopher~P. Burgess, Xavier
  Glorot, Matthew Botvinick, Shakir Mohamed, and Alexander Lerchner.
\newblock beta-vae: Learning basic visual concepts with a constrained
  variational framework.
\newblock In \emph{International Conference on Learning Representations
  ({ICLR})}, 2017.

\bibitem[Hinton et~al.(2015)Hinton, Vinyals, and Dean]{hinton2015distill}
Geoffrey Hinton, Oriol Vinyals, and Jeffrey Dean.
\newblock Distilling the knowledge in a neural network.
\newblock In \emph{Deep Learning and Representation Learning Workshop
  ({NIPS})}, 2015.

\bibitem[Hinton \& Salakhutdinov(2006)Hinton and
  Salakhutdinov]{Hinton2006ReducingTD}
Geoffrey~E. Hinton and Ruslan Salakhutdinov.
\newblock Reducing the dimensionality of data with neural networks.
\newblock \emph{Science}, 2006.

\bibitem[Jiao et~al.(2020)Jiao, Yin, Shang, Jiang, Chen, Li, Wang, and
  Liu]{jiao2019tiny-bert}
Xiaoqi Jiao, Yichun Yin, Lifeng Shang, Xin Jiang, Xiao Chen, Linlin Li, Fang
  Wang, and Qun Liu.
\newblock Tinybert: Distilling {BERT} for natural language understanding.
\newblock In \emph{Conference on Empirical Methods in Natural Language
  Processing ({EMNLP})}, 2020.

\bibitem[Kingma \& Ba(2015)Kingma and Ba]{kingma2015adam}
Diederik~P. Kingma and Jimmy Ba.
\newblock Adam: A method for stochastic optimization.
\newblock In \emph{International Conference on Learning Representations
  ({ICLR})}, 2015.

\bibitem[Kingma \& Welling(2014)Kingma and Welling]{kingma2014auto}
Diederik~P. Kingma and Max Welling.
\newblock Auto-encoding variational bayes.
\newblock In \emph{International Conference on Learning Representations
  ({ICLR})}, 2014.

\bibitem[Kingma et~al.(2016)Kingma, Salimans, Jozefowicz, Chen, Sutskever, and
  Welling]{Kingma2016improved}
Diederik~P. Kingma, Tim Salimans, Rafal Jozefowicz, Xi~Chen, Ilya Sutskever,
  and Max Welling.
\newblock Improved variational inference with inverse autoregressive flow.
\newblock In \emph{Advances in Neural Information Processing Systems
  ({NeurIPS})}, 2016.

\bibitem[Kisa et~al.(2014)Kisa, den Broeck, Choi, and
  Darwiche]{kisa2014probabilistic}
Doga Kisa, Guy~Van den Broeck, Arthur Choi, and Adnan Darwiche.
\newblock Probabilistic sentential decision diagrams.
\newblock In \emph{International Conference on Principles of Knowledge
  Representation and Reasoning ({KR})}, 2014.

\bibitem[Krizhevsky(2009)]{cifar100}
Alex Krizhevsky.
\newblock Learning multiple layers of features from tiny images.
\newblock Technical report, U. of Toronto, 2009.

\bibitem[Lang et~al.(2022)Lang, Mundt, Ventola, Peharz, and
  Kersting]{lang2022diff-sampling-spns}
Steven Lang, Martin Mundt, Fabrizio Ventola, Robert Peharz, and Kristian
  Kersting.
\newblock Elevating perceptual sample quality in probabilistic circuits through
  differentiable sampling.
\newblock In \emph{Proceedings of Machine Learning Research ({PMLR})}, 2022.

\bibitem[Larochelle \& Murray(2011)Larochelle and Murray]{Larochelle2011TheNA}
H.~Larochelle and Iain Murray.
\newblock The neural autoregressive distribution estimator.
\newblock In \emph{International Joint Conference on Artificial Intelligence
  ({IJCAI})}, 2011.

\bibitem[LeCun et~al.(1998)LeCun, Bottou, Bengio, and Haffner]{lecun1998mnist}
Yann LeCun, L{\'{e}}on Bottou, Yoshua Bengio, and Patrick Haffner.
\newblock Gradient-based learning applied to document recognition.
\newblock \emph{Proc. {IEEE}}, 1998.

\bibitem[Li{\'e}vin et~al.(2019)Li{\'e}vin, Dittadi, Maal{\o}e, and
  Winther]{lievin2019towards}
Valentin Li{\'e}vin, Andrea Dittadi, Lars Maal{\o}e, and Ole Winther.
\newblock Towards hierarchical discrete variational autoencoders.
\newblock In \emph{Proceedings of the 2nd Symposium on Advances in Approximate
  Bayesian Inference}, 2019.

\bibitem[Liu et~al.(2022)Liu, Mandt, and den Broeck]{liu2022lossless}
Anji Liu, Stephan Mandt, and Guy~Van den Broeck.
\newblock Lossless compression with probabilistic circuits.
\newblock In \emph{International Conference on Learning Representations
  ({ICLR})}, 2022.

\bibitem[Liu et~al.(2023{\natexlab{a}})Liu, Zhang, and den
  Broeck]{liu2023scaling}
Anji Liu, Honghua Zhang, and Guy~Van den Broeck.
\newblock Scaling up probabilistic circuits by latent variable distillation.
\newblock In \emph{International Conference on Learning Representations
  ({ICLR})}, 2023{\natexlab{a}}.

\bibitem[Liu et~al.(2023{\natexlab{b}})Liu, Liu, Van~den Broeck, and
  Liang]{liu2023understanding}
Xuejie Liu, Anji Liu, Guy Van~den Broeck, and Yitao Liang.
\newblock Understanding the distillation process from deep generative models to
  tractable probabilistic circuits.
\newblock In \emph{International Conference on Machine Learning ({ICML})},
  2023{\natexlab{b}}.

\bibitem[Liu et~al.(2017)Liu, Li, Shen, Huang, Yan, and
  Zhang]{liu2017learning-efficient-cnn}
Zhuang Liu, Jianguo Li, Zhiqiang Shen, Gao Huang, Shoumeng Yan, and Changshui
  Zhang.
\newblock Learning efficient convolutional networks through network slimming.
\newblock In \emph{International Conference on Computer Vision ({ICCV})}, 2017.

\bibitem[Liu et~al.(2015)Liu, Luo, Wang, and Tang]{liu2015faceattributes}
Ziwei Liu, Ping Luo, Xiaogang Wang, and Xiaoou Tang.
\newblock Deep learning face attributes in the wild.
\newblock In \emph{International Conference on Computer Vision ({ICCV})},
  December 2015.

\bibitem[Loconte et~al.(2023)Loconte, Mauro, Peharz, and
  Vergari]{loconte2023turn}
Lorenzo Loconte, Nicola~Di Mauro, Robert Peharz, and Antonio Vergari.
\newblock How to turn your knowledge graph embeddings into generative models
  via probabilistic circuits.
\newblock In \emph{Advances in Neural Information Processing Systems
  ({NeurIPS})}, 2023.

\bibitem[Loconte et~al.(2024{\natexlab{a}})Loconte, Mengel, and
  Vergari]{loconte2024sum}
Lorenzo Loconte, Stefan Mengel, and Antonio Vergari.
\newblock Sum of squares circuits.
\newblock \emph{arXiv preprint arXiv:2408.11778}, 2024{\natexlab{a}}.

\bibitem[Loconte et~al.(2024{\natexlab{b}})Loconte, Sladek, Mengel, Trapp,
  Solin, Gillis, and Vergari]{loconte2024subtractive}
Lorenzo Loconte, Aleksanteri~Mikulus Sladek, Stefan Mengel, Martin Trapp, Arno
  Solin, Nicolas Gillis, and Antonio Vergari.
\newblock Subtractive mixture models via squaring: Representation and learning.
\newblock In \emph{International Conference on Learning Representations
  ({ICLR})}, 2024{\natexlab{b}}.

\bibitem[Loshchilov \& Hutter(2017)Loshchilov and Hutter]{loshchilov2017adamw}
Ilya Loshchilov and Frank Hutter.
\newblock Fixing weight decay regularization in adam.
\newblock \emph{arXiv preprint arXiv:1711.05101}, 2017.

\bibitem[Louizos \& Welling(2016)Louizos and Welling]{Louizos2016Structured}
Christos Louizos and Max Welling.
\newblock Structured and efficient variational deep learning with matrix
  gaussian posteriors.
\newblock In \emph{International Conference on Machine Learning ({ICML})},
  2016.

\bibitem[Lowd \& Davis(2010)Lowd and Davis]{Lowd2010LearningMN}
Daniel Lowd and Jesse Davis.
\newblock Learning markov network structure with decision trees.
\newblock \emph{2010 IEEE International Conference on Data Mining}, 2010.

\bibitem[Lu et~al.(2017)Lu, Guo, and Renals]{lu2016highway}
Liang Lu, Michelle Guo, and Steve Renals.
\newblock Knowledge distillation for small-footprint highway networks.
\newblock In \emph{International Conference on Acoustics, Speech, and Signal
  Processing ({ICASSP})}, 2017.

\bibitem[Luo et~al.(2018)Luo, Cai, Zhang, Xu, and Yuan]{Luo2018MultivariateTS}
Yonghong Luo, Xiangrui Cai, Y.~Zhang, Jun Xu, and Xiaojie Yuan.
\newblock Multivariate time series imputation with generative adversarial
  networks.
\newblock In \emph{Advances in Neural Information Processing Systems
  ({NeurIPS})}, 2018.

\bibitem[Martires(2024)]{martires2024pnc}
Pedro Zuidberg~Dos Martires.
\newblock Probabilistic neural circuits.
\newblock In \emph{Association for the Advancement of Artificial Intelligence
  ({AAAI})}, 2024.

\bibitem[Mau{\'a} et~al.(2017)Mau{\'a}, Cozman, Conaty, and
  Campos]{maua2017credal}
Denis~D Mau{\'a}, Fabio~G Cozman, Diarmaid Conaty, and Cassio~P Campos.
\newblock Credal sum-product networks.
\newblock In \emph{International Symposium on Imprecise Probabilities: Theories
  and Applications ({ISIPTA})}, 2017.

\bibitem[Mau{\'a} et~al.(2018)Mau{\'a}, Conaty, Cozman, Poppenhaeger, and
  de~Campos]{maua2018robustifying}
Denis~Deratani Mau{\'a}, Diarmaid Conaty, Fabio~Gagliardi Cozman, Katja
  Poppenhaeger, and Cassio~Polpo de~Campos.
\newblock Robustifying sum-product networks.
\newblock \emph{International Journal of Approximate Reasoning ({IJAR})}, 2018.

\bibitem[Molina et~al.(2018)Molina, Vergari, Mauro, Natarajan, Esposito, and
  Kersting]{molina2018mixed}
Alejandro Molina, Antonio Vergari, Nicola~Di Mauro, Sriraam Natarajan, Floriana
  Esposito, and Kristian Kersting.
\newblock Mixed sum-product networks: {A} deep architecture for hybrid domains.
\newblock In \emph{Association for the Advancement of Artificial Intelligence
  ({AAAI})}, 2018.

\bibitem[Mordvintsev et~al.(2015)Mordvintsev, Olah, and
  Tyka]{mordvintsev2015deepdream}
Alexander Mordvintsev, Christopher Olah, and Mike Tyka.
\newblock Inceptionism: Going deeper into neural networks, 2015.
\newblock URL
  \url{https://ai.googleblog.com/2015/06/inceptionism-going-deeper-into-neural.html}.

\bibitem[Mou et~al.(2016)Mou, Jia, Xu, Li, Zhang, and Jin]{mou2016word-emb}
Lili Mou, Ran Jia, Yan Xu, Ge~Li, Lu~Zhang, and Zhi Jin.
\newblock Distilling word embeddings: An encoding approach.
\newblock In \emph{International Conference on Information and Knowledge
  Management ({CIKM})}, 2016.

\bibitem[Netzer et~al.(2011)Netzer, Wang, Coates, Bissacco, Wu, and Ng]{svhn}
Yuval Netzer, Tao Wang, Adam Coates, Alessandro Bissacco, Bo~Wu, and Andrew~Y.
  Ng.
\newblock Reading digits in natural images with unsupervised feature learning.
\newblock In \emph{NIPS Workshop on Deep Learning and Unsupervised Feature
  Learning}, 2011.

\bibitem[Peharz et~al.(2013)Peharz, Geiger, and Pernkopf]{peharz2013greedy}
Robert Peharz, Bernhard~C Geiger, and Franz Pernkopf.
\newblock Greedy part-wise learning of sum-product networks.
\newblock In \emph{European Conference on Machine Learning and Principles and
  Practice of Knowledge Discovery in Databases ({ECML PKDD})}, 2013.

\bibitem[Peharz et~al.(2017)Peharz, Gens, Pernkopf, and
  Domingos]{peharz2017latent}
Robert Peharz, Robert Gens, Franz Pernkopf, and Pedro~M. Domingos.
\newblock On the latent variable interpretation in sum-product networks.
\newblock \emph{Transactions on Pattern Analysis and Machine Intelligence
  ({TPAMI})}, 2017.

\bibitem[Peharz et~al.(2020{\natexlab{a}})Peharz, Lang, Vergari, Stelzner,
  Molina, Trapp, den Broeck, Kersting, and Ghahramani]{peharz2020einet}
Robert Peharz, Steven Lang, Antonio Vergari, Karl Stelzner, Alejandro Molina,
  Martin Trapp, Guy~Van den Broeck, Kristian Kersting, and Zoubin Ghahramani.
\newblock Einsum networks: Fast and scalable learning of tractable
  probabilistic circuits.
\newblock In \emph{International Conference on Machine Learning ({ICML})},
  2020{\natexlab{a}}.

\bibitem[Peharz et~al.(2020{\natexlab{b}})Peharz, Vergari, Stelzner, Molina,
  Shao, Trapp, Kersting, and Ghahramani]{peharz20a-rat-spn}
Robert Peharz, Antonio Vergari, Karl Stelzner, Alejandro Molina, Xiaoting Shao,
  Martin Trapp, Kristian Kersting, and Zoubin Ghahramani.
\newblock Random sum-product networks: A simple and effective approach to
  probabilistic deep learning.
\newblock In \emph{Uncertainty in Artificial Intelligence ({UAI})},
  2020{\natexlab{b}}.

\bibitem[Pevn\'{y} et~al.(2020)Pevn\'{y}, Sm\'{i}dl, Trapp, Pol\'{a}\v{c}ek,
  and Oberhuber]{tom2020sptn}
Tom\'{a}\v{s} Pevn\'{y}, V\'{a}clav Sm\'{i}dl, Martin Trapp, Ond\v{r}ej
  Pol\'{a}\v{c}ek, and Tom\'{a}\v{s} Oberhuber.
\newblock Sum-product-transform networks: Exploiting symmetries using
  invertible transformations.
\newblock In \emph{International Conference on Probabilistic Graphical Models
  ({PGM})}, 2020.

\bibitem[Poon \& Domingos(2011)Poon and Domingos]{poon2011spn}
Hoifung Poon and Pedro~M. Domingos.
\newblock Sum-product networks: {A} new deep architecture.
\newblock In \emph{Uncertainty in Artificial Intelligence ({UAI})}, 2011.

\bibitem[Rahman et~al.(2014)Rahman, Kothalkar, and Gogate]{rahman2014cutset}
Tahrima Rahman, Prasanna Kothalkar, and Vibhav Gogate.
\newblock Cutset networks: A simple, tractable, and scalable approach for
  improving the accuracy of chow-liu trees.
\newblock In \emph{ECML PKDD}, 2014.

\bibitem[Ramsay et~al.(2019)Ramsay, Kilgour, Roblek, and
  Sharifi]{ramsay2018low-dimensional-bf}
David~B. Ramsay, Kevin Kilgour, Dominik Roblek, and Matthew Sharifi.
\newblock Low-dimensional bottleneck features for on-device continuous speech
  recognition.
\newblock In \emph{Conference of the International Speech Communication
  Association ({ISCA})}, 2019.

\bibitem[Razavi et~al.(2019)Razavi, van~den Oord, and
  Vinyals]{razavi2019generating}
Ali Razavi, A{\"a}ron van~den Oord, and Oriol Vinyals.
\newblock Generating diverse high-fidelity images with vq-vae-2.
\newblock In \emph{Advances in Neural Information Processing Systems
  ({NeurIPS})}, 2019.

\bibitem[Rezende \& Mohamed(2015)Rezende and
  Mohamed]{Rezende2015VariationalFlows}
Danilo~Jimenez Rezende and Shakir Mohamed.
\newblock Variational inference with normalizing flows.
\newblock In \emph{International Conference on Machine Learning ({ICML})},
  2015.

\bibitem[Rezende et~al.(2014)Rezende, Mohamed, and
  Wierstra]{rezende2014stochasticba}
Danilo~Jimenez Rezende, Shakir Mohamed, and Daan Wierstra.
\newblock Stochastic backpropagation and approximate inference in deep
  generative models.
\newblock In \emph{International Conference on Machine Learning ({ICML})},
  2014.

\bibitem[Rooshenas \& Lowd(2014)Rooshenas and Lowd]{rooshenas14learning}
Amirmohammad Rooshenas and Daniel Lowd.
\newblock Learning sum-product networks with direct and indirect variable
  interactions.
\newblock In \emph{International Conference on Machine Learning ({ICML})},
  2014.

\bibitem[Rubin(1989)]{Rubin1989MultipleIF}
Donald~B. Rubin.
\newblock Multiple imputation for nonresponse in surveys.
\newblock 1989.

\bibitem[Salah et~al.(2011)Salah, Vincent, Muller, Gloro, and
  Bengio]{salah2011contractive}
Rifai Salah, P~Vincent, X~Muller, X~Gloro, and Y~Bengio.
\newblock Contractive auto-encoders: Explicit invariance during feature
  extraction.
\newblock In \emph{International Conference on Machine Learning ({ICML})},
  2011.

\bibitem[Shao et~al.(2022)Shao, Molina, Vergari, Stelzner, Peharz, Liebig, and
  Kersting]{shao2022cspn}
Xiaoting Shao, Alejandro Molina, Antonio Vergari, Karl Stelzner, Robert Peharz,
  Thomas Liebig, and Kristian Kersting.
\newblock Conditional sum-product networks: Modular probabilistic circuits via
  gate functions.
\newblock \emph{International Journal of Approximate Reasoning ({IJAR})}, 2022.

\bibitem[Sharir \& Shashua(2018)Sharir and Shashua]{sharir2018sum}
Or~Sharir and Amnon Shashua.
\newblock Sum-product-quotient networks.
\newblock In \emph{International Conference on Artificial Intelligence and
  Statistics ({AISTATS})}, 2018.

\bibitem[Shih et~al.(2021)Shih, Sadigh, and Ermon]{shih2021hyperspn}
Andy Shih, Dorsa Sadigh, and Stefano Ermon.
\newblock Hyperspns: Compact and expressive probabilistic circuits.
\newblock In \emph{Advances in Neural Information Processing Systems
  ({NeurIPS})}, 2021.

\bibitem[Sidheekh et~al.(2023)Sidheekh, Kersting, and
  Natarajan]{sidheekh2023probabilistic}
Sahil Sidheekh, Kristian Kersting, and Sriraam Natarajan.
\newblock Probabilistic flow circuits: Towards unified deep models for
  tractable probabilistic inference.
\newblock In \emph{Uncertainty in Artificial Intelligence ({UAI})}, 2023.

\bibitem[S{\o}nderby et~al.(2016)S{\o}nderby, Raiko, Maal{\o}e, S{\o}nderby,
  and Winther]{Sonderby2016}
Casper~Kaae S{\o}nderby, Tapani Raiko, Lars Maal{\o}e, S{\o}ren~Kaae
  S{\o}nderby, and Ole Winther.
\newblock Ladder variational autoencoders.
\newblock In \emph{Advances in Neural Information Processing Systems
  ({NeurIPS})}, 2016.

\bibitem[Tan \& Peharz(2019)Tan and Peharz]{tan2019hierarchical}
Ping~Liang Tan and Robert Peharz.
\newblock Hierarchical decompositional mixtures of variational autoencoders.
\newblock In Kamalika Chaudhuri and Ruslan Salakhutdinov (eds.), \emph{ICML},
  2019.

\bibitem[Tang et~al.(2019)Tang, Lu, Liu, Mou, Vechtomova, and
  Lin]{tang2019distilling-bert}
Raphael Tang, Yao Lu, Linqing Liu, Lili Mou, Olga Vechtomova, and Jimmy Lin.
\newblock Distilling task-specific knowledge from bert into simple neural
  networks.
\newblock \emph{arXiv preprint arXiv:1903.12136}, 2019.

\bibitem[Tomczak \& Welling(2018)Tomczak and Welling]{Tomczak2018}
Jakub~M. Tomczak and Max Welling.
\newblock Vae with a vampprior.
\newblock In \emph{International Conference on Artificial Intelligence and
  Statistics ({AISTATS})}, 2018.

\bibitem[Trapp et~al.(2019)Trapp, Peharz, Ge, Pernkopf, and
  Ghahramani]{trapp2019bayesian}
Martin Trapp, Robert Peharz, Hong Ge, Franz Pernkopf, and Zoubin Ghahramani.
\newblock Bayesian learning of sum-product networks.
\newblock In \emph{Advances in Neural Information Processing Systems
  ({NeurIPS})}, 2019.

\bibitem[Trapp et~al.(2020)Trapp, Peharz, Pernkopf, and
  Rasmussen]{trapp2020deepgaussianprocess}
Martin Trapp, Robert Peharz, Franz Pernkopf, and Carl~Edward Rasmussen.
\newblock Deep structured mixtures of gaussian processes.
\newblock In \emph{International Conference on Artificial Intelligence and
  Statistics ({AISTATS})}, 2020.

\bibitem[Vahdat \& Kautz(2020{\natexlab{a}})Vahdat and Kautz]{Vahdat2020}
Arash Vahdat and Jan Kautz.
\newblock Nvae: A deep hierarchical variational autoencoder.
\newblock In \emph{Advances in Neural Information Processing Systems
  ({NeurIPS})}, 2020{\natexlab{a}}.

\bibitem[Vahdat \& Kautz(2020{\natexlab{b}})Vahdat and Kautz]{vahdat2020nvae}
Arash Vahdat and Jan Kautz.
\newblock {NVAE}: A deep hierarchical variational autoencoder.
\newblock In \emph{Advances in Neural Information Processing Systems
  ({NeurIPS})}, 2020{\natexlab{b}}.

\bibitem[van~den Oord et~al.(2017)van~den Oord, Vinyals, and
  Kavukcuoglu]{vanDenOord2017}
Aaron van~den Oord, Oriol Vinyals, and Koray Kavukcuoglu.
\newblock Neural discrete representation learning.
\newblock In \emph{Advances in Neural Information Processing Systems
  ({NeurIPS})}, 2017.

\bibitem[Vaswani et~al.(2017)Vaswani, Shazeer, Parmar, Uszkoreit, Jones, Gomez,
  Kaiser, and Polosukhin]{Vaswani2017AttentionIA}
Ashish Vaswani, Noam~M. Shazeer, Niki Parmar, Jakob Uszkoreit, Llion Jones,
  Aidan~N. Gomez, Lukasz Kaiser, and Illia Polosukhin.
\newblock Attention is all you need.
\newblock In \emph{Advances in Neural Information Processing Systems
  ({NeurIPS})}, 2017.

\bibitem[Ventola et~al.(2020)Ventola, Stelzner, Molina, and
  Kersting]{ventola2020residual}
Fabrizio Ventola, Karl Stelzner, Alejandro Molina, and Kristian Kersting.
\newblock Residual sum-product networks.
\newblock In \emph{European Workshop on Probabilistic Graphical Models}, 2020.

\bibitem[Ventola et~al.(2023{\natexlab{a}})Ventola, Braun, Yu, Mundt, and
  Kersting]{ventola2023probabilistic}
Fabrizio Ventola, Steven Braun, Zhongjie Yu, Martin Mundt, and Kristian
  Kersting.
\newblock Probabilistic circuits that know what they don't know.
\newblock In \emph{Uncertainty in Artificial Intelligence ({UAI})},
  2023{\natexlab{a}}.

\bibitem[Ventola et~al.(2023{\natexlab{b}})Ventola, Braun, Yu, Mundt, and
  Kersting]{ventola2023tdi}
Fabrizio~G. Ventola, Steven Braun, Zhongjie Yu, Martin Mundt, and Kristian
  Kersting.
\newblock Probabilistic circuits that know what they don't know.
\newblock In \emph{Uncertainty in Artificial Intelligence ({UAI})},
  2023{\natexlab{b}}.

\bibitem[Vergari et~al.(2018)Vergari, Peharz, Mauro, Molina, Kersting, and
  Esposito]{Vergari2018SumProductAE}
Antonio Vergari, Robert Peharz, Nicola~Di Mauro, Alejandro Molina, Kristian
  Kersting, and Floriana Esposito.
\newblock Sum-product autoencoding: Encoding and decoding representations using
  sum-product networks.
\newblock In \emph{Association for the Advancement of Artificial Intelligence
  ({AAAI})}, 2018.

\bibitem[Vergari et~al.(2021)Vergari, Choi, Liu, Teso, and den
  Broeck]{vergari2021compositional}
Antonio Vergari, YooJung Choi, Anji Liu, Stefano Teso, and Guy~Van den Broeck.
\newblock A compositional atlas of tractable circuit operations for
  probabilistic inference.
\newblock In \emph{Advances in Neural Information Processing Systems
  ({NeurIPS})}, 2021.

\bibitem[Vincent et~al.(2008)Vincent, Larochelle, Bengio, and
  Manzagol]{vincent2008extracting}
Pascal Vincent, Hugo Larochelle, Yoshua Bengio, and Pierre-Antoine Manzagol.
\newblock Extracting and composing robust features with denoising autoencoders.
\newblock In \emph{International Conference on Machine Learning ({ICML})},
  2008.

\bibitem[Wang \& Broeck(2024)Wang and Broeck]{wang2024relationship}
Benjie Wang and Guy Van~den Broeck.
\newblock On the relationship between monotone and squared probabilistic
  circuits.
\newblock \emph{arXiv preprint arXiv:2408.00876}, 2024.

\bibitem[Wang(2021{\natexlab{a}})]{Wang2021DataFreeKD}
Z.~Wang.
\newblock Data-free knowledge distillation with soft targeted transfer set
  synthesis.
\newblock In \emph{Association for the Advancement of Artificial Intelligence
  ({AAAI})}, 2021{\natexlab{a}}.

\bibitem[Wang(2021{\natexlab{b}})]{wang2021zskdb}
Zi~Wang.
\newblock Zero-shot knowledge distillation from a decision-based black-box
  model.
\newblock In \emph{International Conference on Machine Learning ({ICML})},
  2021{\natexlab{b}}.

\bibitem[Xiao et~al.(2017)Xiao, Rasul, and Vollgraf]{xiao2017fmnist}
Han Xiao, Kashif Rasul, and Roland Vollgraf.
\newblock Fashion-mnist: a novel image dataset for benchmarking machine
  learning algorithms.
\newblock \emph{arXiv preprint arXiv:1708.07747}, 2017.

\bibitem[Yang et~al.(2023)Yang, Gala, and Peharz]{yang23bayesian}
Yang Yang, Gennaro Gala, and Robert Peharz.
\newblock Bayesian structure scores for probabilistic circuits.
\newblock In \emph{International Conference on Artificial Intelligence and
  Statistics ({AISTATS})}, 2023.

\bibitem[Yim et~al.(2017)Yim, Joo, Bae, and Kim]{yim2017agf}
Junho Yim, Donggyu Joo, Ji{-}Hoon Bae, and Junmo Kim.
\newblock A gift from knowledge distillation: Fast optimization, network
  minimization and transfer learning.
\newblock In \emph{Conference on Computer Vision and Pattern Recognition
  ({CVPR})}, 2017.

\bibitem[Yin et~al.(2020)Yin, Molchanov, Alvarez, Li, Mallya, Hoiem, Jha, and
  Kautz]{hongxu2019deepinversion}
Hongxu Yin, Pavlo Molchanov, Jose~M. Alvarez, Zhizhong Li, Arun Mallya, Derek
  Hoiem, Niraj~K. Jha, and Jan Kautz.
\newblock Dreaming to distill: Data-free knowledge transfer via deepinversion.
\newblock In \emph{Conference on Computer Vision and Pattern Recognition
  ({CVPR})}, 2020.

\bibitem[Yoon et~al.(2018)Yoon, Jordon, and van~der Schaar]{Yoon2018GAINMD}
Jinsung Yoon, James Jordon, and Mihaela van~der Schaar.
\newblock {GAIN}: Missing data imputation using generative adversarial nets.
\newblock In \emph{International Conference on Machine Learning ({ICML})},
  2018.

\bibitem[Yu et~al.(2015)Yu, Zhang, Song, Seff, and Xiao]{lsun}
Fisher Yu, Yinda Zhang, Shuran Song, Ari Seff, and Jianxiong Xiao.
\newblock {LSUN:} construction of a large-scale image dataset using deep
  learning with humans in the loop.
\newblock \emph{arXiv preprint arXiv:1506.03365}, 2015.

\bibitem[Yu et~al.(2021)Yu, Zhu, Trapp, Skryagin, and
  Kersting]{yu2021uai_momogps}
Zhongjie Yu, Mingye Zhu, Martin Trapp, Arseny Skryagin, and Kristian Kersting.
\newblock Leveraging probabilistic circuits for nonparametric multi-output
  regression.
\newblock In \emph{Uncertainty in Artificial Intelligence ({UAI})}, 2021.

\bibitem[Yu et~al.(2022)Yu, Dhami, and Kersting]{yu2022sumproductattention}
Zhongjie Yu, Devendra~Singh Dhami, and Kristian Kersting.
\newblock Sum-product-attention networks: Leveraging self-attention in
  energy-based probabilistic circuits.
\newblock In \emph{The 5th Workshop on Tractable Probabilistic Modeling}, 2022.

\bibitem[Yu et~al.(2024)Yu, Trapp, and Kersting]{yu2024characteristic}
Zhongjie Yu, Martin Trapp, and Kristian Kersting.
\newblock Characteristic circuits.
\newblock In \emph{Advances in Neural Information Processing Systems
  ({NeurIPS})}, 2024.

\bibitem[Zhang et~al.(2023)Zhang, Dang, Peng, and den
  Broeck]{zhang2023tractable}
Honghua Zhang, Meihua Dang, Nanyun Peng, and Guy~Van den Broeck.
\newblock Tractable control for autoregressive language generation.
\newblock In \emph{International Conference on Machine Learning ({ICML})},
  2023.

\bibitem[Zheng et~al.(2018)Zheng, Pronobis, and Rao]{zheng2018graph}
Kaiyu Zheng, Andrzej Pronobis, and Rajesh P.~N. Rao.
\newblock Learning graph-structured sum-product networks for probabilistic
  semantic maps.
\newblock In \emph{Association for the Advancement of Artificial Intelligence
  ({AAAI})}, 2018.

\bibitem[Zhou et~al.(2018)Zhou, Wu, Ni, Zhou, Wen, and
  Zou]{zhou2018dorefanet-tl}
Shuchang Zhou, Yuxin Wu, Zekun Ni, Xinyu Zhou, He~Wen, and Yuheng Zou.
\newblock {DoReFa-Net}: Training low bitwidth convolutional neural networks
  with low bitwidth gradients.
\newblock \emph{arXiv preprint arXiv:1606.06160}, 2018.

\end{thebibliography}
\bibliographystyle{tmlr}

\cleardoublepage
\appendix
\crefalias{figure}{appfig}
\crefalias{table}{apptab}
\crefalias{algorithm}{appalg}

\section{Encoding Algorithm}

\begin{algorithm}[H]%
  \caption{APC Encoding Procedure}
  \label{alg:encoding}
  \begin{algorithmic}[1]
    \Require Probabilistic circuit $\gC$ over data $\X$ and embeddings $\Z$, an input data point $\x \in \X$

    \State

    \Procedure{encode}{$\gC$: circuit, $\x$: data}
    \State $\textsc{forward}(\textsc{root}(\gC), \x)$ \Comment{Forward pass; at each sum unit we store its inputs}
    \State $\bigl[\x, \z\bigr] = \textsc{sample}(\textsc{root}(\gC))$ \Comment{Sample from conditioned PC $\left.\gC\right\rvert_{\X=\x}$}
    \State \textbf{return} $\z$
    \EndProcedure

    \State

    \Procedure{forward}{$n$: unit, $\x$: data} \Comment{General forward pass for circuit units}

    \If{$n == \inputunit$}
      \If{$\x_n$ is missing}
      \State \textbf{return} $1.0$  \Comment{Marginalize missing inputs}
      \Else
      \State \textbf{return} $p_n(\x_n)$ \Comment{Evaluate PDF of input units}
      \EndIf
    \EndIf

    \State $\bm{\gamma} \gets \bigl[c(\x_c) \text{ for } c \in \inscope(n)\bigr]$  \Comment{Collect inputs for sum/product units}
    \State $n.\bm{\gamma} \gets \bm{\gamma}$  \Comment{Cache inputs for conditional sampling pass}

    \If{$n == \bigotimes$}
      \State \textbf{return} $\prod_{c \in \inscope(n)} \bm{\gamma}_c$ \Comment{Product unit}
    \EndIf

    \If{$n == \bigoplus$}
      \State \textbf{return} $\sum_{c \in \inscope(n)} \bm{\gamma}_c \cdot \theta_n^c$  \Comment{Sum unit}
    \EndIf
    \EndProcedure

    \State

    \Procedure{sample}{$n$: unit} \Comment{Sampling pass for circuit units}

    \If{$n == \inputunit$}
        \State \textbf{return} $\x_n \sim p_n(\X_n)$  \Comment{Sample from PDF represented by input unit}
    \EndIf

    \If{$n == \bigotimes$}
      \State \textbf{return} $\textsc{concat}(\bigl[ \textsc{sample}(c) \text{ for } c \in \inscope(n) \bigr])$ \Comment{Product unit simply sample all inputs}
    \EndIf

    \If{$n == \bigoplus$}
      \State $\bm{\theta}_n' \gets \textsc{condition\_weights}(\bm{\theta}_n, n.\bm{\gamma})$  \Comment{Condition weights on forward pass likelihoods}
      \State $\mathbf{s} \gets \textsc{simple}(\bm{\theta}_n')$ \Comment{Sample one-hot encoded index}
      \State \textbf{return} $\textsc{dot}(\mathbf{s}, \bigl[ \textsc{sample}(c) \text{ for } c \in \inscope(n) \bigr])$ \Comment{Dot product indexes input samples}
    \EndIf
    \EndProcedure

    \State

    \Procedure{condition\_weights}{$\bm{\theta}$: weights, $\bm{\gamma}$: inputs}
    \For{$i \in 1 \dots \textsc{size}(\bm{\theta})$}
      \State $\theta_{i} \gets \theta_{c} \cdot \gamma_i$  \Comment{Reweight weights based on input likelihoods}
    \EndFor

    \State $s = \sum_{i} \theta_{i}$  \Comment{Compute normalization constant}

    \For{$i \in 1 \dots \textsc{size}(\bm{\theta})$}
      \State $\theta_{i} \gets \theta_{i} / s$  \Comment{Normalize weight distribution}
    \EndFor
    \State \textbf{return} $\bm{\theta}$
    \EndProcedure
  \end{algorithmic}
\end{algorithm}

Conditional sampling in PCs, which forms the basis of our encoding procedure (\cref{alg:encoding}), consists of two passes through the
circuit. First, a forward pass computes the marginal likelihood of the evidence, $\fn{p_{\gC}}{\x}$, for a given data
sample $\x$. During this pass, likelihoods $\bm{\gamma}$ are cached at the inputs of each sum unit. Subsequently, a
sampling pass traverses the circuit from the root to the input units. At each sum unit, it uses the cached likelihoods
$\bm{\gamma}$ from the forward pass to reweight the mixture parameters, effectively applying Bayes' rule to form a
posterior over its inputs. It then samples a path according to these conditioned weights using the differentiable SIMPLE
estimator. This process yields a sample from the conditional distribution $\fn{p_{\gC}}{\Z \cbar \X = \x}$, which serves
as the embedding $\z$.

\section{Experiment and Evaluation Protocol: Additional Details }
\label{app:eval-protocol}

All experiments and models are implemented in PyTorch~\citep{Ansel_PyTorch_2_Faster_2024} and PyTorch
Lightning~\citep{Falcon_PyTorch_Lightning_2019}. Our implementation is available as open-source software at
\url{https://github.com/placeholder-url}.

\paragraph{Datasets} We evaluate our models on various image and tabular datasets. For image data, we include
MNIST \citep{lecun1998mnist}, Fashion MNIST (F-MNIST) \citep{xiao2017fmnist}, CIFAR-10 \citep{cifar100}, CelebA
\citep{liu2015faceattributes}, SVHN \citep{svhn}, Flowers \citep{gurnani2017flowers}, LSUN (Church) \citep{lsun}, and
Tiny-ImageNet \citep{imagenet}. Note that Tiny-ImageNet is occasionally abbreviated as ImageNet in our tables for
brevity. For tabular data, we utilize the 20 datasets from the binary density estimation benchmark DEBD
\citep{Lowd2010LearningMN,Haaren2012MarkovNS,Bekker2015TractableLF,Larochelle2011TheNA}.

\paragraph{Models} In our empirical evaluation, we compare the proposed \methods against
several models to demonstrate their effectiveness. These
include SPAE-ACT and SPAE-CAT~\citep{vergari2018spae} as a circuit-based comparison, as well as vanilla variational autoencoder (VAE)
and, to also include more missing-data specific methods MIWAE~\citep{mattei2019miwae} and missForest~\citep{stekhoven2011missForest}. For all autoencoding models, we use an embedding dimension $d$ of 64 for MNIST and F-MNIST, 256 for all other image-based
datasets, and 4,8,16,32, and 64 for tabular datasets, depending on their number of features to ensure $d \ll |\X|$. All
models are trained under the same conditions with common and well-established practices.

\paragraph{Metrics}
We report MSE ($\downarrow$) to measure reconstruction fidelity and SSIM ($\uparrow$) to assess perceptual quality based on structural and visual similarity. To assess encoding
robustness, we analyze MSE and SSIM across varying levels of MCAR-style input data corruption, where uniformly random corruption is
incrementally increased from 0\% to 95\% in steps of 5\%. We then compute the average reconstruction error over all
corruption levels for each metric. To ensure that results are not specific to MCAR-style
missing data, \cref{tab:mar-corruptions} presents evaluations using nine additional MAR-style corruption variants.

\paragraph{Missing Data}
For vanilla VAEs, missing values are imputed using a constant zero value. We also explored alternative
imputation strategies, including mean imputation and learned normalization (similar to LayerNorm). In our experiments on
MNIST, mean imputation offered improved reconstruction quality at the cost of a reduction in downstream task accuracy.
For more challenging datasets like SVHN and CIFAR, the choice of imputation method did not yield statistically
significant differences in performance. All other methods handle missing data without the need for imputation.

\paragraph{Model Capacity}
For all experiments, we ensure that the model encoder and decoder networks are approximately the same size and depth
across all models. We use the same neural decoder architecture across APCs and VAE models to ensure that performance differences can be solely attributed to the encoder.

\paragraph{Circuit Input Units}
For circuit-based encoders, we model data distributions on images with Binomial input units, and Bernoulli input units for the
DEBD tabular datasets. Embedding distributions are chosen as Gaussian input units in all cases. We want to highlight once
more that the \method framework, in principle, allows for the choice of arbitrary data and \emph{embedding} distributions.

\paragraph{Sum-Product Autoencoding}
For SPAE, we retrieve the correctly sized embedding vector from a layer in the circuit graph, which consists of exactly
$d$ units (embedding size). In addition, when evaluating reconstructions, we report results only for SPAE (instead of SPAE-ACT and
SPAE-CAT) because the reconstruction output is identical for both SPAE variants, regardless of whether activation
embeddings (ACT) or categorical embeddings (CAT) are used. This equivalence is formally established in Proposition 3 of
\citet{vergari2018spae}. However, in all other evaluations, we distinguish between SPAE-ACT and SPAE-CAT, since
their embeddings are inherently different when used in e.g., downstream tasks.

\paragraph{Autoencoder Training} Each model is trained for 10,000 iterations using the AdamW optimizer
\citep{kingma2015adam,loshchilov2017adamw}, and convergence was confirmed for all models by the end of this training
period. We use the MSE as $\mathcal{L}_{\text{REC}}$ for all models. Training is carried out with a
batch size of 512, except for CelebA where a batch size of 256 is used due to larger model sizes and VRAM constraints.
The initial learning rate is set to 0.1 for \methods and 0.005 for AEs and VAEs. The learning rate is reduced by a
factor of 10 at 66\% and 90\% of the training progress. Additionally, to enhance stability and avoid numerical issues or
exploding gradients during the training phase, we utilize an exponential learning rate warmup over the first 2\% of
training iterations. Empirically, this approach mitigates random training difficulties without affecting the final
performance across different random seeds. While, for \methods, we could potentially pretrain the PC-encoder with MLE on
the marginal $\fn{p}{\X}$ to speed up convergence, we refrain from doing so in our experiments, to keep the comparison
between APC and VAE-based models as fair as possible. All models, with the exception of MIWAE, were trained on the
complete datasets without missing values. For MIWAE, training was performed using MCAR-style data in which 50\% of
entries were missing.

\paragraph{Downstream Task Training}
We train the logistic regression downstream task models with a batch size of 512 for 5,000 iterations on dataset
embeddings from the respective encoders. We employed an initial learning rate of 0.05, which was reduced by a factor of
0.1 at 66\% and 90\% of the training progress, along with an exponential learning rate warmup over the first 2\% of
iterations. For SPAE-ACT models specifically, we found it necessary to use the AdamW optimizer with a learning rate of
0.01 for 100,0000 iterations (20$\times$ longer) and normalize the embeddings to achieve comparable downstream task
performance better than a random guessing baseline.

\subsection{Model Architectures}
\label{app:eval-protocol:model-architectures}

\paragraph{Neural Encoder and Decoder.} For tabular data, we employ a simple
linear feed-forward style network with four hidden layers and leaky ReLU
($\alpha= 0.1$) activations as encoder and decoder. We use convolutional and residual layers with ReLU activations instead for image data. For further
details, we refer the reader to
\texttt{apc.models.\{encoder,decoder\}.nn\_\{encoder,decoder\}.py} files in our
source code repository.

\paragraph{EinsumNetwork Encoder for Tabular Data.} We use
EinsumNetworks~\citep{peharz2020einet} as circuit structure for \methods and SPAEs with Bernoulli
input units for the tabular DEBD
dataset~\citep{Lowd2010LearningMN,Haaren2012MarkovNS,Bekker2015TractableLF,Larochelle2011TheNA}.
For \methods, embedding input units are inserted at the lowest layer and randomly
shuffled with the data distribution input units. For all experiments, we keep a depth
of 4, with a single repetition and 32 input unit distributions, and the sum units per
scope at each layer.

\paragraph{ConvPc Encoder for Image Data.} For image data, we construct a simple ``convolutional'' circuit for \methods and SPAEs, where we
first map each pixel to its density represented by an input unit layer with 256 output channels (number of input unit
distributions per scope). We then successively reduce the height and width by a factor of 2 with product layers, which
builds the product over all scopes in disjoint neighboring windows (think of non-overlapping convolution windows with
stride being the window size). After each product layer, a sum layer maps all channels of each scope to a vector of sum
unit outputs (think of convolution in- and out-channels). We repeat this until we end up with a single dimension where
the scopes cover the full image input. This allows us to choose a certain depth (how often do we want to repeat the
product-sum combination), and the choice of number of sum output units per sum layer. Embedding input units are inserted
randomly at the lowest layer with product units attached to $|\Z|$ data input units, constructing $|\Z|$ combinations of
$\fn{p_{\gC}}{x_{i}}\fn{p_{\gC}}{z_{j}}$. More advanced strategies for embedding insertions remain an avenue for future
investigation; for example, embedding input units could be decoupled and integrated at varying hierarchical levels within the
circuit.

\cleardoublepage

\section{Reconstruction Performance: Additional Results}
\label{app:rec-additional-results}

This section provides a more detailed exposition of the reconstruction capabilities of \methods and the compared models,
supplementing the empirical evaluations presented in the main body. We include additional quantitative results, such as
SSIM for image datasets, further explore the impact of various MAR corruption patterns beyond the MCAR scenarios discussed earlier, and present
extended qualitative results through more comprehensive sets of reconstruction visualizations across all datasets and
corruption types. Furthermore, we offer an initial exploration into the potential of APC embeddings for
out-of-distribution detection by analyzing embedding likelihoods.

\subsection{Reconstructions: Comparing Circuit and Neural Encoder (Quantitative Results)}

\begin{table}[t]
  \caption{%
\textbf{\methods achieve lower reconstruction errors than PC, VAE, and missForest baselines across varying levels of MCAR data}.
  Average reconstruction performance on eight image datasets under increasing percentages of MCAR-style
    randomly missing pixels (0\%-95\%).
    }
  \label{tab:reconstruction-mse}
  \begin{center}
    \begin{tabular}{lrrrrr}

    \maybetoprule
                                     & \header{\method}    & \header{SPAE}       & \header{VAE}        & \header{MIWAE}     & \header{missForest} \\
    \midrule
MNIST~\citep{lecun1998mnist}         & \resB{5.07}{0.04}   & \res{28.46}{0.43}   & \res{35.38}{1.74}   & \res{9.16}{0.92}   & \res{24.51}{0.63}   \\
F-MNIST~\citep{xiao2017fmnist}       & \resB{4.33}{0.02}   & \res{18.96}{0.11}   & \res{39.80}{1.10}   & \res{12.54}{0.74}  & \res{24.53}{0.02}   \\
CIFAR~\citep{cifar100}               & \resB{20.65}{0.09}  & \res{78.32}{3.09}   & \res{55.81}{0.47}   & \res{32.80}{0.27}  & \res{57.61}{0.16}   \\
CelebA~\citep{liu2015faceattributes} & \resB{166.89}{0.65} & \res{1318.73}{6.37} & \res{1095.90}{14.0} & \res{576.47}{24.8} & \res{1855.67}{2.29}      \\
SVHN~\citep{svhn}                    & \resB{4.87}{0.00}   & \res{35.38}{1.92}   & \res{41.01}{0.15}   & \res{17.28}{0.29}  & \res{44.19}{0.25}   \\
Flowers~\citep{gurnani2017flowers}   & \resB{45.22}{0.14}  & \res{119.31}{5.23}  & \res{107.75}{11.8}  & \res{58.03}{0.37}  & \res{52.65}{0.14}   \\
LSUN~\citep{lsun}                    & \resB{63.70}{0.20}  & \res{322.22}{3.48}  & \res{254.75}{2.39}  & \res{123.58}{1.40} & \res{115.28}{0.15}  \\
ImageNet~\citep{imagenet}            & \resB{118.70}{0.16} & \res{365.06}{2.20}  & \res{1286.51}{10.4} & \res{204.10}{2.04} & \res{295.40}{0.34}  \\

    \maybebottomrule
\end{tabular}

  \end{center}
\end{table}

\begin{table}[t]
  \caption{%
    \textbf{\methods achieve lowest reconstruction error on 18 out of 20 tabular dataset.}
    \methods outperform the PC, VAE and missForest baselines on the 20 DEBD binary tabular dataset and achieve the best reconstruction performance on 18 of the 20 datasets under increasing percentages of MCAR-style
    randomly missing data (0\%-95\%).}
  \label{tab:reconstruction-debd}
  \begin{center}
    \tablesize     \begin{tabular}{lrrrrr}
      \maybetoprule
            & \header{\method}   & \header{SPAE}      & \header{VAE}       & \header{MIWAE}     & \header{missForest} \\
      \midrule
accidents   & \resB{6.16}{0.04}  & \res{12.55}{0.86}  & \res{7.56}{0.28}   & \res{7.34}{0.05}  & \res{6.75}{0.01}   \\
ad          & \res{5.57}{0.03}   & \res{231.84}{15.4} & \res{6.04}{0.17}   & \resB{3.50}{0.06} & \res{5.57}{0.03}   \\
baudio      & \resB{6.50}{0.03}  & \res{15.60}{1.33}  & \res{7.44}{0.01}   & \res{7.81}{0.04}  & \res{7.50}{0.01}   \\
bbc         & \resB{34.66}{0.15} & \res{161.59}{10.8} & \res{37.45}{0.55}  & \res{47.08}{0.60} & \res{35.02}{0.15}  \\
bnetflix    & \resB{10.08}{0.02} & \res{20.53}{0.96}  & \res{13.49}{0.77}  & \res{14.81}{0.04} & \res{10.80}{0.00}  \\
book        & \resB{3.65}{0.02}  & \res{67.10}{9.40}  & \res{3.73}{0.02}   & \res{4.40}{0.06}  & \res{3.87}{0.03}   \\
c20ng       & \resB{19.12}{0.04} & \res{123.50}{5.85} & \res{20.69}{0.05}  & \res{25.50}{0.28} & \res{20.34}{0.02}  \\
cr52        & \resB{11.94}{0.10} & \res{134.30}{7.81} & \res{12.83}{0.18}  & \res{15.10}{0.19} & \res{12.75}{0.09}  \\
cwebkb      & \resB{21.32}{0.14} & \res{109.07}{4.38} & \res{22.94}{0.15}  & \res{28.53}{0.32} & \res{22.71}{0.07}  \\
dna         & \resB{15.89}{0.12} & \res{28.76}{0.91}  & \res{16.46}{0.17}  & \res{20.46}{0.30} & \resB{15.89}{0.04} \\
jester      & \resB{9.04}{0.02}  & \res{20.98}{0.33}  & \res{17.44}{0.10}  & \res{15.04}{0.08} & \res{10.66}{0.01}  \\
kdd         & \resB{0.19}{0.00}  & \res{4.12}{0.57}   & \res{0.19}{0.00}   & \res{0.22}{0.00}  & \res{0.20}{0.00}   \\
kosarek     & \resB{1.35}{0.04}  & \res{13.57}{3.63}  & \res{1.42}{0.01}   & \res{1.67}{0.01}  & \res{1.40}{0.00}   \\
moviereview & \resB{48.06}{0.15} & \res{141.19}{4.27} & \res{50.08}{0.22}  & \res{70.07}{0.13} & \res{48.36}{0.06}  \\
msnbc       & \res{1.05}{0.03}   & \res{2.23}{0.39}   & \res{1.03}{0.01}   & \res{1.17}{0.00}  & \resB{0.99}{0.00}  \\
nltcs       & \resB{1.12}{0.02}  & \res{2.48}{0.00}   & \res{1.60}{0.01}   & \res{1.37}{0.01}  & \res{1.48}{0.00}   \\
plants      & \resB{2.52}{0.04}  & \res{11.62}{0.54}  & \res{3.93}{0.05}   & \res{2.93}{0.01}  & \res{4.67}{0.01}   \\
pumsb\_star & \resB{5.82}{0.16}  & \res{23.30}{0.74}  & \res{12.47}{0.40}  & \res{6.53}{0.10}  & \res{11.74}{0.02}  \\
tmovie      & \resB{7.32}{0.06}  & \res{57.94}{5.49}  & \res{8.72}{0.03}   & \res{9.04}{0.04}  & \res{10.11}{0.02}  \\
tretail     & \resB{1.22}{0.01}  & \res{6.22}{1.63}   & \res{1.23}{0.00}   & \res{1.58}{0.02}  & \resB{1.22}{0.00}  \\
voting      & \resB{27.71}{0.18} & \res{264.53}{7.30} & \res{122.24}{52.8} & \res{39.04}{0.46} & \res{119.03}{0.12}

      \maybebottomrule
    \end{tabular}

  \end{center}
\end{table}

In \cref{sec:eval:reconstruction}, we visually demonstrated the robustness of \methods against increasing data
corruption by plotting reconstruction error against the percentage of missing values (cf. \cref{fig:rec-missing-graph}).
To supplement this visual analysis with a direct numerical comparison, we additionally provide detailed quantitative results.
\cref{tab:reconstruction-mse} presents the average MSE across all MCAR corruption
levels for the image datasets, numerically aggregating and substantiating the trends observed in the plots. We further extend this
quantitative analysis to tabular data to demonstrate the generalizability of our findings.
\cref{tab:reconstruction-debd} provides the corresponding reconstruction performance on the 20 DEBD datasets.

\subsection{Missing at Random (MAR) Style Corruptions}

\begin{table}[H]
  \caption{\textbf{\methods ranks first on all MAR-style corruption types across all datasets.} Model ranking per corruption type, aggregated across datasets. Each row represents a different MAR-style corruption
    applied to the input images. The severity of each corruption is linearly increased, and the average reconstruction
    MSE/SSIM is measured. The models are ranked based on their robustness to these corruptions aggregated across all
    image datasets. The reported values represent the ranking based on the average MSE/SSIM. See
    \cref{app:fig:rec-examples-B,app:fig:rec-examples-D} for reconstruction visualizations.}
  \label{tab:mar-corruptions}
  \setlength{\tabcolsep}{3.0pt}
  \begin{center}
    \tablesize 

\begin{tabular}{lc p{0.5em} *{4}{c} p{0.5em} *{4}{c}}
  \maybetoprule
                                 & &       & \multicolumn{4}{c}{MSE}                                                       & & \multicolumn{4}{c}{SSIM} \\
                                   \cmidrule(lr){4-7} \cmidrule(lr){9-12}
Corruption      &   &   & \header{\method} & \header{SPAE} & \header{VAE} & \header{MIWAE} &   & \header{\method} & \header{SPAE} & \header{VAE} & \header{MIWAE} \\
    \midrule
left-to-right    &  &  & \textbf{1} & 4 & 2 & 3 &  & \textbf{1} & 3 & 2 & 4 \\
right-to-left    &  &  & \textbf{1} & 4 & 2 & 3 &  & \textbf{1} & 2 & 3 & 4 \\
top-to-bottom    &  &  & \textbf{1} & 4 & 2 & 3 &  & \textbf{1} & 3 & 2 & 4 \\
bottom-to-top    &  &  & \textbf{1} & 4 & 2 & 3 &  & \textbf{1} & 3 & 2 & 4 \\
border-to-center &  &  & \textbf{1} & 3 & 2 & 4 &  & \textbf{1} & 2 & 3 & 4 \\
center-to-border &  &  & \textbf{1} & 3 & 4 & 2 &  & \textbf{1} & 2 & 3 & 4 \\
horizontal-band  &  &  & \textbf{1} & 3 & 4 & 2 &  & \textbf{1} & 4 & 2 & 3 \\
vertical-band    &  &  & \textbf{1} & 2 & 3 & 4 &  & \textbf{1} & 3 & 2 & 4 \\
salt-and-pepper  &  &  & \textbf{1} & 3 & 2 & 4 &  & \textbf{1} & 4 & 2 & 3 \\
  \maybebottomrule
\end{tabular}

  \end{center}
\end{table}

To further investigate model robustness beyond \emph{randomly} missing data, \cref{tab:mar-corruptions} presents model
rankings across different MAR-style corruption types aggregated over all datasets. For each corruption type, we repeat
the experiments presented in \cref{tab:reconstruction-mse} for all image datasets, assign each method a rank from best
(1.) to worst (4.) and then report their final rank, sorted by the average rank over all datasets. We only
compare autoencoding methods, as they reconstruct the full image from a low-dimensional latent space. In contrast,
missForest operates directly in the input space, making it an inappropriate comparison for a ranking in this context.
\methods consistently achieve the top rank for all but one corruption type and metric, indicating that their superior
performance extends beyond handling missing data to various other forms of data corruption. Vanilla VAEs appear to
outrank its missing-data-specific variant MIWAE in both MSE and SSIM measurements. This could be attributed to the fact
that MIWAE specifically trains on a specific type of missing data, MCAR in our case. This leads to a decreased
performance in reconstruction when the pattern of missing data substantially changes. We can visually confirm this is
\cref{app:fig:rec-examples-C,app:fig:rec-examples-D}, where MIWAE is better at reconstructing MCAR-style missing data
than VAEs, but fails similarly when confronted with MAR-style corruptions (horizontal, vertical, center, border).

\subsection{Structured Similarity Index (SSIM) Evaluations}
\label{app:images-mcar-ssim}

\begin{table}[t]
  \caption{%
\textbf{\methods achieve higher SSIM scores compared to PC, VAE, and missForest baselines across varying levels of MCAR data}.
  Average reconstruction SSIM performance on eight image datasets under increasing percentages of MCAR-style
    randomly missing pixels (0\%-95\%).
    }

  \label{app:tab:reconstruction-ssim}

  \begin{center}
    \tablesize \begin{tabular}{lrrrrr}
    \maybetoprule

         & \header{\method}   & \header{SPAE}     & \header{VAE}      & \header{MIWAE}    & \header{missForest} \\
    \midrule
MNIST    & \resB{93.33}{0.04} & \res{69.87}{0.36} & \res{50.74}{0.82} & \res{86.81}{1.14} & \res{62.29}{0.46}   \\
F-MNIST  & \resB{87.40}{0.05} & \res{68.89}{0.23} & \res{51.75}{0.38} & \res{78.24}{0.43} & \res{54.00}{0.04}   \\
CIFAR    & \resB{78.20}{0.07} & \res{50.61}{0.88} & \res{61.64}{0.14} & \res{74.25}{0.15} & \res{57.98}{0.13}   \\
CelebA   & \resB{79.52}{0.04} & \res{49.02}{0.17} & \res{50.30}{0.30} & \res{61.26}{0.66} & \res{37.16}{0.10}      \\
SVHN     & \resB{90.29}{0.03} & \res{54.88}{0.91} & \res{61.79}{0.18} & \res{76.23}{0.47} & \res{56.86}{0.17}   \\
Flowers  & \resB{60.40}{0.16} & \res{44.01}{0.82} & \res{50.63}{0.43} & \res{56.68}{0.11} & \res{56.94}{0.10}   \\
LSUN     & \resB{77.14}{0.07} & \res{50.73}{0.52} & \res{55.63}{0.28} & \res{59.80}{0.16} & \res{27.32}{0.02}   \\
ImageNet & \resB{71.55}{0.04} & \res{50.62}{0.36} & \res{47.76}{0.23} & \res{61.58}{0.15} & \res{50.98}{0.05}

    \maybebottomrule
\end{tabular}

  \end{center}
\end{table}

\begin{figure}[t]
  \begin{center}
    \includegraphics[width=\textwidth]{ 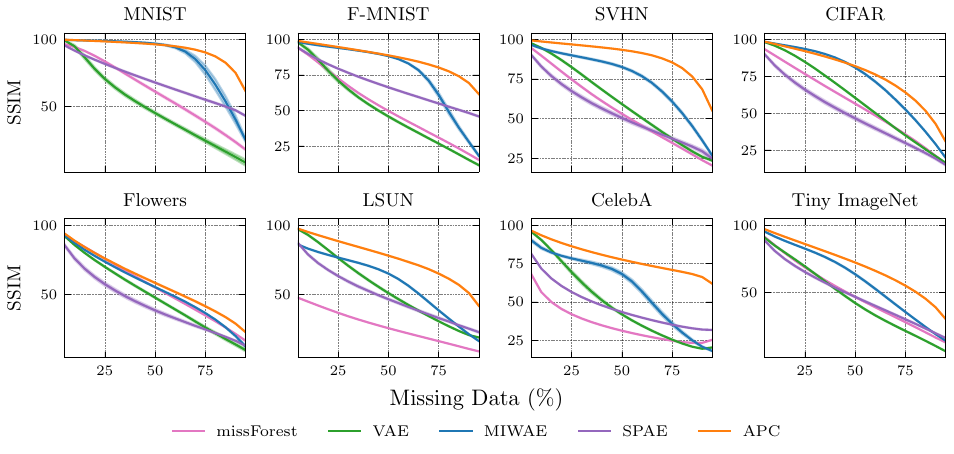 }
  \end{center}
  \caption{%
    \textbf{\methods can deliver higher reconstruction SSIM scores compared to PC, VAE, and missForest baselines in the asymptotic regime of MCAR-style randomly missing data}. Reconstruction SSIM is reported across all image datasets as the degree of MCAR corruption increases from 0\% to 95\%.
  }
  \label{app:fig:rec-missing-graph-ssim}
\end{figure}

While MSE provides a quantitative measure of pixel-wise differences, SSIM is also employed to offer a complementary perspective on reconstruction quality. SSIM is designed to better align with human visual perception by evaluating luminance, contrast, and structure similarities between the original and reconstructed images. Consistent with the MSE results, the findings presented in \cref{app:tab:reconstruction-ssim} (complementary to \cref{tab:reconstruction-mse}) and \cref{app:fig:rec-missing-graph-ssim} (complementary to \cref{fig:rec-missing-graph}) demonstrate that \methods significantly outperform all other compared methods in preserving image fidelity across all datasets, even as the proportion of missing data increases.

\subsection{Additional Reconstruction Visualizations}

\begin{figure}[t]
  \centering
  \tablesize \includegraphics[width=\textwidth]{ 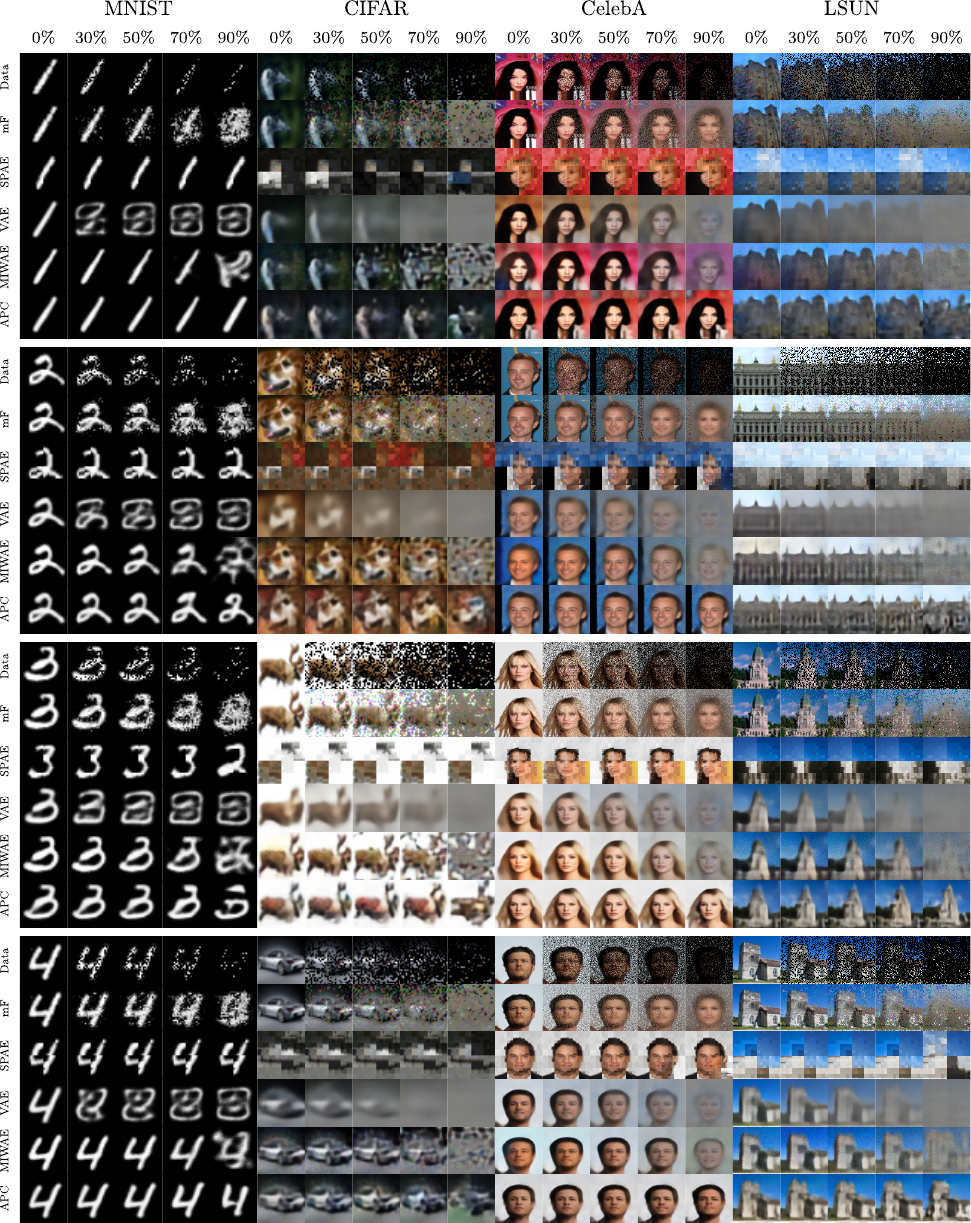 }
  \caption{Reconstructions of various MCAR corruptions for MNIST, CIFAR, CelebA, and LSUN.}
  \label{app:fig:rec-examples-A}
\end{figure}

\begin{figure}[t]
  \centering
  \tablesize \includegraphics[width=\textwidth]{ 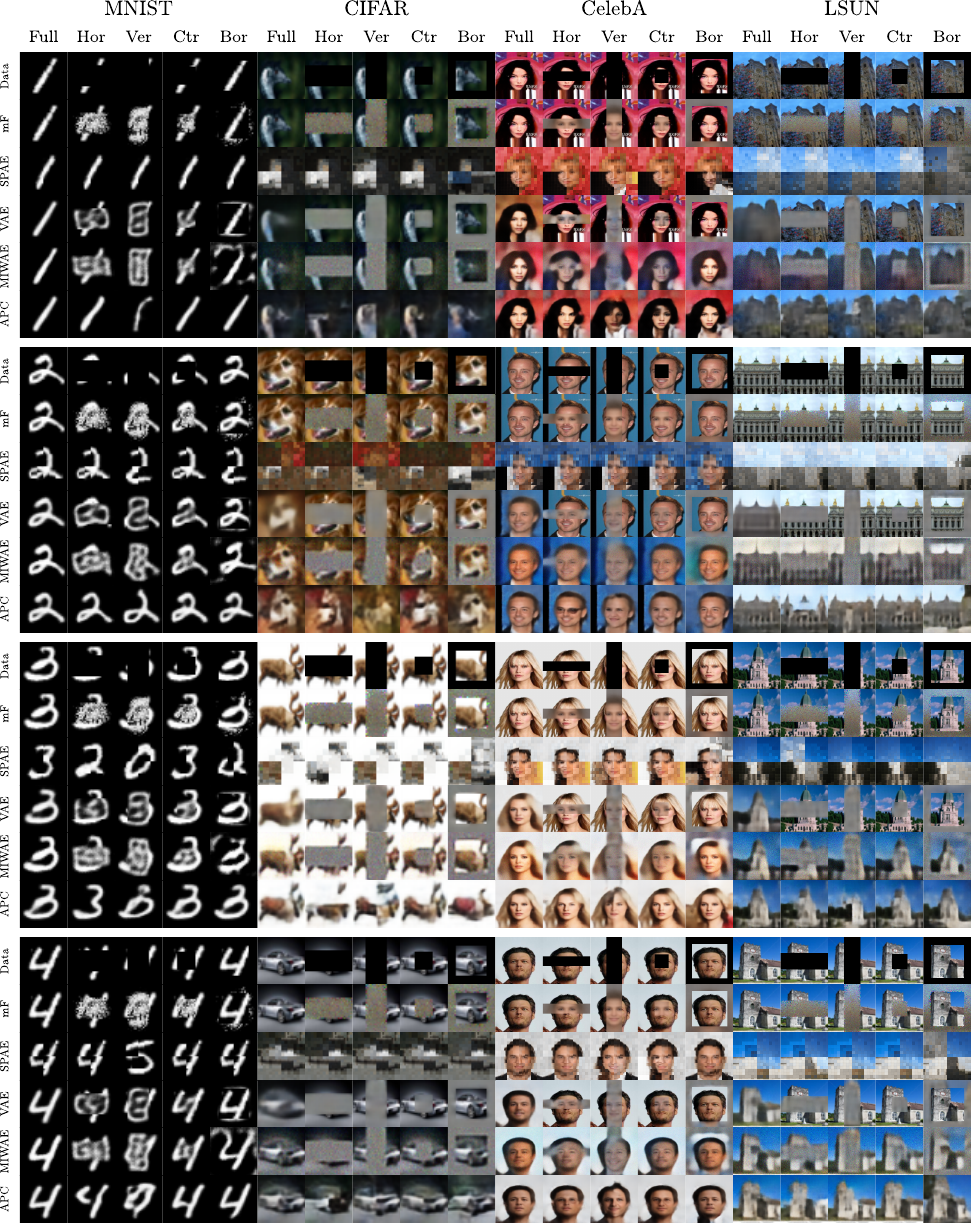 }
  \caption{Reconstructions of various MAR corruptions for MNIST, CIFAR, CelebA, and LSUN.}
  \label{app:fig:rec-examples-C}
\end{figure}

\begin{figure}[t]
  \centering
  \tablesize \includegraphics[width=\textwidth]{ 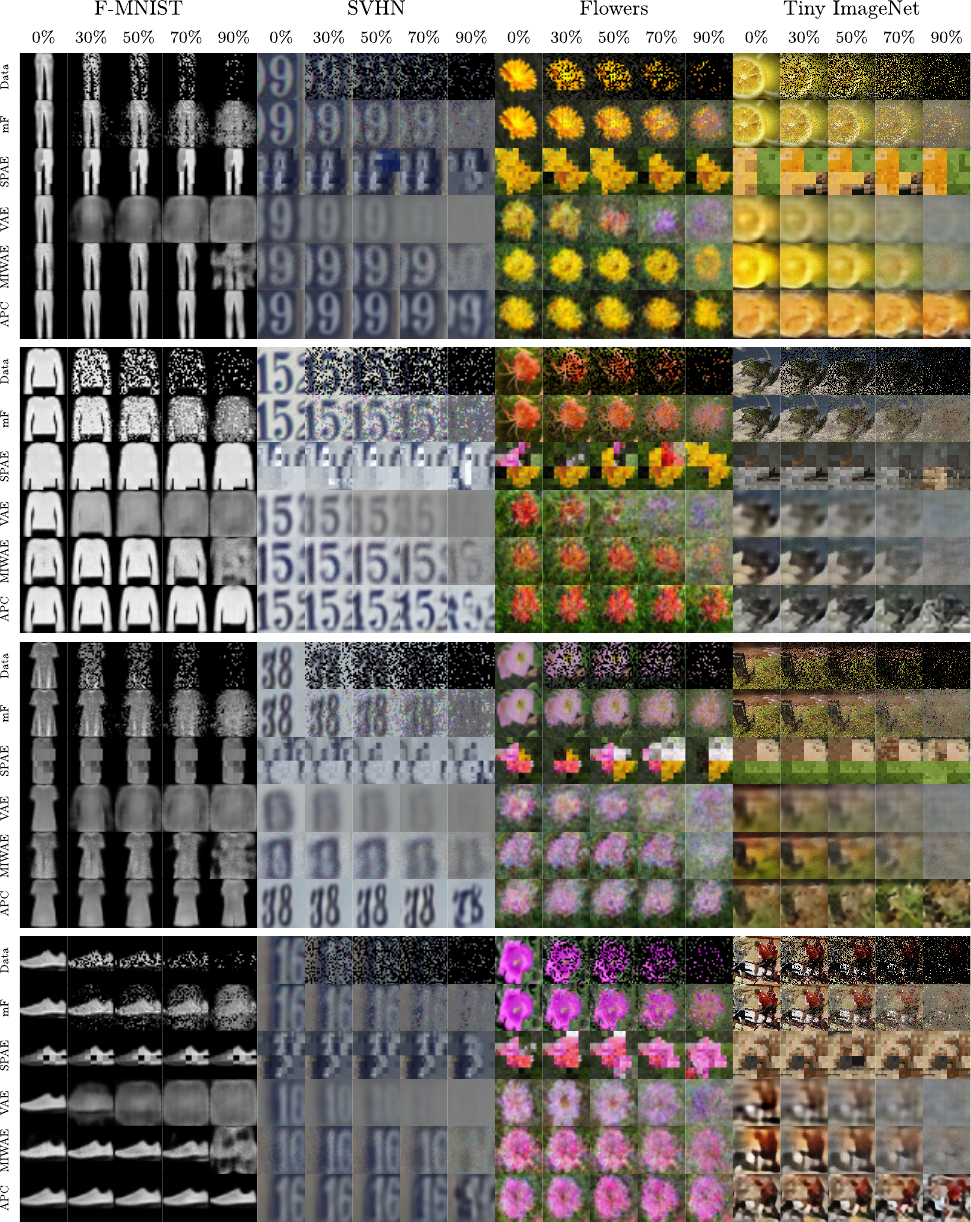 }
  \caption{Reconstructions of various MCAR corruptions for F-MNIST, SVHN, Flowers, and Tiny-ImageNet.}
  \label{app:fig:rec-examples-B}
\end{figure}

\begin{figure}[t]
  \centering
  \tablesize \includegraphics[width=\textwidth]{ 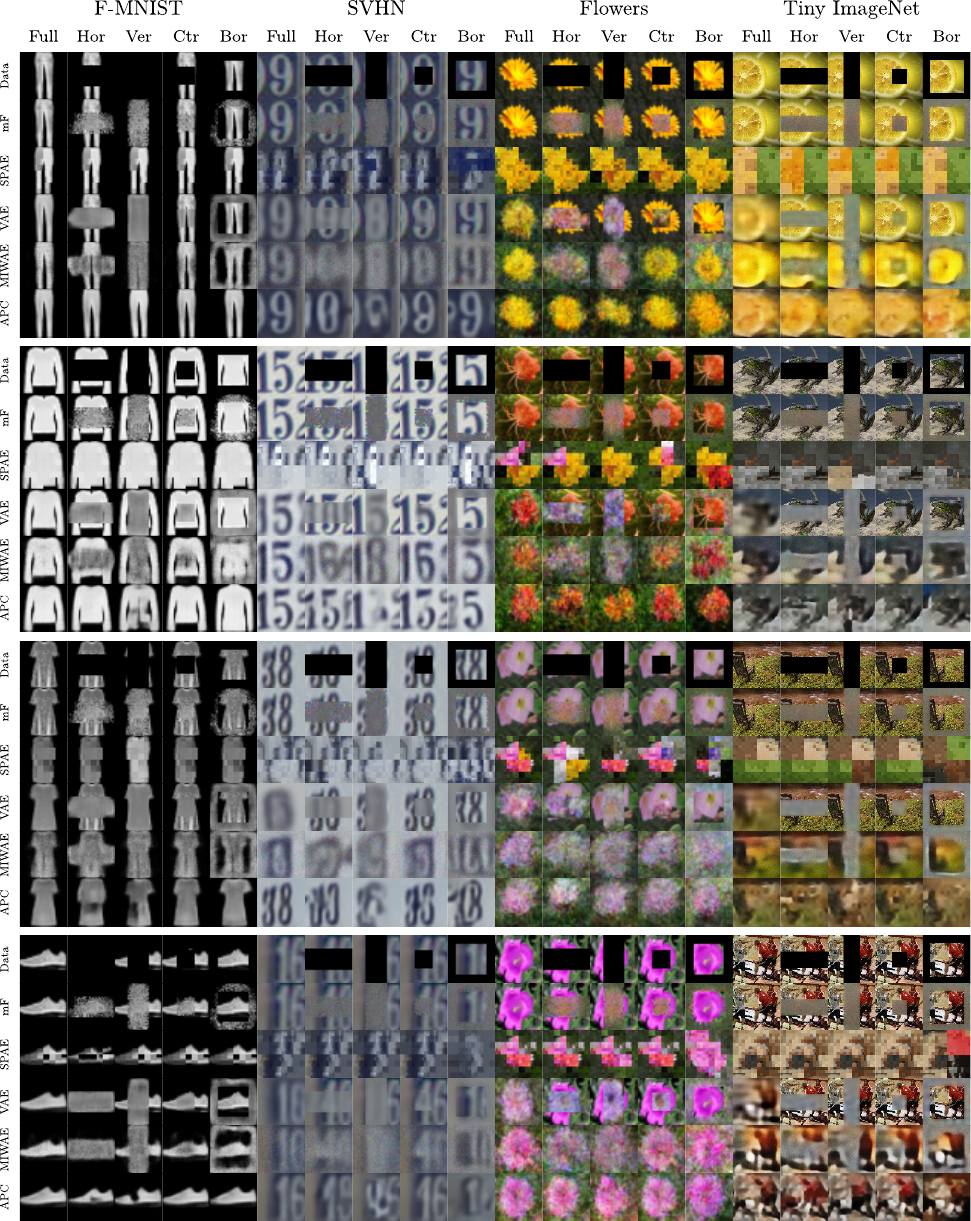 }
  \caption{Reconstructions of various MAR corruptions for F-MNIST, SVHN, Flowers, and Tiny-ImageNet.}
  \label{app:fig:rec-examples-D}
\end{figure}

To provide further visual insights and extend the quantitative reconstructions presented in \cref{fig:rec-examples}, we additionally show MCAR-style corruptions for 0\%, 30\%, 50\%, 70\%, and 90\% uniformly random missing data in \cref{app:fig:rec-examples-A,app:fig:rec-examples-B} and horizontal bands, vertical band, center and border MAR-style corruptions in \cref{app:fig:rec-examples-C,app:fig:rec-examples-D} on all image datasets for all methods. These results further provide evidence that \methods outperform the alternative methods in all investigated corruption scenarios, providing better reconstructions across the bench.

\cleardoublepage

\section{Autoencoding Probabilistic Circuit Variants}
\label{app:sec:eval:variants}

Having established the contribution of each component through our ablation studies in \cref{sec:eval:ablation}, we now build
upon the core \method framework and additionally explore several alternative configurations motivated by previous work of
sum-product autoencoding~\citep{vergari2018spae}. These variations allow us to examine the impact of different
encoding and decoding strategies within the \method framework. Specifically, we consider the following:

\begin{itemize}
  \item \textbf{\method$_{\text{pc}}$}: We replace the neural network decoder with the circuit encoder $\gC$ itself.
        Leveraging the flexibility of PCs, reconstructions $\hat{\x}$ are generated by sampling from the conditional
        distribution $\fn{p_{\gC}}{\X \cbar \Z}$. Differentiable sampling enables end-to-end training of this
        configuration, using the same objective as standard \methods to optimize the circuit for both encoding and decoding.
  \item \textbf{\method-SPAE$_{\text{pc}}$}: This configuration trains an \method using the SPAE scheme for encoding and
        decoding, employing the circuit $\gC$ for both. This means the encoding and decoding processes are based on ACT
        embeddings (CAT embeddings lead to identical reconstructions as per \cite{vergari2018spae}). Note, that
        KLD regularization of embeddings is omitted due to the different nature of ACT and CAT embeddings compared to
        the default \method formulation, and $\mathcal{L}_{\text{NLL}}$ is limited to $\fn{p_{\gC}}{\X}$, as explicit
        embedding representations $\Z$ are not directly modeled in the circuit in this variant.
  \item \textbf{\method-SPAE-ACT$_{\text{nn}}$}: We combine the ACT encoding from SPAE with a neural decoder. Similar to
        \method-SPAE$_{\text{pc}}$, KLD regularization is not applied and $\mathcal{L}_{\text{NLL}}$ is limited to
        $\fn{p_{\gC}}{\X}$. In our experiments, we found that due to circuit activation statistics, it was necessary to
        add a batch normalization layer on top of the circuit activations for the training with a neural decoder to be
        stable.
  \item \textbf{\method-SPAE-CAT$_{\text{nn}}$}: Analogous to \method-SPAE-ACT$_{\text{nn}}$, this variant utilizes the CAT
        embedding encoding approach from SPAE, coupled with a neural decoder for reconstruction. Training follows the
        same strategy as \method-SPAE-ACT$_{\text{nn}}$, excluding KLD regularization and limiting
        $\mathcal{L}_{\text{NLL}}$ to $\fn{p_{\gC}}{\X}$.
\end{itemize}

\begin{table}[t]
  \caption{%
\textbf{\methods achieve lower reconstruction errors than PC and SPAE-based variants across varying levels of MCAR data}.
  Average reconstruction performance on eight image datasets under increasing percentages of MCAR-style
    randomly missing pixels (0\%-95\%).
    }

  \label{tab:reconstruction-images-apc-variants}

  \begin{center}
    \small 
\begin{tabular}{lrrrrrr}
    \maybetoprule

            &                    &                    &                    & \header{APC-}               & \header{APC-}                   & \header{APC-}                   \\
            & \header{APC}       & \header{APC\pc}    & \header{SPAE}      & \header{SPAE$_{\text{pc}}$} & \header{SPAE-ACT$_{\text{nn}}$} & \header{SPAE-CAT$_{\text{nn}}$} \\
    \midrule
    \multicolumn{7}{c}{MSE $\downarrow$}                                                                                                                           \\

MNIST    & \resB{5.07}{0.04}   & \res{17.17}{0.41}  & \res{28.46}{0.43}   & \res{49.82}{3.01}   & \res{14.71}{0.25}   & \res{20.07}{1.00}   \\
F-MNIST  & \resB{4.33}{0.02}   & \res{17.09}{0.89}  & \res{18.96}{0.11}   & \res{49.46}{2.75}   & \res{12.79}{0.62}   & \res{17.97}{3.58}   \\
CIFAR    & \resB{20.65}{0.09}  & \res{59.22}{0.30}  & \res{78.32}{3.09}   & \res{102.74}{0.93}  & \res{31.88}{0.53}   & \res{42.44}{0.26}   \\
CelebA   & \resB{166.89}{0.65} & \res{644.36}{14.2} & \res{1318.73}{6.37} & \res{2969.30}{235.} & \res{1173.57}{156.} & \res{1220.76}{318.} \\
SVHN     & \resB{4.87}{0.00}   & \res{32.27}{0.38}  & \res{35.38}{1.92}   & \res{78.01}{4.37}   & \res{17.59}{0.24}   & \res{18.74}{0.35}   \\
Flowers  & \resB{45.22}{0.14}  & \res{58.65}{0.72}  & \res{119.31}{5.23}  & \res{110.05}{6.19}  & \res{72.46}{0.60}   & \res{75.05}{2.91}   \\
LSUN     & \resB{63.70}{0.20}  & \res{185.92}{3.62} & \res{322.22}{3.48}  & \res{489.87}{18.7}  & \res{194.10}{16.6}  & \res{185.69}{3.11}  \\
ImageNet & \resB{118.70}{0.16} & \res{276.92}{3.07} & \res{365.06}{2.20}  & \res{566.89}{17.2}  & \res{369.20}{35.7}  & \res{267.96}{20.6}  \\

    \midrule
    \multicolumn{7}{c}{SSIM $\uparrow$} \\

MNIST & \resB{93.33}{0.04} & \res{79.87}{0.57}  & \res{69.87}{0.36} & \res{45.21}{3.15} & \res{82.63}{0.35} & \res{64.52}{2.46}                     \\
F-MNIST                    & \resB{87.40}{0.05} & \res{69.44}{1.00} & \res{68.89}{0.23} & \res{39.11}{1.18} & \res{75.81}{0.58} & \res{60.89}{4.83} \\
CIFAR                      & \resB{78.20}{0.07} & \res{58.23}{0.31} & \res{50.61}{0.88} & \res{44.18}{0.17} & \res{71.41}{0.18} & \res{61.31}{0.14} \\
CelebA                     & \resB{79.52}{0.04} & \res{58.79}{0.33} & \res{49.02}{0.17} & \res{26.71}{1.17} & \res{49.03}{0.83} & \res{44.56}{3.77} \\
SVHN                       & \resB{90.29}{0.03} & \res{62.82}{0.24} & \res{54.88}{0.91} & \res{39.22}{0.76} & \res{75.60}{0.25} & \res{67.86}{0.36} \\
Flowers                    & \resB{60.40}{0.16} & \res{59.16}{0.15} & \res{44.01}{0.82} & \res{40.80}{0.87} & \res{50.35}{0.18} & \res{51.86}{1.12} \\
LSUN                       & \resB{77.14}{0.07} & \res{59.09}{0.38} & \res{50.73}{0.52} & \res{37.45}{0.93} & \res{58.72}{0.89} & \res{55.39}{0.39} \\
ImageNet                   & \resB{71.55}{0.04} & \res{55.68}{0.29} & \res{50.62}{0.36} & \res{37.90}{0.77} & \res{55.43}{0.45} & \res{53.50}{0.97} \\

    \maybebottomrule
\end{tabular}

  \end{center}
\end{table}

\begin{table}[t]
  \caption{%
\textbf{\methods achieve lower reconstruction errors than PC and SPAE-based variants across varying levels of MCAR data}.
  Average reconstruction performance on the 20 DEBD binary tabular datasets under increasing percentages of MCAR-style
    randomly missing pixels (0\%-95\%).
    }

  \label{tab:reconstruction-debd-apc-variants}
  \begin{center}
    \small \begin{tabular}{lrrrrrrr}
  \maybetoprule

            &                    &                    &                    & \header{APC-}               & \header{APC-}                   & \header{APC-}                   \\
            & \header{APC}       & \header{APC\pc}    & \header{SPAE}      & \header{SPAE$_{\text{pc}}$} & \header{SPAE-ACT$_{\text{nn}}$} & \header{SPAE-CAT$_{\text{nn}}$} \\
      \midrule
accidents   & \resB{6.16}{0.04}  & \res{29.16}{0.52}  & \res{12.55}{0.86}  & \res{22.16}{1.95}           & \res{8.31}{0.31}                & \res{6.75}{0.01}                \\
ad          & \resB{5.57}{0.03}  & \res{346.14}{5.94} & \res{275.00}{68.1} & \res{374.18}{15.7}          & \res{7.88}{0.41}                & \resB{5.57}{0.03}               \\
baudio      & \resB{6.50}{0.03}  & \res{23.31}{1.31}  & \res{15.60}{1.33}  & \res{16.81}{0.72}           & \res{7.27}{0.16}                & \res{7.50}{0.01}                \\
bbc         & \resB{34.66}{0.15} & \res{238.77}{3.22} & \res{161.59}{10.8} & \res{236.07}{4.29}          & \res{42.77}{0.75}               & \res{35.02}{0.15}               \\
bnetflix    & \resB{10.08}{0.02} & \res{24.87}{0.64}  & \res{20.53}{0.96}  & \res{21.89}{0.34}           & \res{10.56}{0.15}               & \res{10.80}{0.00}               \\
book        & \resB{3.65}{0.02}  & \res{121.44}{2.65} & \res{67.10}{9.40}  & \res{95.96}{4.73}           & \res{3.89}{0.18}                & \res{3.87}{0.03}                \\
c20ng       & \resB{19.12}{0.04} & \res{215.58}{5.08} & \res{123.50}{5.85} & \res{187.31}{5.87}          & \res{20.60}{0.26}               & \res{20.34}{0.02}               \\
cr52        & \resB{11.94}{0.10} & \res{201.12}{5.98} & \res{134.30}{7.81} & \res{217.42}{10.8}          & \res{14.33}{0.34}               & \res{12.75}{0.09}               \\
cwebkb      & \resB{21.32}{0.14} & \res{193.05}{3.82} & \res{109.07}{4.38} & \res{189.50}{7.03}          & \res{24.40}{0.47}               & \res{22.70}{0.07}               \\
dna         & \resB{15.89}{0.12} & \res{46.77}{1.38}  & \res{28.76}{0.91}  & \res{34.41}{1.21}           & \res{21.19}{0.45}               & \resB{15.89}{0.04}              \\
jester      & \resB{9.04}{0.02}  & \res{24.39}{0.47}  & \res{20.98}{0.33}  & \res{21.08}{0.83}           & \res{10.20}{0.12}               & \res{10.66}{0.01}               \\
kdd         & \resB{0.19}{0.00}  & \res{13.20}{1.73}  & \res{4.12}{0.57}   & \res{7.55}{1.30}            & \res{0.33}{0.03}                & \res{0.20}{0.00}                \\
kosarek     & \resB{1.35}{0.04}  & \res{43.28}{2.04}  & \res{13.57}{3.63}  & \res{36.47}{3.56}           & \res{2.15}{0.11}                & \res{1.40}{0.00}                \\
moviereview & \resB{48.06}{0.15} & \res{226.20}{4.90} & \res{141.19}{4.27} & \res{234.37}{10.9}          & \res{60.11}{0.99}               & \res{48.36}{0.06}               \\
msnbc       & \res{1.05}{0.03}   & \res{2.68}{0.43}   & \res{2.23}{0.39}   & \res{2.12}{0.34}            & \res{1.67}{0.01}                & \resB{0.99}{0.00}               \\
nltcs       & \resB{1.12}{0.02}  & \res{3.26}{0.27}   & \res{2.48}{0.00}   & \res{2.86}{0.39}            & \res{1.66}{0.14}                & \res{1.48}{0.00}                \\
plants      & \resB{2.52}{0.04}  & \res{16.44}{1.20}  & \res{11.62}{0.54}  & \res{12.04}{0.40}           & \res{3.22}{0.16}                & \res{4.67}{0.01}                \\
pumsb\_star & \resB{5.82}{0.16}  & \res{42.21}{1.94}  & \res{23.30}{0.74}  & \res{32.76}{1.53}           & \res{8.99}{0.65}                & \res{11.74}{0.02}               \\
tmovie      & \resB{7.32}{0.06}  & \res{116.74}{1.89} & \res{57.94}{5.49}  & \res{90.92}{10.4}           & \res{10.39}{0.34}               & \res{10.11}{0.02}               \\
tretail     & \resB{1.22}{0.01}  & \res{30.08}{2.20}  & \res{6.22}{1.63}   & \res{34.79}{4.17}           & \res{2.12}{0.25}                & \resB{1.22}{0.00}               \\
voting      & \resB{27.71}{0.18} & \res{311.78}{3.31} & \res{264.53}{7.30} & \res{303.02}{13.6}          & \res{91.81}{6.90}               & \res{119.05}{0.11}              \\
         \maybebottomrule
\end{tabular}

  \end{center}
\end{table}

In \cref{tab:reconstruction-images-apc-variants,tab:reconstruction-debd-apc-variants}, we report the reconstruction
performance of all \method variants on image and tabular datasets with progressively increasing proportions of randomly
missing pixels, contrasting them with full \method and SPAE models. We evaluate reconstruction quality using the average MCAR-style corruption reconstruction error for both MSE and SSIM, which
capture the impact of varying corruption levels. As shown, \methods maintain the overall best performance across all
datasets. However, we make three key observations. (1) Utilizing the encoding PC directly as a
decoder (\method\pc) yields inferior performance across all tasks, with a median relative decrease of $7.3\times$ on tabular
data $2.7\times$ in image data reconstruction, compared to employing a neural network decoder, highlighting, that the
additional modeling capacity and flexibility of a neural decoder is crucial.
(2) Integrating the SPAE scheme within the \method framework (\method-SPAE\pc) also led to a reduction in performance, further
suggesting that a PC decoder is less effective in this context. (3) Incorporating SPAE-ACT and SPAE-CAT encoding
mechanisms into the \method framework, while retaining a neural network decoder, consistently improved upon the performance
of the baseline SPAE method in all evaluated scenarios, achieving median improvements of $2.5\times$ for ACT and
$1.7\times$ for CAT embeddings on image data, and median $3.9\times$ and $4.8\times$ respectively on tabular datasets.
We use median values here to reduce the impact of extreme outliers, which we obtained on some of the datasets.

We want to highlight that CAT embeddings~\citep{vergari2018spae} can be interpreted as a special
case of \methods. They are the result of running \texttt{MaxProdMPE} on the circuit $\bar{\gC}$ over
$\V = (\X, \mathbf{H})$ where $\mathbf{H}$ are sum unit indicator random variables obtained from the latent variable
interpretation introduced in \cite{peharz2017latent} and described in \cref{sec:background}. In contrast, the \method framework allows embedding random
variables to appear at \emph{arbitrary positions} in the circuit graph represented by \emph{arbitrary distributions}.
This comparison highlights the flexibility of \methods compared to the prior autoencoding scheme introduced in
\cite{vergari2018spae}.

\cleardoublepage

\section{Knowledge Distillation: Algorithm \& Additional Results}
\label{app:knowledge-distillation}

\begin{algorithm}[H]%
  \caption{Data-Free Knowledge Distillation from \text{VAE} to \text{\method}}
  \label{alg:knowledge-distillation}
  \begin{algorithmic}[1]
    \Require Pre-trained Teacher VAE, Student \method, Iterations $T$
    \Procedure{KnowledgeDistillation}{$\text{VAE}, \text{\method}, T$}
      \For{$t = 1, 2, \ldots, T$}
        \State $\z \sim \fn{p_{\gC}}{\Z}$ \Comment{Sample embedding from $\text{\method}$ prior}
        \State $\hat{\x}_{\text{VAE}} = \fn{g_{\text{VAE}}}{\z}$ \Comment{Generate synthetic data with $\text{VAE}$ decoder}
        \State $\z_{\text{VAE}} = \fn{f_{\text{VAE}}}{\hat{\x}_{\text{VAE}}}$ \Comment{Encode synthetic data with $\text{VAE}$ encoder}
        \State $\z_{\text{\method}} \sim \fn{p_{\gC}}{\Z \cbar \x_{\text{VAE}}}$ \Comment{Encode synthetic data with $\text{\method}$ encoder}
        \State $\hat{\x}_{\text{\method}} = \fn{g_{\text{\method}}}{\z_{\text{\method}}}$ \Comment{Reconstruct with $\text{\method}$ decoder}
        \State $\mathcal{L} = \fn{\text{MSE}}{\hat{\x}_{\text{VAE}}, \hat{\x}_{\text{\method}}} + \fn{\text{KLD}}{\fn{p_{\gC}}{\Z \cbar \hat{\x}_{\text{VAE}}} \ccbar \fn{\mathcal{N}}{0, 1}} - \log \fn{p_{\gC}}{\hat{\x}_{\text{VAE}}, \z_{\text{VAE}}}$ \Comment{Distillation loss}
        \State Update $\text{\method}$ parameters to minimize $\mathcal{L}$ \Comment{Gradient descent}
      \EndFor
      \State \textbf{return} Distilled $\text{\method}$
    \EndProcedure
  \end{algorithmic}
\end{algorithm}

\cref{alg:knowledge-distillation} outlines the process of data-free knowledge distillation from a VAE to an \method of similar capacity and the same decoder architecture. The procedure
iteratively refines the \method to learn from the VAE without the need for original training data. At each iteration, an
embedding $\z$ is first sampled from the \method prior $\fn{p_{\gC}}{\Z}$. This embedding is then passed through the VAE
decoder $\fn{g_{\text{VAE}}}{\z}$ to generate synthetic data $\hat{\x}_{\text{VAE}}$ that aligns with the \method prior. The
synthetic data is subsequently re-encoded using the VAE encoder
$\z_{\text{VAE}} = \fn{f_{\text{VAE}}}{\hat{\x}_{\text{VAE}}}$, to obtain an embedding according to the VAE's
approximate posterior. In parallel, the \method encodes the same synthetic data using
$\fn{p_{\gC}}{\Z \cbar \x_{\text{VAE}}}$, obtaining an auxiliary embedding $\z_{\text{\method}}$. This embedding is then
passed through the \method decoder $\fn{g_{\text{\method}}}{\z_{\text{\method}}}$ to reconstruct the synthetic VAE sample, producing
$\hat{\x}_{\text{\method}}$. The distillation loss $\mathcal{L}$ is equal to the main training recipe of \methods and consists
of three components: the mean squared error (MSE) between the VAE and \method reconstructions, a Kullback-Leibler divergence
(KLD) term that encourages the \method’s posterior $\fn{p_{\gC}}{\Z \cbar \hat{\x}_{\text{VAE}}}$ to align with a standard
normal prior $\fn{\mathcal{N}}{0, 1}$, and a negative log-likelihood (NLL) term maximizing the joint likelihood of VAE
sample and embedding $\fn{p_{\gC}}{\hat{\x}_{\text{VAE}}, \z_{\text{VAE}}}$. Since all steps are differentiable, we can
end-to-end train this procedure from Line 3 to Line 7 and update the \method parameters via gradient descent to minimize
$\mathcal{L}$.

\begin{table}[t]
  \caption{\textbf{\methods successfully distill the knowledge from a pre-trained VAE in a data-free setting and exceed their teachers robustness against missing data.} Input reconstruction mean squared error ($\downarrow$) and downstream task
    accuracy ($\uparrow$) performance comparison between VAE teacher and distilled \method student models, evaluated with
    full evidence (Full Evi.) and under varying levels of missing data (MCAR). }
  \label{tab:latent-kd}
  \normalsize
  \begin{minipage}{0.48\linewidth}
    \centering
    \caption*{\textbf{Full Evidence}: \methods are successfully distilling the knowledge from their teacher.}
      \begin{tabular}{lr@{\hspace{0.1cm}}c@{\hspace{0.1cm}}r@{\hspace{0.75cm}}}

    \maybetoprule
& \header{Teacher} & $\rightarrow$    & \header{Student} \\
    \midrule
    \multicolumn{4}{c}{MSE ($\downarrow$)} \\[0.25em]
 MNIST       & \res{3.15}{0.02}    & & \res{6.84}{0.17}    \\
 F-MNIST     & \res{5.84}{0.02}    & & \res{8.16}{0.10}    \\
 CIFAR       & \res{21.84}{0.07}   & & \res{25.34}{0.30}   \\
 CelebA      & \res{336.00}{22.7}  & & \res{402.53}{21.1}  \\
 Flowers     & \res{115.04}{2.12}  & & \res{57.66}{1.33}   \\
 LSUN        & \res{104.91}{14.3}  & & \res{119.07}{11.1}  \\
 SVHN        & \res{15.04}{0.48}   & & \res{13.14}{0.50}   \\
 ImageNet    & \res{1375.88}{30.5} & & \res{230.16}{11.3}  \\
 accidents   & \res{10.38}{0.67}   & & \res{19.03}{1.79}   \\
 ad          & \res{3.97}{0.22}    & & \res{11.75}{0.84}   \\
 baudio      & \res{11.47}{0.18}   & & \res{20.35}{2.95}   \\
 bbc         & \res{71.49}{0.55}   & & \res{80.06}{3.29}   \\
 bnetflix    & \res{17.61}{0.24}   & & \res{21.59}{3.31}   \\
 book        & \res{7.47}{0.06}    & & \res{13.09}{3.13}   \\
 c20ng       & \res{38.61}{0.03}   & & \res{64.73}{10.3}   \\
 cr52        & \res{22.66}{0.21}   & & \res{31.68}{2.27}   \\
 cwebkb      & \res{43.63}{0.36}   & & \res{57.91}{3.21}   \\
 dna         & \res{32.34}{0.07}   & & \res{40.59}{3.39}   \\
 jester      & \res{16.38}{0.23}   & & \res{25.35}{2.74}   \\
 kdd         & \res{0.37}{0.00}    & & \res{0.43}{0.04}    \\
 kosarek     & \res{2.36}{0.03}    & & \res{2.96}{0.15}    \\
 moviereview & \res{102.20}{0.80}  & & \res{131.58}{13.0}  \\
 msnbc       & \res{1.58}{0.09}    & & \res{1.96}{0.02}    \\
 nltcs       & \res{1.05}{0.00}    & & \res{1.32}{0.02}    \\
 plants      & \res{2.65}{0.23}    & & \res{3.55}{0.37}    \\
 pumsb\_star & \res{8.14}{0.70}    & & \res{20.08}{5.77}   \\
 tmovie      & \res{13.15}{0.03}   & & \res{19.06}{1.22}   \\
 tretail     & \res{2.20}{0.00}    & & \res{2.54}{0.06}    \\
     voting  & \res{51.00}{0.97}   & & \res{57.25}{2.31}   \\

\midrule
    \multicolumn{4}{c}{Downstream Accuracy ($\uparrow$)} \\[0.25em]
MNIST    & \res{96.12}{0.12}&  & \res{90.39}{0.07} \\
F-MNIST  & \res{83.02}{0.16}&  & \res{81.04}{0.21} \\
SVHN     & \res{56.92}{3.78}&  & \res{23.98}{0.35} \\
CIFAR    & \res{47.14}{0.37}&  & \res{38.62}{0.42} \\
Flowers  & \res{9.87}{0.65} &  & \res{12.84}{0.23} \\
ImageNet & \res{6.60}{0.11} &  & \res{2.97}{0.14}  \\
\maybebottomrule
  \end{tabular}

  \end{minipage}
  \hfill
  \begin{minipage}{0.48\linewidth}
    \centering
    \caption*{\textbf{MCAR Corruptions}: \methods surpass their teacher in robustness against corruptions.}
      \begin{tabular}{lr@{\hspace{0.1cm}}c@{\hspace{0.1cm}}r@{\hspace{0.75cm}}}

    \maybetoprule
            & \header{Teacher} & $\rightarrow$    & \header{Student} \\
    \midrule
    \multicolumn{4}{c}{MSE ($\downarrow$)} \\[0.25em]
 MNIST       & \res{35.38}{1.74}   & & \res{6.15}{0.11}    \\
 F-MNIST     & \res{39.80}{1.10}   & & \res{5.33}{0.05}    \\
 CIFAR       & \res{55.81}{0.47}   & & \res{19.61}{0.10}   \\
 CelebA      & \res{1095.90}{14.0} & & \res{201.70}{10.4}  \\
 Flowers     & \res{107.75}{11.8}  & & \res{31.43}{0.51}   \\
 LSUN        & \res{254.75}{2.39}  & & \res{68.71}{4.23}   \\
 SVHN        & \res{41.01}{0.15}   & & \res{7.68}{0.19}    \\
 ImageNet    & \res{1286.51}{10.4} & & \res{139.72}{6.39}  \\
 accidents   & \res{7.56}{0.28}    & & \res{7.76}{0.37}    \\
 ad          & \res{6.04}{0.17}    & & \res{5.68}{0.23}    \\
 baudio      & \res{7.44}{0.01}    & & \res{7.94}{0.40}    \\
 bbc         & \res{37.45}{0.55}   & & \res{39.04}{1.20}   \\
 bnetflix    & \res{13.49}{0.77}   & & \res{10.46}{0.41}   \\
 book        & \res{3.73}{0.02}    & & \res{6.12}{0.85}    \\
 c20ng       & \res{20.69}{0.05}   & & \res{29.86}{1.85}   \\
 cr52        & \res{12.83}{0.18}   & & \res{15.98}{0.61}   \\
 cwebkb      & \res{22.94}{0.15}   & & \res{28.13}{1.24}   \\
 dna         & \res{16.46}{0.17}   & & \res{19.29}{1.09}   \\
 jester      & \res{17.44}{0.10}   & & \res{10.37}{0.43}   \\
 kdd         & \res{0.19}{0.00}    & & \res{0.22}{0.01}    \\
 kosarek     & \res{1.42}{0.01}    & & \res{1.58}{0.05}    \\
 moviereview & \res{50.08}{0.22}   & & \res{62.55}{3.66}   \\
 msnbc       & \res{1.03}{0.01}    & & \res{1.07}{0.04}    \\
 nltcs       & \res{1.60}{0.01}    & & \res{1.12}{0.01}    \\
 plants      & \res{3.93}{0.05}    & & \res{2.55}{0.05}    \\
 pumsb\_star & \res{12.47}{0.40}   & & \res{9.22}{2.80}    \\
 tmovie      & \res{8.72}{0.03}    & & \res{8.81}{0.33}    \\
 tretail     & \res{1.23}{0.00}    & & \res{1.22}{0.01}    \\
     voting  & \res{122.24}{52.8}  & & \res{28.94}{0.82}   \\

    \midrule
    \multicolumn{4}{c}{Downstream Accuracy ($\uparrow$)} \\[0.25em]
 MNIST    & \res{27.49}{2.23} & & \res{87.77}{0.12} \\
 F-MNIST  & \res{30.30}{2.64} & & \res{79.41}{0.24} \\
 SVHN     & \res{36.38}{1.99} & & \res{22.93}{0.29} \\
 CIFAR    & \res{33.53}{0.50} & & \res{37.66}{0.33} \\
 Flowers  & \res{4.53}{0.35}  & & \res{12.56}{0.27} \\
 ImageNet & \res{5.54}{0.37}  & & \res{2.97}{0.15}  \\
\maybebottomrule
  \end{tabular}

  \end{minipage}
\end{table}

To demonstrate that knowledge distillation from VAEs to \methods generalizes beyond the two exemplary datasets considered in \cref{sec:eval:knowledge-distillation}, we extend our evaluation to all image and tabular datasets introduced in \cref{sec:eval}. We report results for both teacher and student reconstructions, and downstream task accuracy under full evidence (Full Evi.) and MCAR-style corruptions in \cref{tab:latent-kd}.

\cleardoublepage

\section{Embedding-based Out-of-Distribution Detection Using \methods}

\begin{figure}[H]
  \centering
  \includegraphics[width=\textwidth]{ 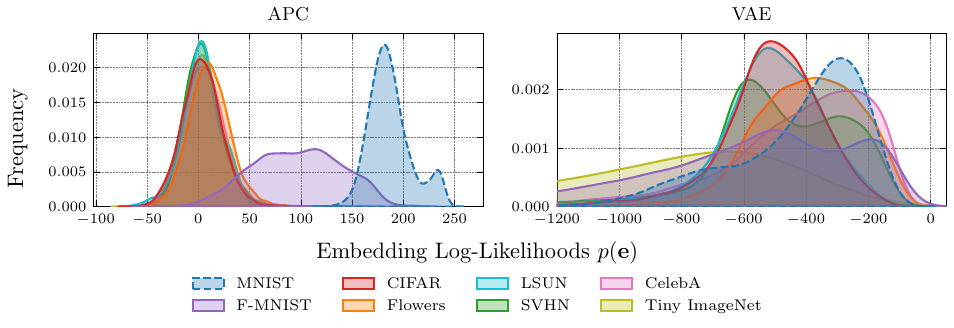 }
  \caption{
\textbf{While VAEs map in-distribution and out-of-distribution data to the same
    embedding range, \methods cleanly separate MNIST from other datasets.} As expected, we see some minor overlap
    with F-MNIST due to their similar pixel statistics, whereas all other datasets have close to no overlap with MNIST.
    \method and VAE models under the lens of out-of-distribution data: Models were trained on MNIST and embedding
    log-likelihoods $\fn{p}{\z}$ were evaluated for a range of out-of-distribution datasets. For \methods we can tractably
    obtain $\fn{p}{\z} = \int \fn{p}{\x, \z}d\x$ with marginalization. For VAEs, we need to compute an aggregate
    posterior from training dataset embeddings. }
  \label{fig:ood-mnist}
\end{figure}

In addition to generative tasks, we can also investigate the empirical distribution of embedding likelihoods produced by
the model. Such analysis could allow out-of-distribution (OOD) detection, enabling decoding processes or
downstream applications to proactively \textit{reject} embeddings that deviate significantly from the expected embedding
distribution, as determined by a predefined threshold. As an initial exploration, \cref{fig:ood-mnist} compares the
embedding log-likelihood distributions of \methods and VAEs for in-distribution (MNIST) and out-of-distribution datasets. We
selected MNIST for our OOD experiments because its hand-drawn digit images represent a highly constrained and artificial
distribution that differs from naturally occurring images. This distinctive characteristic, which MNIST shares with
F-MNIST, establishes a clear distributional boundary that theoretically enables models to assign lower likelihood
scores to out-of-distribution samples compared to in-distribution data points. As previously established, \methods allow us
to evaluate the marginal embedding log-likelihood exactly and tractably. At the same time, VAEs require approximating this
distribution using a post-hoc aggregate posterior from training data. As evident in \cref{fig:ood-mnist}, \method
embeddings separate in-distribution from out-of-distribution data in the likelihood space, with minimal overlap except
for F-MNIST, which shares similar pixel statistics with MNIST. In contrast, VAE embeddings exhibit substantial
overlap between in-distribution and out-of-distribution likelihood scores, making OOD detection almost impossible. This
demonstrates that \methods' explicit probabilistic formulation provides strong reconstruction and representation
learning capabilities and hints at out-of-distribution detection capabilities without requiring additional
mechanisms or models. We leave further in-depth analysis on this topic to future work.

\cleardoublepage

\section{Differentiable Sampling with SIMPLE: Details}
\label{app:simple}

\begin{algorithm}[H]%
  \caption{Sampling sum units with SIMPLE}
  \label{alg:simple}
  \begin{algorithmic}[1]
    \Require Normalized weights $\bm{\theta} \in [0,1]^D$ for sum unit $n$, where $D = |\inscope(n)|$.
    \Procedure{simple}{$\bm{\theta}$}
      \State $\bm{\log\theta} \gets \log(\bm{\theta})$ \Comment{Convert probabilities to log-probabilities (logits)}
      \State Draw $\mathbf{g} \sim \text{Gumbel}(0,1)^D$ \Comment{Draw independent Gumbel noise for each component}
      \State $\mathbf{z} \gets \bm{\log\theta} + \mathbf{g}$ \Comment{Perturb log-probabilities with Gumbel noise}
      \State $d^* \gets \argmax_{j=1,\ldots,D} z_j$ \Comment{Identify the index of the maximum perturbed logit (discrete sample)}
      \State $\mathbf{s} \gets \text{one-hot}(d^*, D)$ \Comment{Construct a one-hot vector $\mathbf{s}$ where $s_{d^*} = 1$}
      \State \textbf{return} $(\mathbf{s} - \bm{\theta}).detach() + \bm{\theta}$ \Comment{Return $\mathbf{s}$ for forward pass, but only pass gradients of $\bm{\theta}$ (\textit{detach})}
    \EndProcedure
  \end{algorithmic}
\end{algorithm}

We additionally improve the differentiable sampling approach for PCs originally proposed in
\cite{lang2022diff-sampling-spns} by replacing the Gumbel-Softmax gradient estimator with
SIMPLE~\citep{ahmed2022simple} for $k$-subset sampling (with $k=1$) when sampling from sum units as outlined in \cref{alg:simple}. In contrast to Gumbel-Softmax, SIMPLE directly propagates the gradients through the unperturbed paramters $\bm{\theta}$. To quantitatively
evaluate this modification, we conducted a controlled experiment comparing both gradient estimators on a synthetic task:
learning a sum unit distribution with 32,64,128, and 256 inputs to match a known categorical ground truth distribution
over 64 categories. We optimized sum unit weights by minimizing the MSE between one-hot encoded samples from both
distributions, using AdamW with a learning rate of 0.01 for 1,000 iterations with batch size 64, repeating across 10
random seeds. As shown in \cref{fig:simple-vs-gumbel}, SIMPLE demonstrates faster convergence and higher accuracy in
approximating the true distribution, as measured by KLD. Notably, SIMPLE at iteration 100 achieves comparable
performance to Gumbel-Softmax at iteration 1,000, and continues to improve beyond that point. Furthermore, SIMPLE
demonstrates lower variance across different random initializations, indicating more consistent convergence behavior
regardless of starting conditions. Upon convergence, SIMPLE achieves a KLD of $0.0033 \pm 0.00065$, approximately
25$\times$ lower than Gumbel-Softmax's $0.0817 \pm 0.00602$ for $D=64$. These results validate our choice of SIMPLE as
the gradient estimator for differentiable sampling in \methods, enabling more accurate and stable training.

\end{document}